\pdfoutput=1
\documentclass[11pt]{article}

\usepackage[preprint]{acl}

\usepackage{times}
\usepackage{latexsym}

\usepackage[T1]{fontenc}
\usepackage[utf8]{inputenc}

\usepackage{microtype}

\usepackage{inconsolata}

\usepackage{graphicx}

\usepackage{amsmath}
\usepackage{xcolor}
\usepackage{booktabs}
\usepackage{colortbl}
\usepackage{tcolorbox}
\usepackage{tabularx}

\definecolor{linebg}{HTML}{7AA2E3}
\definecolor{headerbg}{HTML}{C0DEFF} % Hex code FFF9D0
\definecolor{headertext}{HTML}{222831}
\definecolor{rowtext}{HTML}{90C8AC}

\usepackage{subcaption}

\newtcolorbox{mybox}{fontupper=\footnotesize}

\newcommand{\yt}[1]{\textcolor{orange}{#1}}

\newcommand{\eh}[1]{\textcolor{purple}{EH: #1}}
\newcommand{\mee}[1]{\textcolor{red}{#1}}

\title{When Context Leads but Parametric Memory Follows in Large Language Models}

\author{Yufei Tao, Adam Hiatt, Erik Haake, Antonie J. Jetter, Ameeta Agrawal \\
Portland State University, USA \\
\texttt{\{yutao, ahiatt, ehaake, ajetter, ameeta\}@pdx.edu}}

\begin{document}
\maketitle

% Abstract
\begin{abstract}

Large language models (LLMs) have demonstrated remarkable progress in leveraging diverse knowledge sources. This study investigates how nine widely used LLMs allocate knowledge between local context and global parameters when answering open-ended questions in knowledge-consistent scenarios. We introduce a novel dataset, \texttt{WikiAtomic}\footnote{The dataset is available at \url{https://github.com/PortNLP/WikiAtomic}.}, and systematically vary context sizes to analyze how LLMs prioritize and utilize the provided information and their parametric knowledge in knowledge-consistent scenarios. Additionally, we also study their tendency to hallucinate under varying context sizes. Our findings reveal consistent patterns across models, including a consistent reliance on both contextual (around 70\%) and parametric (around 30\%) knowledge, and a decrease in hallucinations with increasing context. These insights highlight the importance of more effective context organization and developing models that use input more deterministically for robust performance.

\end{abstract}

%\mee{GPT4o, Claude Opus, Sonnet, Haiku. Llama 70b, 8b, mixtral 8x22, mistral 7b, phi.}

% Dataset Creation Figure
\begin{figure*}[t]
    \centering
    \includegraphics[width=0.85\textwidth]{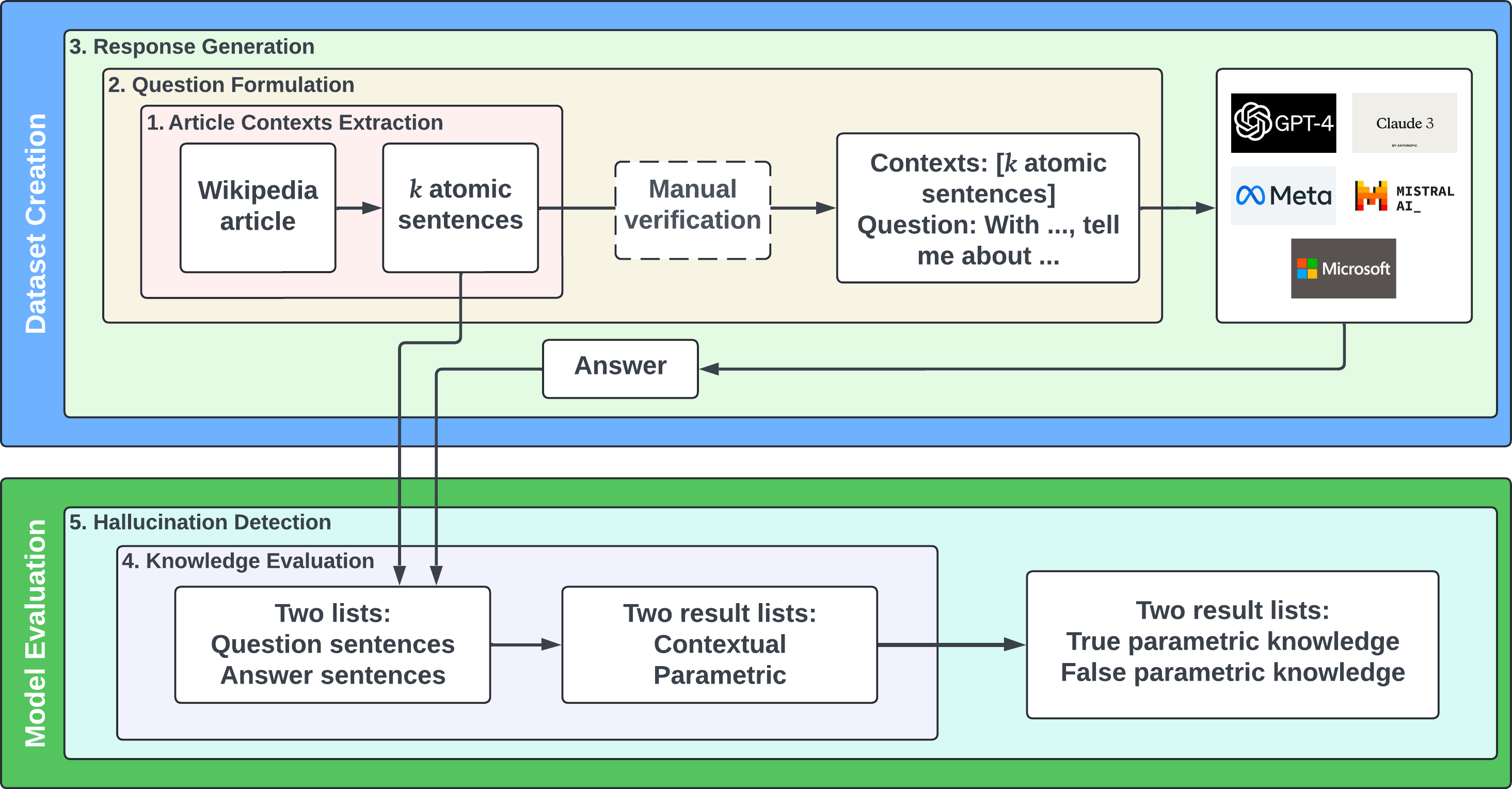}
    \caption{Overview of the dataset creation and model evaluation pipeline}
    \label{fig:pipeline}\vspace{-0.2cm}
\end{figure*}

% Introduction
\section{Introduction}

\begin{comment}
    Titles: 

\eh{Potential Titles: "Unmasking LLM Homogeneity Through Atomic Insights", "Uniform Patterns, Focused Minds: Atomic Insights into LLMs", that's all I got for now. I'll update if I think of any more.}

\mee{"Context Matters: Boosting Knowledge Consistency with More Context"}

\yt{All roads lead to Rome -> All LLMs lead to , no , Why are you all so similar? Way too similar. More contexts, more similar responses, All LLMs guess the answer the same way. with more contexts, they guess more accurate. LLMs are students answer question the same way. LLMs guess what we want in very similar way. How about \textbf{"Same Questions, Same Guess Patterns- The Uniformity of LLMs"}}

\mee{for inspiration} Fantastically ordered prompts and where to find them: Overcoming few-shot prompt order sensitivity
\end{comment}

%\mee{section 4.2 of this paper \url{https://arxiv.org/pdf/2403.08319}, starting with Elzahar's work talks about what seems like knowledge-consistent studies, would be good to do a quick survey and see if/how we're related} [knowledge-consistent in hallucination ... ]\eh{I didn't find anything particularly relevant to our paper}

%\cite{yan2023multimodal, li2024developing, hariri2024unlocking}

Large language models (LLMs) have significantly advanced the capabilities of natural language processing. When generating responses, LLMs can use the contextual information provided in a prompt along with or instead of the parametric knowledge embedded during pretraining \cite{petroni-etal-2019-language, Brown-fewShotLearners, heinzerling-inui-2021-language}. 

% This flexibility allows models to tailor responses to specific prompts and leverage extensive embedded knowledge, making LLMs valuable in fields like information retrieval \cite{InformationRetrievalLLMs}, machine translation \cite{zhu2023multilingual}, and question answering \cite{minaee2024large}.

%Parametric knowledge refers to the vast amounts of general knowledge encoded in the model's weights during pretraining, whereas contextual knowledge is the specific information provided in the prompt.

%Previous studies, such as those by \citet{neeman2022disentqa} and \citet{li2024prompting}, investigated whether large language models (LLMs) favor global parametric knowledge over local contextual information.  

In order to generate accurate and coherent responses, LLMs need to effectively combine their parametric knowledge with provided contextual information, and understanding the balance between these two sources of information is crucial \cite{neeman2022disentqa,li2024prompting}. Most prior work has explored this question through the lens of counterfactual data where the parametric and contextual knowledge are in conflict \cite{longpre-etal-2021-entity, xu2024knowledge}. Some studies suggest that the models prefer contextual knowledge, while others suggest they prioritize parametric knowledge \cite{krishna-etal-2021-hurdles, zhou2023contextfaithful}. 

% In many real-world scenarios, however, contextual knowledge complements rather than conflicts with parametric knowledge. In question answering tasks, a user may provide the model with true information about a subject to elicit a targeted response. Understanding how models integrate different sources of knowledge in knowledge-consistent scenarios is a critical.
In many real-world scenarios such as question answering or summarization, however, contextual knowledge may {\em complement} rather than conflict with parametric knowledge. Understanding how models integrate different sources of knowledge in knowledge-consistent scenarios is critical.

Moreover, while parametric knowledge enables LLMs to generate coherent text, it also contributes to the risk of hallucinations where responses are coherent and confident but factually incorrect or irrelevant \cite{Ji-Hallucination-Survey, Tian2023FinetuningLM, wang2023survey, luo2024hallucination}. %The likelihood of a model producing hallucinations is closely tied to the context provided within the prompt \cite{andriopoulos2023augmenting, liu2023lost, zheng2023does}. 
As such, our work also explores the hallucination tendency in knowledge-consistent scenarios.

%Inadequate or ambiguous context leads the model to rely excessively on its general training, which might result in responses that are erroneous \cite{mallen-etal-2023-trust}. 

Real-world applications often present LLMs with varying amounts of context. Building robust models requires understanding how effectively they leverage this information. This research investigates how the volume of context influences knowledge preference and hallucination tendencies in LLMs within a question answering (QA) framework. We study nine widely used LLMs to determine how different models navigate varying context sizes in knowledge-consistent settings, aiming to answer the main question: How do LLMs prioritize contextual and parametric knowledge when generating responses? Additionally, we investigate the likelihood of generating hallucinations with increasing context, which parts of contexts are used, how similar are various types of knowledge, and further analyses exploring potentially unseen knowledge.

Figure~\ref{fig:pipeline} provides an overview of our investigation pipeline. We create a new dataset, \texttt{WikiAtomic}, of atomic sentences serving as our context. We systematically adjust the amount of factual context provided to the models and prompt them in an open-ended QA task to study the responses generated by various models. Then, we measure the amount of contextual knowledge models recall, the additional parametric knowledge they rely on, and the extent of hallucinations found in their responses.

%Studying the effect of varying context input is crucial for understanding the conditions under which LLMs prioritize contextual knowledge provided in immediate context versus extensive parametric knowledge embedded during pretraining. This investigation also sheds light on the type of parametric knowledge models tend to include in their outputs and the extent to which it supplements contextual knowledge. 

% By utilizing our novel dataset WikiAtomic, we devise a series of experiments to investigate 9 popular LLMs. This newly created dataset opens up avenues to explore several research questions. \eh{Removing this since it's not entirely necessary and we need to cull sentences pretty aggressively at this point}

% In this work, we ask two questions: (1) which sources of knowledge do models prioritize under non-conflicting data from different sources where both contextual and parametric knowledge align? (2) what effect does this have on hallucination rate?

Our results suggest that all models behave very similarly, integrating up to 30\% parametric knowledge in their responses. For smaller contexts, models recall information from all parts of the context. However, for longer contexts, they predominantly focus on the first half, potentially missing key information from other parts of the input. The sequence of information in the responses largely mirrors the context and the additional parametric knowledge included is moderately similar to the contextual knowledge. Lastly, the hallucination score decreases as more context is added. These findings highlight the importance of effective context organization and models that utilize input more predictably.

%Slight adjustments to the question prompt can influence how much the model focuses on the contextual knowledge. 

%the models overly rely on the contextual knowledge, and may be missing out on deeper, more comprehensive insights that parametric knowledge can provide. The LLMs seem to be simply regurgitating information from the prompt without adding additional insights, analysis, or creativity.

Our key contributions include:
\begin{itemize}
    \item A novel dataset, \texttt{WikiAtomic}, comprised of prompt/response pairs in which provided context is incrementally increased.
    \item Comprehensive analyses showing how models utilize contextual and parametric and in what proportion.
    \item We discuss the relationship between hallucination tendency and the amount of context.
\end{itemize}

\section{Task and Terminology}
We structure our investigation as an open-ended question answering task to gain deeper insights into how LLMs utilize and integrate contextual knowledge with their parametric knowledge. By incrementally increasing the amount of context, we analyze the models' behavior, focusing on their ability to prioritize different types of information and their tendency toward factual hallucination in knowledge-consistent scenarios. %This approach allows us to better understand the mechanisms behind LLMs' knowledge integration and their reliability in generating accurate responses.

We next introduce some key terminology used throughout the paper.

{\textbf{Topic:} This is the title of the Wikipedia article.} The \texttt{WikiAtomic} dataset includes 200 distinct articles, and questions are asked about these topics.

\textbf{Atomic Sentence}: This is a sentence that contains a single piece of information that cannot be broken down into simpler components without losing its meaning \cite{liu-etal-2023-revisiting}. 

{\textbf{Context}: For each topic, a context consists of $k$ atomic sentences provided in a prompt. In our experiments, $2 \leq k \leq 50$, where contexts consist of increments of 2 sentences from 2 to 30 (e.g., 2, 4, 6, etc.), followed by increments of 5 sentences until reaching a total of 50 sentences. For each topic, this results in contexts of 20 different sizes, creating a total of 4,000 topic-context instances.

{\textbf{Response}: Given the context, the model generates a response that includes some degree of both contextual and parametric knowledge. This response is then atomized. A sentence directly derived from the provided context is considered as {\textbf{contextual  (local) knowledge}}. In contrast, a sentence that is {\em not} entailed from the context is considered as {\textbf{parametric (global) knowledge}. {Following prior work \cite{neeman2022disentqa}, contextual knowledge comes from external sources provided during inference, while parametric knowledge is knowledge encoded (or “memorized”) in the model parameters. As such, if a sentence  introduces information not present in the context, it is considered a parametric knowledge sentence.}

%$R_c$ and $R_p$ for overlap of responses with context and non-context (parametric), respectively. Given 9 LLMs, totaling 40,000 data points in our dataset.

\begin{figure*}
    \centering
    \includegraphics[width=0.93\linewidth]{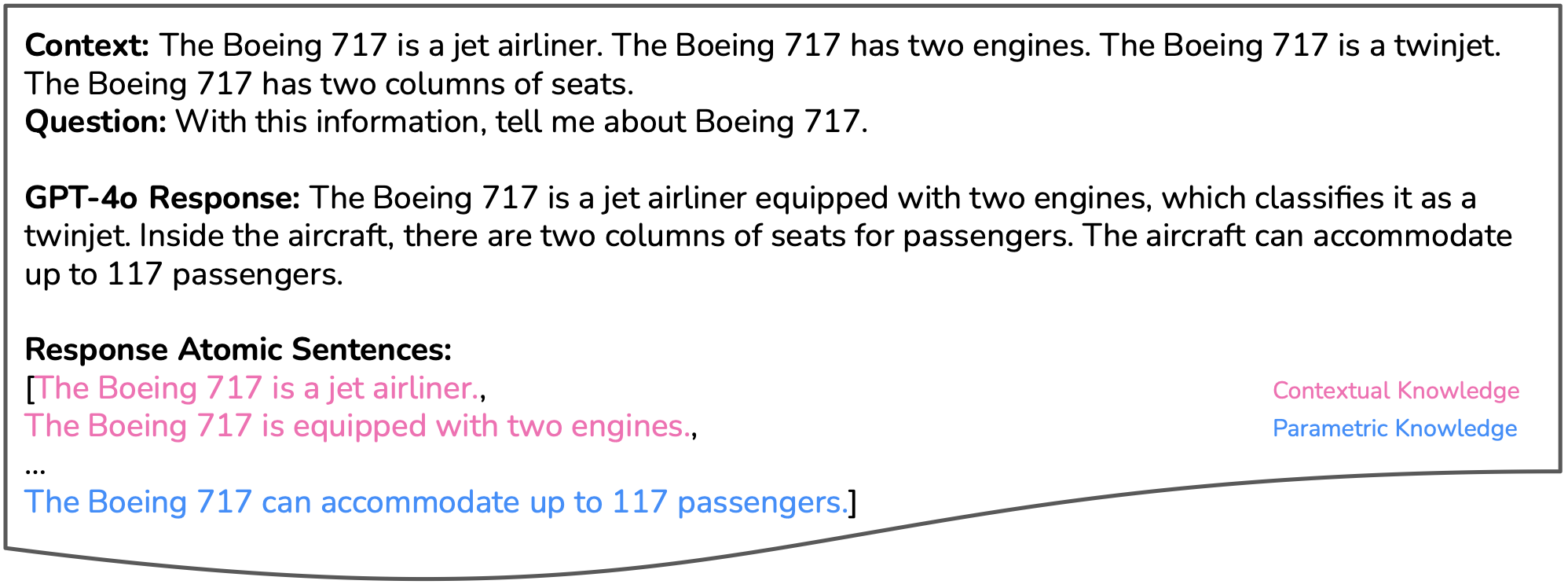}
    \caption{An example of Context, Question, Model Response (GPT-4o) and the list of Atomic Response mapped to  contextual knowledge and parametric knowledge}
    \label{fig:real_example}
\end{figure*}

\section{\texttt{WikiAtomic} Dataset}
%We study the behavior of LLMs while incrementally increasing amounts of context in an open-ended question answering task, focusing on the model's ability to prioritize different types of information and their tendency of factual hallucination in knowledge-consistent scenarios. In order to facilitate this investigation, 
We created a novel dataset -- \texttt{WikiAtomic} -- consisting of 200 articles from Wikipedia. We selected Wikipedia as the basis for creating our knowledge-consistent dataset  because its extensive  data has historically been integral to the training of numerous LLMs, explicitly in models like GPT-2, GPT-3, BERT, T5, and BLOOM, and implicitly in large-scale aggregate corpora such as The Pile \cite{gao2020pile} and Common Crawl \cite{luccioni2021whats}. This extensive use allows us to reasonably assume that information from Wikipedia, particularly from older articles, is present in the pretraining data of several LLMs we study. %Second, Wikipedia provides reasonably accurate information, establishing a solid baseline for hallucination detection.% in our study.

%which aggregate data from various sources, including Wikipedia. 

\medskip
\noindent \textbf{\em Extracting Wikipedia Articles} \quad We selected 200 high-quality articles from Wikipedia\footnote{\url{https://huggingface.com/datasets/wikipedia}}, each over 1000 words, covering diverse topics from science and technology to history, culture, and prominent figures. After manually removing low-quality texts, we ensured a diverse collection for our open-ended question answering task.

% We used Wikipedia Pageviews API to get each page's view count numbers over last 5 years. Previous study has shown LLMs struggle to learn long-tail knowledge \cite{kandpal2023large}, we want to explore would our selected models would respond differently when the subject is not popular. 

\medskip
\noindent \textbf{\em Converting to Atomic Sentences}\label{sec:convert-atomic} \quad 
To precisely control the number of contexts in our question, all articles are decomposed into a set of atomic sentences \cite{liu-etal-2023-revisiting}. Sentences containing  multiple pieces of information can complicate evaluations, particularly those involving entailment \cite{kim2024fables}. By breaking down sentences into atomic sentences, we ensure that each unit presents a single, unambiguous piece of information which enhances the accuracy of entailment assessments.  Following \citet{min-etal-2023-factscore}, we use GPT-4o to extract atomic sentences by providing the definition of atomic facts and instructing the model to perform multiple passes on an article to break each sentence down to individual atomic sentences (the prompt is included in Appendix~\ref{app:article_extraction}). Consequently, \emph{k} atomic sentences are extracted for each article (in our experiments, $k=50$) for a total of \emph{$200 * 50 = 10000$} atomic sentences in \texttt{WikiAtomic}. Example \ref{example:atomic_sentence_example} shows an original sentence and corresponding atomic sentences.

\vspace{-0.2cm}
\newcounter{example}
\setcounter{example}{0}
\renewcommand{\theexample}{\arabic{example}}

% \smallskip
\begin{tcolorbox}[fontupper=\small, colback=pink!10!white, colframe=pink!95!black, title=\textit{\textbf{Example \refstepcounter{example}\label{example:atomic_sentence_example}\theexample: Atomic Sentence}}, fonttitle=\sffamily]
\textbf{Original Sentence:} \\
\texttt{Mariah Carey, born on March 27, 1969, in Huntington, New York.} \\
\\
\textbf{Atomic Sentences:} \\
\texttt{1. Mariah Carey was born on March 27, 1969. \\
2. Mariah Carey was born in Huntington, New York.}
\end{tcolorbox}

We manually verify a subset of 1000 atomic sentences (20 articles, each with 50 atomic sentences) to ensure that the sentences (i)   were correctly atomized and, (ii) originate from the corresponding Wikipedia article. Only 12 out of 1000 sentences were found to be insufficiently atomic or had their meaning changed. {GPT-4o tended to struggle the most with sentences that have complex structures, such as those with multiple clauses or extensive use of commas. This led to errors in atomic sentence extraction where the sentences either contained too much information to be considered atomic or the meaning got altered in the process. Here is an example of each failure case: \medskip \\
\textbf{(a). Not atomic enough example:} \smallskip\\
\textbf{Original sentence:} \texttt{In 1774, the British passed the Intolerable Acts to punish the colonists in Boston for the Boston Tea Party.} \\
\textbf{GPT-4o’s atomic sentence:} \texttt{`In 1774, the British passed the Intolerable Acts to punish the colonists in Boston.',} … \\
\textbf{Correct atomic sentences:} \texttt{`In 1774, the British passed the Intolerable Acts.', `The Intolerable Acts were intended to punish the colonists in Boston.'} ... \medskip \\
\textbf{(b). Changed meaning example:} \smallskip\\
\textbf{Original sentence:} … \texttt{However, Denmark, on the losing side of the Napoleon wars, lost Norway to Sweden, on the winning side.} \\
\textbf{GPT 4o’s atomic sentences:} … \texttt{`Norway was on the losing side of the Napoleonic Wars.', `Sweden was on the winning side of the Napoleonic Wars.'}}
\medskip

Overall, GPT-4o achieved 98.8\% accuracy in atomic sentence extraction. The sentences that did not meet our criteria for atomicity were revised by referring back to their original Wikipedia articles.

% NEED TO REVISIT THINKING ABOUT WHERE TO PUT THESE CITATIONS!!!!!!!!!!!!!!!
% (revisit)To extract atomic facts from the LLM's response, we derive an approach from FactScore \cite{min-etal-2023-factscore} and MiniCheck \cite{minicheck}. We take each sentence in the response and 'atomize' it into individual pieces of context using GPT-4o. 

% We take this approach because each sentence from a source article may contain multiple facts. For example, the sentence "January is a cold month consisting of 31 days." contains multiple individual facts about January, and with our approach will be broken down into, "January is a cold month." and "January consists of 31 days." 

% Additionally, we instruct the model to include the proper noun in the split sentences. This helps to avoid producing pronouns such as, "\textit{It} consists of 31 days." (where \textit{It} refers to January in this example) which lacks semantic specificity and could reduce accuracy during evaluation.

% \begin{figure}
%     \centering
%     \includegraphics[width=1\linewidth]{fact_check.png}
%     \caption{Prompt used for determining local contexts. Using GPT4's JSON capabilities allowed us to easily work with the response and did not harm the accuracy. NOTE: I am going to make this look nicer, just wanted a placeholder}
%     \label{fig:entailment}
% \end{figure}

\section{Experiments and Evaluation}

This section outlines the experiment setup, evaluation metrics, and the models studied.

\subsection{Experiments}
 Figure~\ref{fig:real_example} illustrates a sample instance of our question answering task that includes the input (a list of atomic sentences as context along with a question prompt) and the output which is a model's response further atomized.

\noindent \textbf{\em Question Formulation} \quad We prompt the models to answer a question given the contexts using a {\em semi-restrict} format:  ``\texttt{With this information, tell me about \{Topic\}}''\footnote{When there is no context ($k=0$), the question is simply ``\texttt{Tell me about} \{{Topic}\}''.}. The effects of varying the question format are further explored in Section~\ref{sec:further}.

\begin{comment}
As shown in Example \ref{example:question_answer_format}, 
\medskip
\begin{tcolorbox}[colback=cyan!10!white, colframe=cyan!80!black, title=\textit{\textbf{Example  \refstepcounter{example}\label{example:question_answer_format}\theexample: Question Format}}, fonttitle=\sffamily]
\texttt{Contexts:} \{\textit{k atomic sentences about a topic}\}. \\
\texttt{Question: With this information, tell me about} \{\textit{Topic}\}.
\end{tcolorbox}
\medskip
\end{comment}

\noindent \textbf{\em Response Generation} \quad The responses generated by the LLMs are converted into atomic sentences using the same method described earlier in Section~\ref{sec:convert-atomic}\footnote{We used the same converting to atomic sentences extraction pipeline which was validated earlier.}. %This time, we extract all atomic sentences from the response without any . 

Ultimately, we obtain two lists for each question-answer pair:  atomic contexts (atomic sentences from the context)  and atomic responses (atomic sentences from the response), allowing us to directly compare them and minimize discrepancies.

\subsection{Evaluation}
Here, we describe the metrics for detecting contextual and parametric knowledge, as well as model hallucination, focusing on evaluating the faithfulness and factual accuracy of responses.

%\mee{a line why traditional QA quality (coherence, relevance, consistency using G-Eval) evaluation is unnecessary; instead we focus on faithfulness and factual accuracy via hallucination}

\noindent \textbf{\em Contextual vs. parametric knowledge detection} \quad 

When contextual and parametric knowledge largely align with subtle differences, it becomes challenging to distinguish between the two. When no context is provided, the model’s response serves as a baseline of global parametric knowledge. We calculate the overlap between the {\em response} (at $k=0$) and the longest {\em input context} ($k=50$) to estimate knowledge consistency. As shown in Figure~\ref{fig:topics_overlap}, parametric knowledge across five topics shows high consistency with \texttt{WikiAtomic}. This shows that the model can produce similar responses whether it is provided with detailed context or not, supporting our experiments in a non-conflict setting. This differentiates our work from previous studies that relied on counterfactual datasets.

%\textbf{Analysis 2} Consistency/diversity in parametric knowledge.

%\mee{if we have results for multiple runs for '0' global, upload here}

%See Figure~\ref{fig:5_0_global}
\begin{comment}
\begin{figure}
    \centering
    \includegraphics[width=\linewidth]{figures/5_0_global.png}
    \caption{Pairwise cosine similarities between SBERT embeddings of three responses to the same open-ended question (0 context) from GPT-4o for 5 random topics. \mee{so blue bar = sim(r1, r2), green bar = sim(r2, r3), red bar = sim(r3, r1)?}\eh{yes, but switch 2nd and 3rd bar. though this needs to be updated, do to a mistake. The new, correct one is already generated, just needs to be put in}} 
    \label{fig:5_0_global}
\end{figure}
\end{comment}

\begin{figure}[t!]
    \centering
\includegraphics[width=0.4\textwidth]{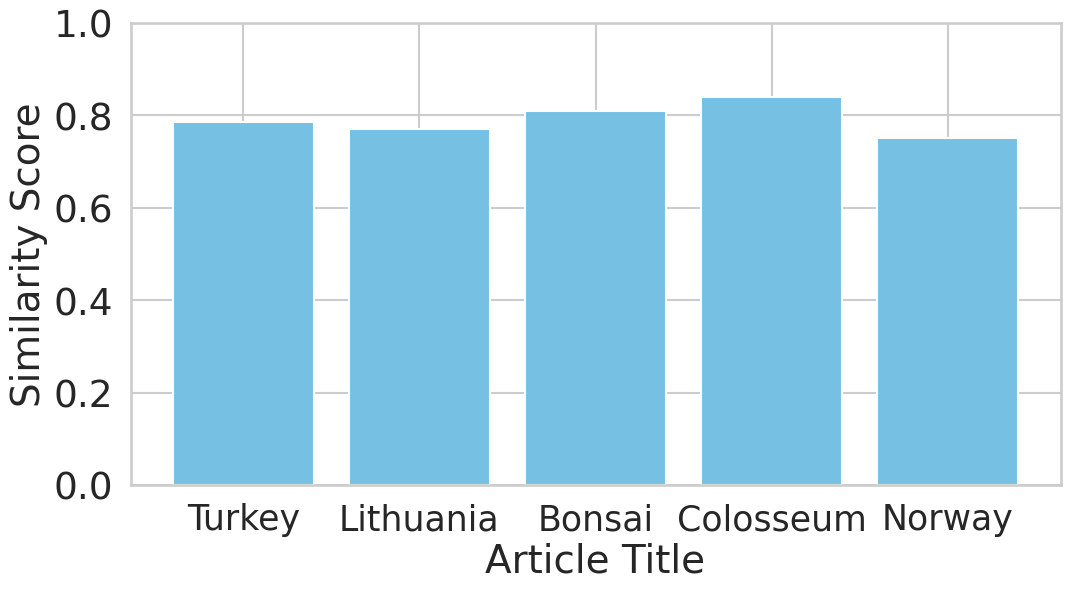}
    \caption{Knowledge-consistency between parametric knowledge and input context of \texttt{WikiAtomic} topics, computed using SBERT \cite{reimers2019sentencebert}}
\label{fig:topics_overlap}
\end{figure}

\begin{figure*}[t!]
    \centering
        \begin{subfigure}[t]{0.49\textwidth}
\includegraphics[width=\textwidth]{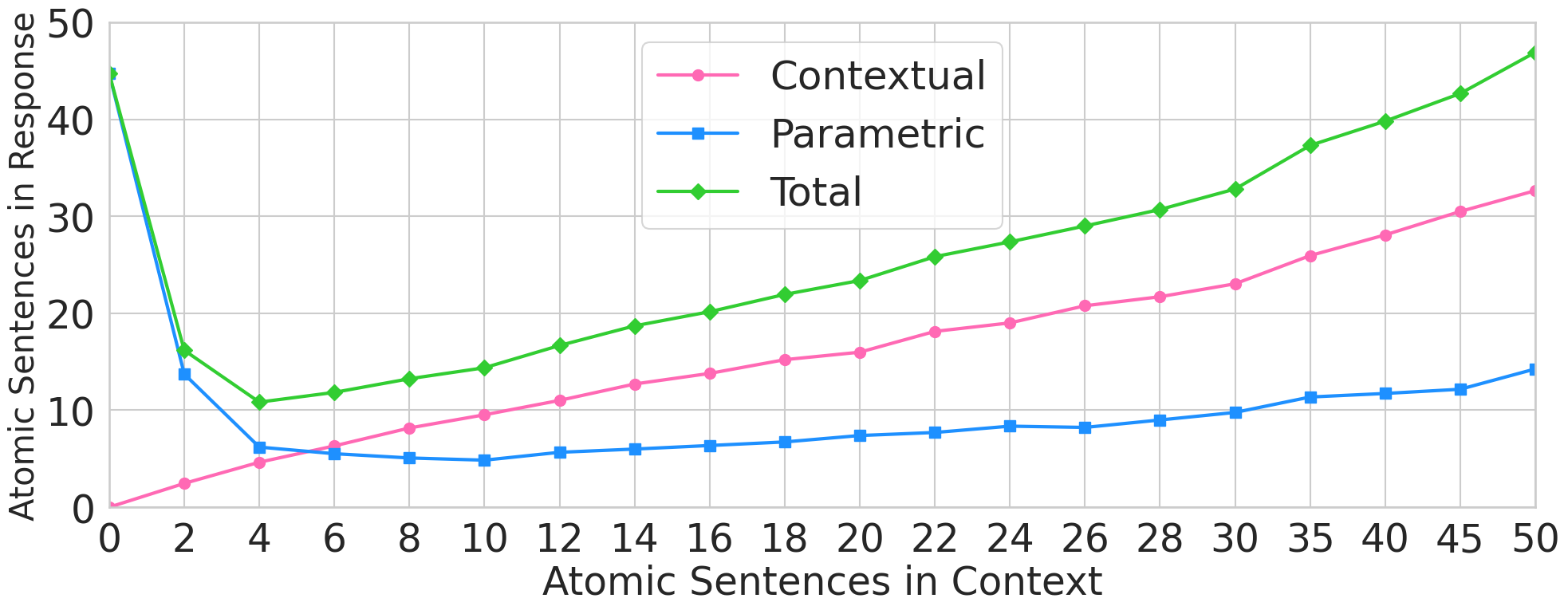}
        \caption{GPT-4o}
        \label{fig:line_gpt_4o}
    \end{subfigure}
        \hfill
    \begin{subfigure}[t]{0.49\textwidth}
\includegraphics[width=\textwidth]{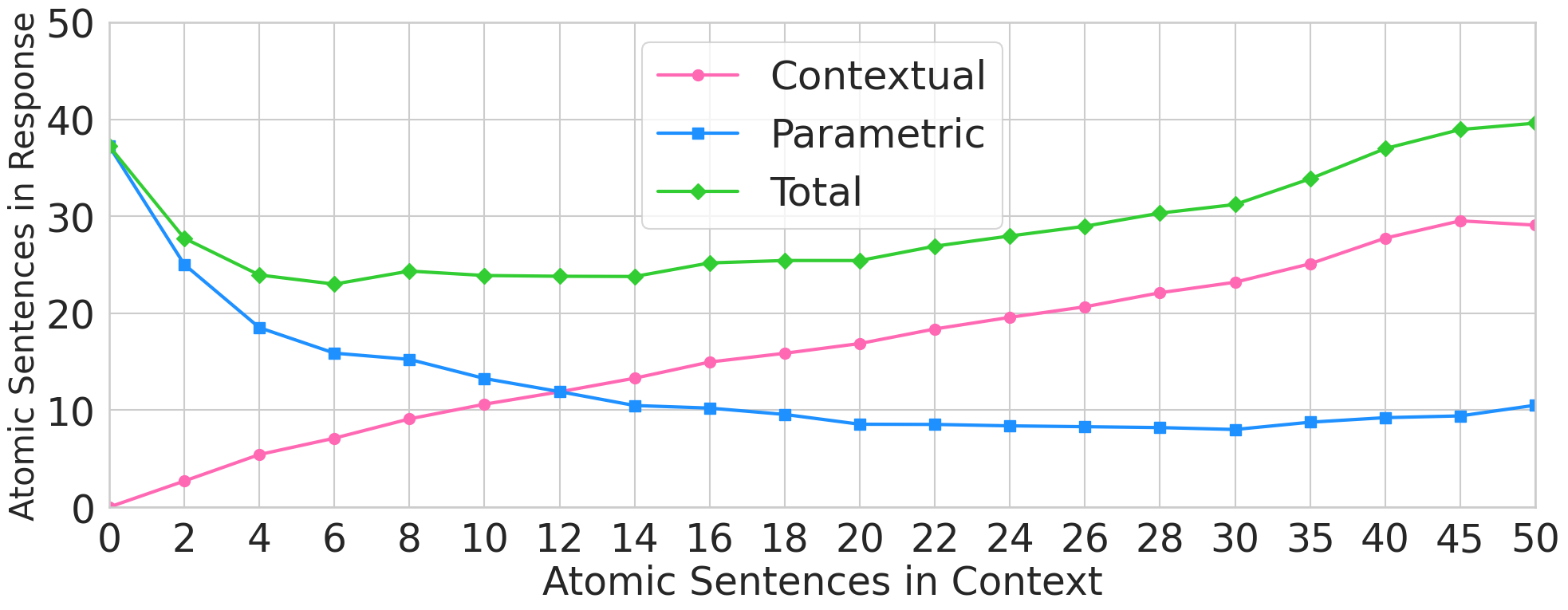}
        \caption{Claude Opus }
        \label{fig:line_opus}
    \end{subfigure}
    \hfill
    \begin{subfigure}[t]{0.49\textwidth}
\includegraphics[width=\textwidth]{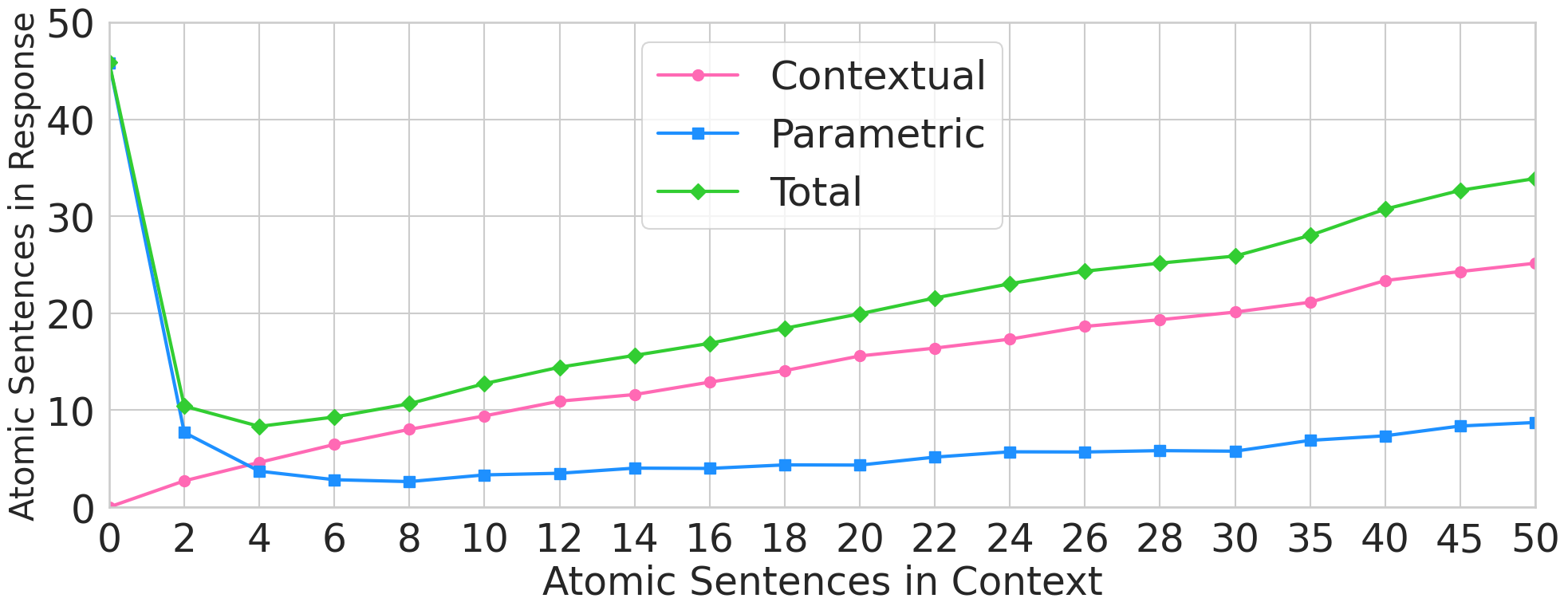}
        \caption{Llama 3 70B }
        \label{fig:llama3_70b}
    \end{subfigure}
    \hfill
    \begin{subfigure}[t]{0.49\textwidth}
\includegraphics[width=\textwidth]
{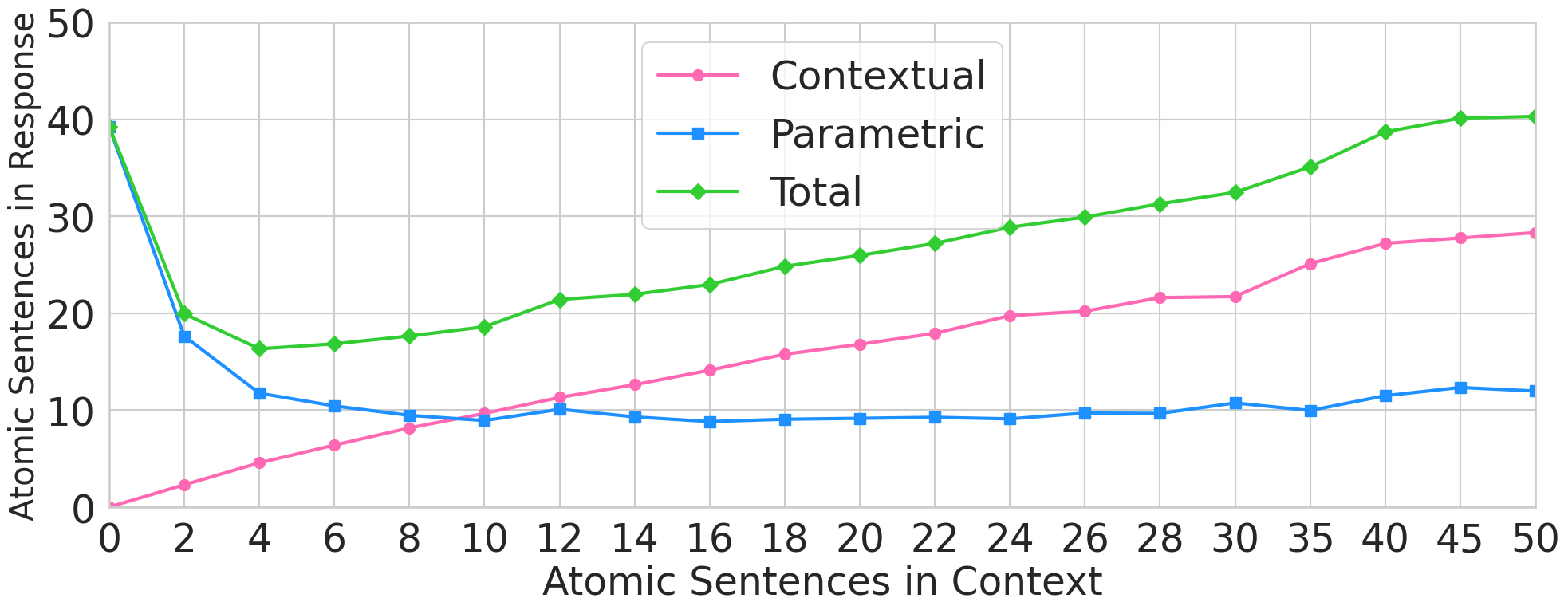}
    \caption{Mistral 8x22B }
        \label{fig:mistral7b}
    \end{subfigure}
    \caption{Contextual (local), parametric (global), and total sentences in responses for (a) GPT-4o, (b) Claude Opus, (c) Llama 3 70B, and (d) Mistral 8x22B. On the $x$-axis, $k=0$ serves as the baseline when no context is provided.}
    \label{fig:combined_lines}
\end{figure*}

Using the \textit{atomic contexts} as a reference, we categorize the \textit{atomic responses} as either contextual or parametric knowledge. We use the Natural Language Inference INFUSE framework\footnote{\url{https://github.com/HJZnlp/Infuse}}  \cite{zhang2024finegrained} to assess the faithfulness of responses. INFUSE calculates entailment scores for each sentence in what is considered the summary (responses from the model), with scores ranging from 0 to 1. A score of 0 indicates that the sentence is not entailed from the context, whereas a score of 1 signifies fully entailed from the context. We empirically set the threshold at 0.5 based on an examination of preliminary results, thus helping us identify sentences in the response as either contextual or parametric\footnote{Most sentences were clearly either entailed or not, with only about 10\% falling in the middle ambiguous range. We discuss the results of an ablation study later in Section~\ref{sec:ambiguous}.}. { This process involved three annotators who independently reviewed a subset of these ambiguously scored sentences. They assessed whether each of these sentences were contextual or parametric based on the definition and whether the scores align with their decisions. The threshold of 0.5 was chosen based on a consensus from this process, as it most effectively distinguished between the two categories.} Preliminary experiments with other metrics such as ROUGE, METEOR, and partial match yielded similar patterns (results included in Appendix \ref{app:alternative_methods_evaluation}).

\noindent \textbf{\em Hallucination detection} \quad For sentences classified as parametric knowledge, we further assess whether they are factually accurate or hallucinated as LLMs are known to hallucinate \cite{huang2023survey, xu2024hallucination, bai2024hallucination}. We use the FActScore framework\footnote{\url{https://github.com/shmsw25/FActScore}} \cite{min-etal-2023-factscore} that uses an external knowledge source to verify each sentence with scores ranging from 0 to 100; a higher score indicates fewer hallucinations and more factually accurate responses.

%(Wikipedia dump from 4/1/2023 by default) \ah{We don't use this dump, we use the same dump as the evaluation, I think 03/01/2022. Do we need to say which dump though? Also it only returns factscore, not a number of metrics}. It returns a number of metrics, including a `FactScore` which constitutes our hallucination metric. 
% We will also use SelfCheckGPT\cite{manakul-etal-2023-selfcheckgpt}. \yt{will add more when we start this part}.

\subsection{Models}
We study nine models ranging from small models to state-of-the-art LLMs, including both open-source and closed-source options (for implementation details, see Appendix~\ref{app:implementation_detail}). These models include: {\bf GPT-4o} (\texttt{gpt-4o-2024-05-13}), {\bf Claude 3 Opus} (\texttt{claude-3-opus-20240229}), {\bf Sonnet} (\texttt{claude-3-sonnet-20240229}) and {\bf Haiku}  (\texttt{clau\\ de-3-haiku-20240307})), {\bf Llama 3}\footnote{\url{https://ai.meta.com/blog/meta-llama-3/}} {\bf 70B} (\texttt{Meta-\\Llama-3-70B-Instruct}) and {\bf 8B} (\texttt{Meta-Llama-3\\-8B-Instruct}), {\bf Mixtral} 8x22b (\texttt{Mixtral-8x22B\\-Instruct-v0.1}), {\bf Mistral 7B} (\texttt{Mistral-7B-Inst\\ruct-v0.2}) \cite{jiang2023mistral}, and  {\bf Phi-3}\footnote{\url{https://huggingface.co/microsoft/Phi-3-mini-4k-instruct-gguf}} (\texttt{Phi-3-mini-4k-instruct-gguf}) \cite{abdin2024phi3}.

\section{Results and Analysis}
Now, we present and discuss the results of our experiments.

\definecolor{linebg}{HTML}{7AA2E3}
\definecolor{headerbg}{HTML}{FFCBCB} % Hex code FFF9D0
\definecolor{headertext}{HTML}{474E68}
\definecolor{rowtext}{HTML}{2F86A6}

\begin{table*}[t!]
    \centering
    \small
    \begin{tabularx}{0.9\textwidth}{lXXXXXX}
        % \rowcolor{headerbg}
        \toprule
        \textbf{{Model}} & \textbf{{$k=10$ }} & \textbf{{$k=20$}} & \textbf{{$k=30$ }} & \textbf{{$k=40$}} & \textbf{{$k=50$}} \\
        \midrule
        GPT-4o & 0.69 / 0.31 & 0.68 / 0.32 & 0.69 / 0.31 & 0.68 / 0.32 & 0.67 / 0.33 \\
         Claude 3 Opus & 0.48 / 0.52 & 0.67 / 0.33 & 0.73 / 0.27 & 0.74 / 0.26 & 0.72 / 0.28 \\
         Claude 3 Sonnet & 0.50 / 0.50 & 0.64 / 0.36 & 0.68 / 0.32 & 0.67 / 0.33 & 0.66 / 0.34 \\
        Claude 3 Haiku & 0.69 / 0.31 & 0.75 / 0.25 & 0.75 / 0.25 & 0.71 / 0.29 & 0.72 / 0.28 \\
           Llama 3 70B & 0.73 / 0.27 & 0.75 / 0.25 & 0.74 / 0.26 & 0.74 / 0.26 & 0.72 / 0.28 \\
           Llama 3 8B & 0.62 / 0.38 & 0.63 / 0.37 & 0.65 / 0.35 & 0.65 / 0.35 & 0.67 / 0.33 \\
         Mixtral 8x22B & 0.58 / 0.42 & 0.65 / 0.35 & 0.66 / 0.34 & 0.69 / 0.31 & 0.69 / 0.31 \\
          Mistral 7B & 0.48 / 0.52 & 0.61 / 0.39 & 0.66 / 0.34 & 0.67 / 0.33 & 0.69 / 0.31 \\
        Phi-3 & 0.24 / 0.76 & 0.37 / 0.63 & 0.42 / 0.58 & 0.46 / 0.54 & 0.49 / 0.51 \\
        \midrule
        \textbf{{All Models}} & 0.56 / 0.44 & 0.64 / 0.36 & 0.67 / 0.33 & 0.67 / 0.33 & 0.67 / 0.33 \\
       % \textbf{All Models except Phi} & 0.60 / 0.40 & 0.67 / 0.33 & 0.70 / 0.30 & 0.69 / 0.31 & 0.69 / 0.31 \\
        \bottomrule
    \end{tabularx}
    \caption{Ratios of contextual/parametric knowledge in responses}
    \label{tab:ratios}\vspace{-0.4cm}
\end{table*}

\subsection{In knowledge-consistent setting, how do models prioritize sources of knowledge?}
\label{main_plots}

% \ah{Let's say something about it dropping from the baseline. It definitely drops, but that is probably because we are basically telling it to say whatever it knows about the topic rather than guiding it. And it's expected that there would be many more global facts since there are 0 local facts. Also, I think it's very interesting that almost all the models (except Phi) show pretty much the exact same pattern. We should touch on that here.}
Figure~\ref{fig:combined_lines} shows how four models prioritize sources of knowledge (similar results were obtained from other models, plots included in Appendix \ref{app:rest_of_nli}). The plots display changes in the average number of local, global, and total atomic sentences in responses as context increases. Somewhat surprisingly, we found that all models (except Phi-3 in certain cases) show consistently similar patterns. This uniformity among the models suggests a shared underlying mechanism in processing and responding to contextual information. %, while Phi-3's divergence suggests areas for further investigation and potential improvement.

With no context provided ($k=0$), models tend to be  most verbose as expected from an initially open prompt. They also peak in parametric knowledge, which drops drastically with the first four contexts. Interestingly, total response lengths generally match context lengths. From $k=2$ onward, the total sentences and local contextual knowledge steadily increase as context increases. {More contexts were utilized but none of these models utilized 100\% of them in their responses.} Global facts show a consistent trend: after the initial open-ended prompt, the global knowledge drops sharply and remains low, though never to zero, with some models slowly increasing or decreasing but always including some global knowledge in their responses.

%Before reaching $k=20$, the number of contexts nearly matches in responses, slowing down to a maximum of 38 contextual sentences after 45 contexts.

% The global facts line also have some interesting behavior. First, when there are less than 4 contexts in questions, GPT-4o first provide the most number of global facts and drastically reduce after. Second, the number of global facts almost always hovers around 5-10 after the number of contexts in a question exceeds 10, indicating the tendency of providing a small amount of global facts regardless of further increases in context. Even sometimes it has enough contexts from questions to generate quality responses. 

Table~\ref{tab:ratios} presents the  proportions of contextual and parametric knowledge in responses. Larger contexts generally increase the model's reliance on contextual knowledge (about 70\%) while reducing dependence on parametric knowledge (about 30\%). However, in smaller contexts, models show different preferences, with some prioritizing contextual knowledge (GPT-4o, Claude Haiku, Llama 3 70B, Llama 3 8B, and Mixtral 8x22B), and others, parametric knowledge (Claude Opus, Mistral 7B, Phi-3). {Moreover, the average proportion remains similar for $k$ = 30, 40, and 50 suggesting that the ratio of contextual/parametric knowledge is maintained at these different context lengths.}

%Interestingly, smaller models (Mistral 7B and Llama 7B) draw {\em more} up on parametric knowledge, and larger models generate responses more faithfully aligned with the context.

%The x-axis represents the number of atomic sentences in the context, while the 
%The y-axis shows the percentage of contexts recalled, and different lines correspond to different quartiles of context.

\subsection{Which parts of context are used?}

The results of the previous experiment intrigued us, leading us to the next question: which parts of the provided context do LLMs use in their responses? Using the INFUSE framework, we reversed the input position to give each atomic sentence in the context an entailment score, considering sentences with scores greater than 0.5 to be included in the model's response. For $k>4$, inputs were split into quartiles for clearer analysis.

\begin{figure}[t!]
    \centering
    \includegraphics[width=0.47\textwidth]{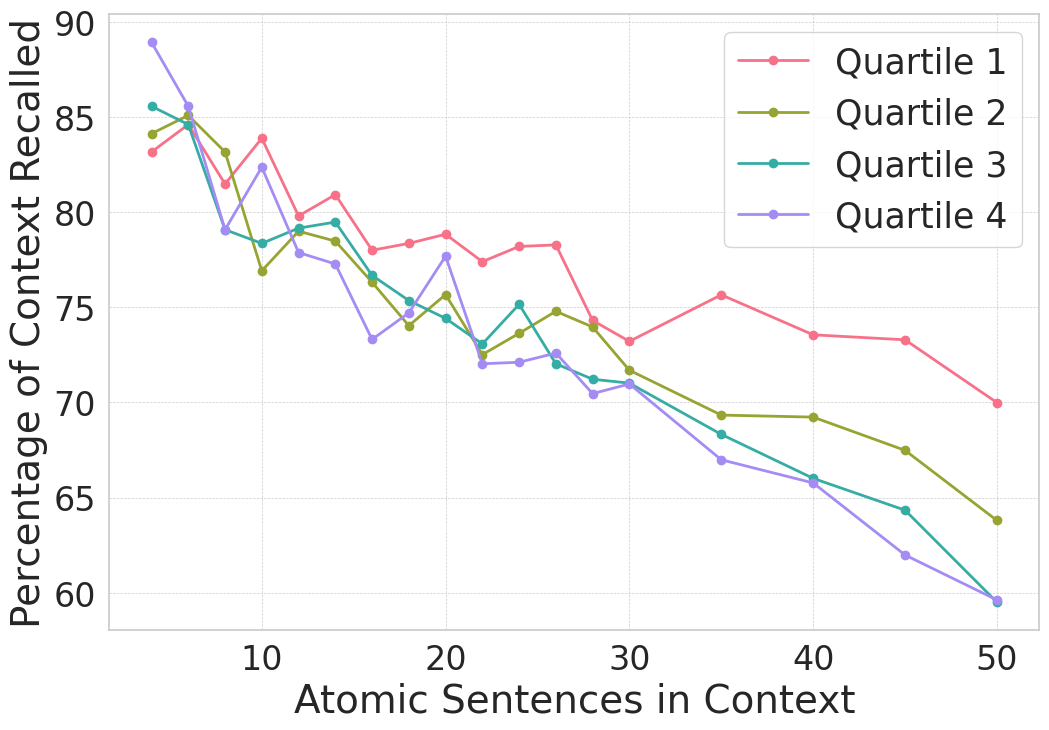}
    \caption{Percentage of each quartile of context recalled in response (GPT-4o)}
    \label{fig:context_positions}
    \vspace{-0.3cm}
\end{figure}

%not consider cases where the number of contexts is lower than 4. 

%Once we have the nested list of scores, we group the lists inside the nested list into 20 groups so we can look at the general pattern of what areas of contexts exists in responses across 200 example sets. 

%The ``lost in the middle'' phenomenon  highlights challenges LLMs face in recalling middle information from long contexts, typically in short response task. 
Figure~\ref{fig:context_positions} (additional results in Appendix \ref{app:position_analysis}) shows that for smaller contexts ($k < 10$), the model treats all portions of the context equally. As context increases, the model predominantly focuses on the first quartile. For $k \geq 25$, the preference gap between quartile 1 and all others further increases, with quartile 2 becoming the next most preferred section. We observe selective attention where certain parts of the data receive more focus than others \cite{liu2024lost}, and find that models struggle more with recalling the bottom half of the data rather than the middle, further confirming the importance of initial contexts. 

%Both findings highlight inherent biases in how models process long data.

% Specifically, with 50 contexts, the first quartile influences 70\% of the model’s responses, the second quartile 65\%, and the third and fourth quartiles each influence 50\%.
%The rest of models also show a decreasing preference for later quartiles and earlier contexts tend to be more likely to be utilized in the responses as the context size expands (See Appendix \ref{app:position_analysis} for each model's graph). This pattern highlights the increasing importance of the initial contexts as more are added.

% \begin{figure}
%     \centering
%     \includegraphics[width=0.49\textwidth]{figures/local_global_similarity/combine_lvg.png}
%     \caption{Similarity between contextual knowledge and corresponding parametric knowledge in responses}
%     \label{fig:combine_lvg}
% \end{figure}

% \subsection{How do contexts map to responses?}

% Seeing that the models tend to prefer earlier contexts, we wonder where those selected contexts end up in the responses.

% On top of looking at how many contexts from questions are used in the model responses. We also investigated where are the contexts ended up in responses.
% We also investigated are there specific areas on contexts in questions LLMs prefer to use in their responses. 
To determine where provided contexts are  used in the responses, we divide contexts and responses into four quartiles and compare each section using cosine similarity of SBERT embeddings \cite{reimers2019sentencebert}, essentially mapping the context quartiles to response quartiles. From Figure~\ref{fig:position}, we observe that the first context quartile maps most closely to the first, and to a lesser extent, the adjacent response quartile  (additional results in Appendix \ref{app:where_contexts_go}). This pattern continues with each subsequent context quartile showing the  highest similarity to its corresponding response quartile, indicating that the model prefers to match the positions of data. This pattern likely mimics the natural sequence  of text when describing a topic, aligning with how information is typically structured. 

\begin{figure}[t!]
    \centering
    \includegraphics[width=0.49\textwidth]{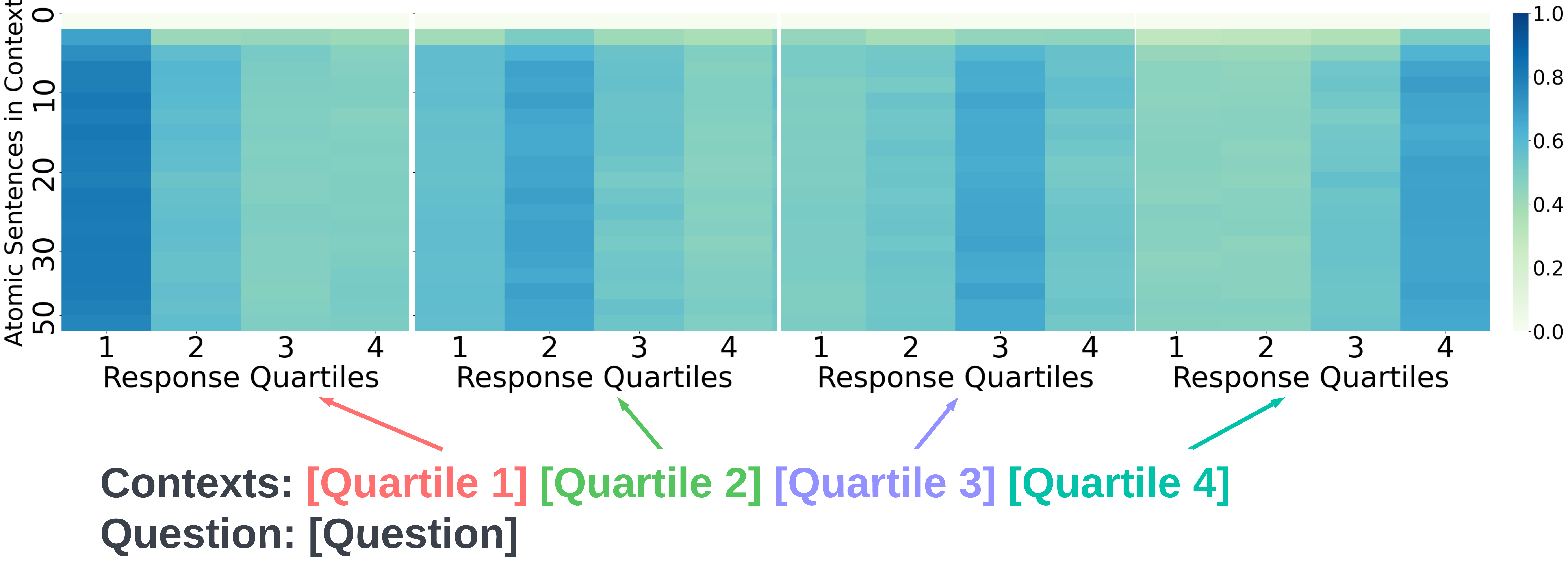}
    \caption{Mapping context quartiles to response quartiles (GPT-4o)}
    \label{fig:position} 
    \vspace{-0.5cm}
\end{figure}

%\footnote{We also experimented with ROUGE score and TF-IDF to compute similarity but they faced the same challenge that couldn't capture precise semantic meaning, plus ROUGE is too sensitive to length of texts. } 
%We decided to use SBERT over TF-IDF and ROUGE because SBERT captures deeper semantic relationships. TF-IDF can only measure term frequency and importance, and it doesn’t account for the context or meaning behind words. ROUGE primarily measures surface-level n-gram overlaps and is too sensitive to length, which is a drawback for us since the length of context sections can be very short compared to the model responses. ROUGE wouldn’t provide an accurate measure of similarity in our case.

% which contexts were used and what position of the contexts were used. 

% We randomly sampled \yt{20} set of examples, and take average of scores. 

% \ah{I think we need to rerun these graphs. You can't create quintiles for integers less than 5 so 0-> 4 contexts is incorrect. We should use quartiles I think so we can run this and start the plot at 4 contexts like in the position graph.} \yt{we didn't split by number of contexts, we did by words}

%\mee{remove? }Second, the heatmaps show when the first, second and last quintile of contexts are more likely to be used in model responses according the intensity of those heat maps.

\subsection{How similar are various types of knowledge?}

Our analysis so far suggests that LLMs consistently add parametric knowledge regardless of the amount of context provided in the question. This prompted us to investigate the relationships between the different types of knowledge. {The graphs were generated as follows: For each Wikipedia topic containing 20 questions (up to 50 contexts), we obtained 20 responses. Each response contains sentences marked as contextual or parametric knowledge.}

%Recall from Figure \ref{fig:combined_lines} that even though the average of added number of parametric knowledge stays around 10 or lower regardless of how many contexts were provided in questions. while the number of global atomic claims are relatively stable,

\medskip
\noindent \textbf{Local vs. Local} \quad We perform  a pairwise comparison of {\bf contextual knowledge} in responses across various context sizes (Figure~\ref{fig:gpt-local-local},  full set of results in Appendix \ref{app:local_vs_local_simialrity}). We observe that the local context remains moderately similar in smaller contexts but becomes increasingly similar with larger contexts. This pattern suggests that models tend to focus on certain types of context, often the earlier parts.

%\mee{response vs. response} At the same time, the heatmap in Figure~\ref{fig:lvg_sonnet} shows high similarity scores across all of Claude Sonnet model's responses. (We only show Sonnet's result because results from rest of the models are all very similar. To see more detailed graph for each LLM, see Appendix \ref{app:local_vs_local_simialrity}.) \ah{Should we make this statement definitive or provide it as a hypothesis?} This shows these models were providing supplemental information to existing contextual knowledge when incorporating parametric knowledge instead of brand-new information. 

\begin{figure}[t!]
    \centering
    \begin{subfigure}[t]{0.31\textwidth}
\raisebox{-0.5\height}{\includegraphics[width=\textwidth]{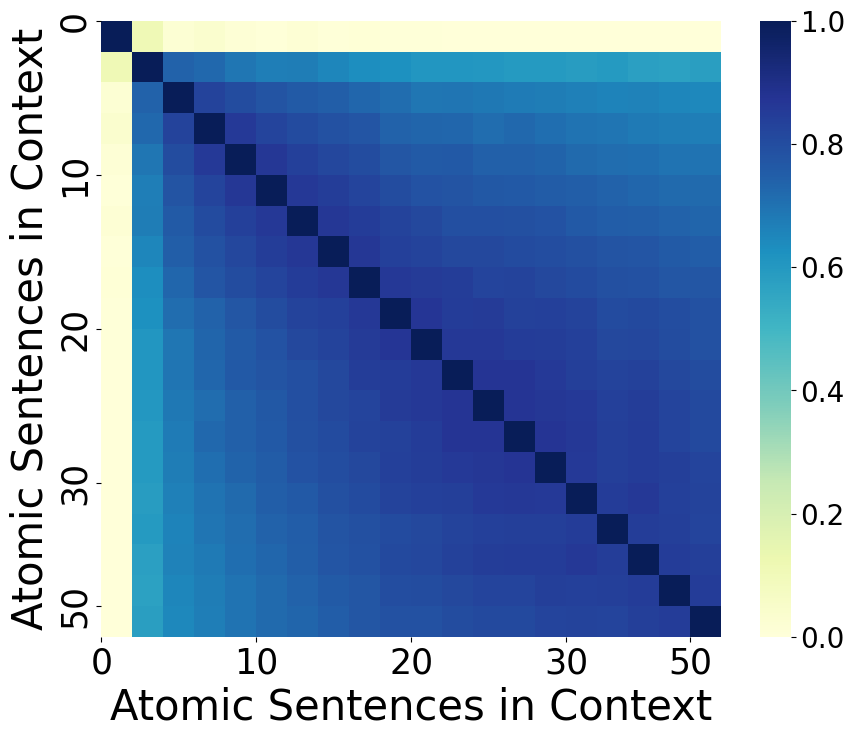}}
    \caption{Local vs. local (GPT-4o)}
    \label{fig:gpt-local-local}
\end{subfigure}
 \hfill
    \begin{subfigure}[t]{0.31\textwidth}
\raisebox{-0.5\height}{\includegraphics[width=\textwidth]{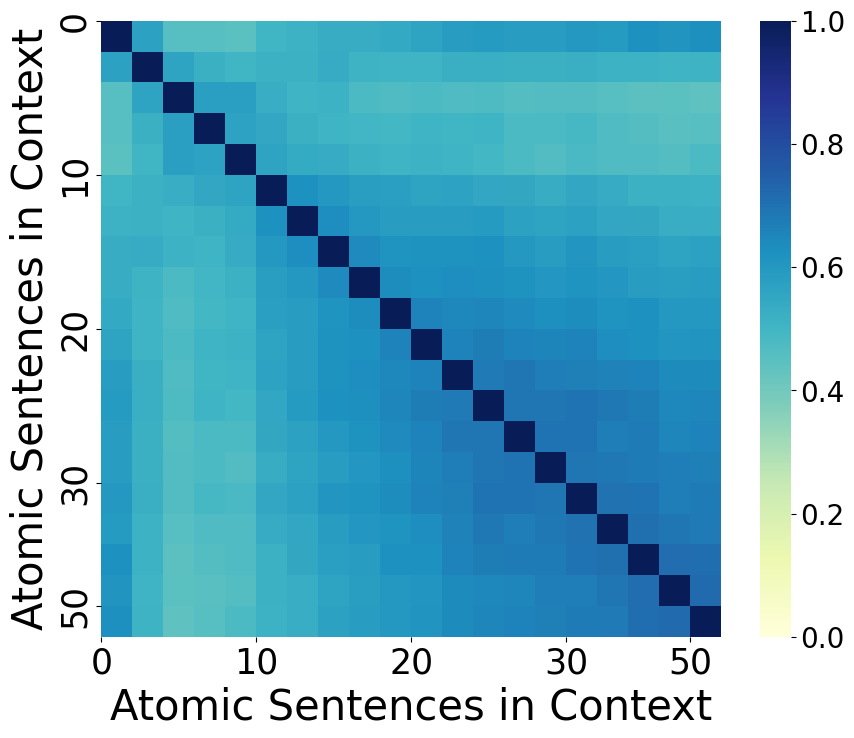}}
    \caption{Global vs. global (GPT-4o)}
    \label{fig:gpt4o_gvg_main}
    \end{subfigure}
    \hfill
    \begin{subfigure}[t]{0.35\textwidth}
{\includegraphics[width=\textwidth]{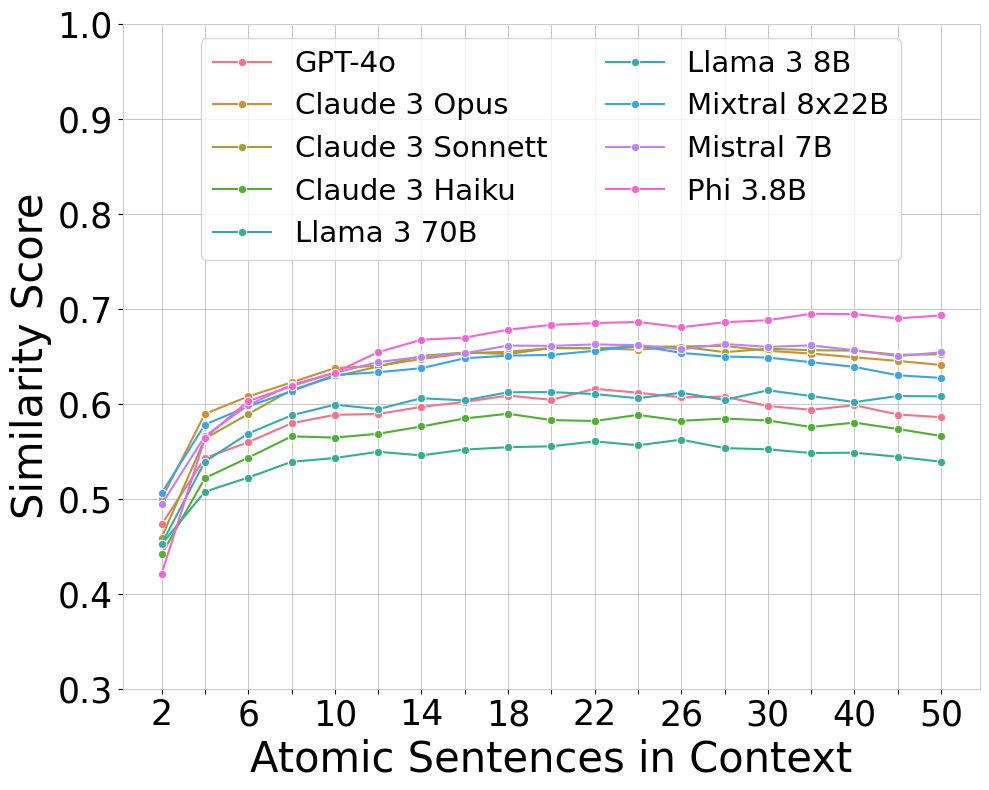}}
    \caption{Local vs. global (All models)}
    \label{fig:combine_lvg}
    \end{subfigure}
    \caption{Similarity between contextual knowledge of responses obtained with varying input lengths (a), similarity between parametric knowledge of responses obtained with varying input lengths (b), and similarity between contextual and parametric knowledge (c).}
\label{fig:global_global_similarity}
\vspace{-0.5cm}
\end{figure}

\medskip
\noindent \textbf{Global vs. Global} \quad Let us now turn to Figure~\ref{fig:gpt4o_gvg_main} (full set of results in Appendix \ref{app:global_global_similarity}) which shows pairwise similarity between {\bf parametric knowledge} in each response. The darker plots near  the diagonal line indicate that the model's parametric knowledge remains similar when context sizes are close, such as context size 8's parametric knowledge being most similar to that of sizes 6 or 10. Initially, with minimal context, models provide a larger variety of information. As context size increases, some models show increased similarity across broader ranges, suggesting responses align more closely with the overall theme of the context. This indicates that models tend to repeatedly add certain pieces of information from a small pool of knowledge, though it is not as uniformly homogeneous as local knowledge.

\medskip
\noindent \textbf{Local vs. Global} \quad Our analysis naturally leads us to our next question: what type of global parametric knowledge is incorporated with local contextual knowledge? Our assumption is that low similarity indicates complementary  parametric knowledge, while high similarity suggests it is supplemental. Figure~\ref{fig:combine_lvg} (detailed graphs in Appendix \ref{app:local_global_similarity}) shows consistent trends across all models. For smaller contexts, the similarity between {\bf contextual knowledge and parametric knowledge} is much lower, suggesting that models incorporate complementary information not directly provided by the context. As context increases, the similarity rises to around 0.6 indicating moderate similarity, with global knowledge sharing some concepts with the provided context, but not merely echoing it. %However, as context increases, so does the similarity between local and global knowledge.

% \ah{I think we should find a good paper to reference for using SBERT for this kind of use case. Maybe it's just the SBERT paper but I think we need to justify our decision. We discussed this in the meeting but we should just look at the local -> global example within a specific sample. It doesn't make sense to compare the local context of 2->10 to the global context of 30->50. We should calculate a similarity score between each local and global set for a specific context length and base our analysis on that.} \yt{Forgot to share about SBERT, I talked to Ameeta a bit about which to measure similarity. the tfid(forgot the name) cosine similarity or rouge, the first one is too weak and can’t capture much, rouge is too sensitive to length which in our case matters a lot. SBERT is overall better. I wrote “we utilized SBERT again…” because I was going to write about this at the first place where we were going to use it. But I FORGOT :) I’ll add that later and yes I’ll look for some references.}

%We utilized SBERT again to compute the similarity score between each response's contextual knowledge and parametric knowledge. We then group the score by the number of atomic sentences in the context and compute the mean for each case. 

% In this specific study, we choose to use cosine similarity score instead of Rouge score because the length of local atomic sentences varies a lot compare to global atomic sentences and Rouge score is sensitive to the text length. 

\subsection{In knowledge-consistent setting, (how much) do models hallucinate?}

In Figure~\ref{fig:factscores}, we present the FActScore hallucination scores across all models. For smaller contexts, models have higher hallucination rate, which improves with additional context and converges with as little as 10 sentences in context. Larger models (GPT-4o, Claude Opus, Claude Sonnet, Claude Haiku, Llama 3 70B, and Mistral 8x22B) consistently show higher FActScores, indicating they are less prone to hallucinations and benefit significantly from additional context. Among  smaller models, Mistral 7B and Llama 3 8B perform well, showing significant improvement with increased context, while Phi-3 tends to hallucinate more, although it also improves with context. Larger models generally stabilize at higher FActScores with fewer fluctuations, indicating a stronger ability to leverage additional context to minimize hallucinations. Smaller models show a more gradual improvement, with some like Phi-3, displaying more variability and a lower overall FActScore, suggesting they are more context-sensitive but can still improve with more information (for detailed results of false parametric knowledge for each model, see Appendix~\ref{app:false_claims}).

\subsection{Further Analyses}\label{sec:further}
%We discuss our results from three further analyses.%: (potentially) unseen knowledge and prompt sensitivity. 

\noindent \textbf{(Potentially) Unseen Knowledge} \quad  
In knowledge-consistent scenarios, models rely on a mix of contextual and parametric knowledge. But in unseen knowledge scenarios, where the context contains new information that the models have potentially not seen, how do models respond? To explore this, we create a recent example around the pro-Palestinian university protests of April 2024 and prompt the models for information on this topic.

\begin{figure}[t!]
    \centering
    \includegraphics[width=0.49\textwidth]{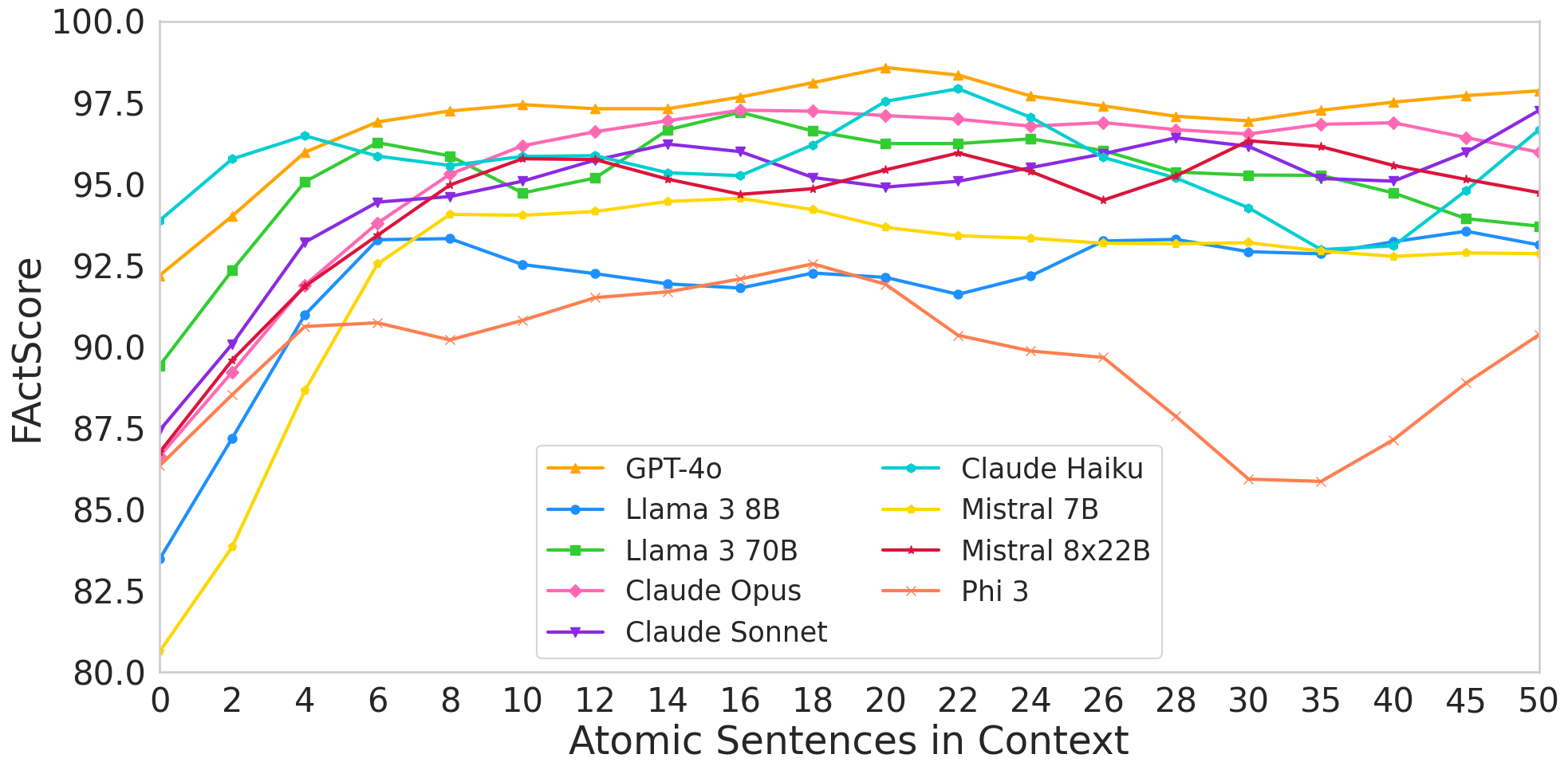}
    \caption{FActScore across all LLMs (higher score indicates lower hallucination)}
    \label{fig:factscores}
\end{figure}

\begin{figure}[t!]
    \centering
    \includegraphics[width=0.49\textwidth]{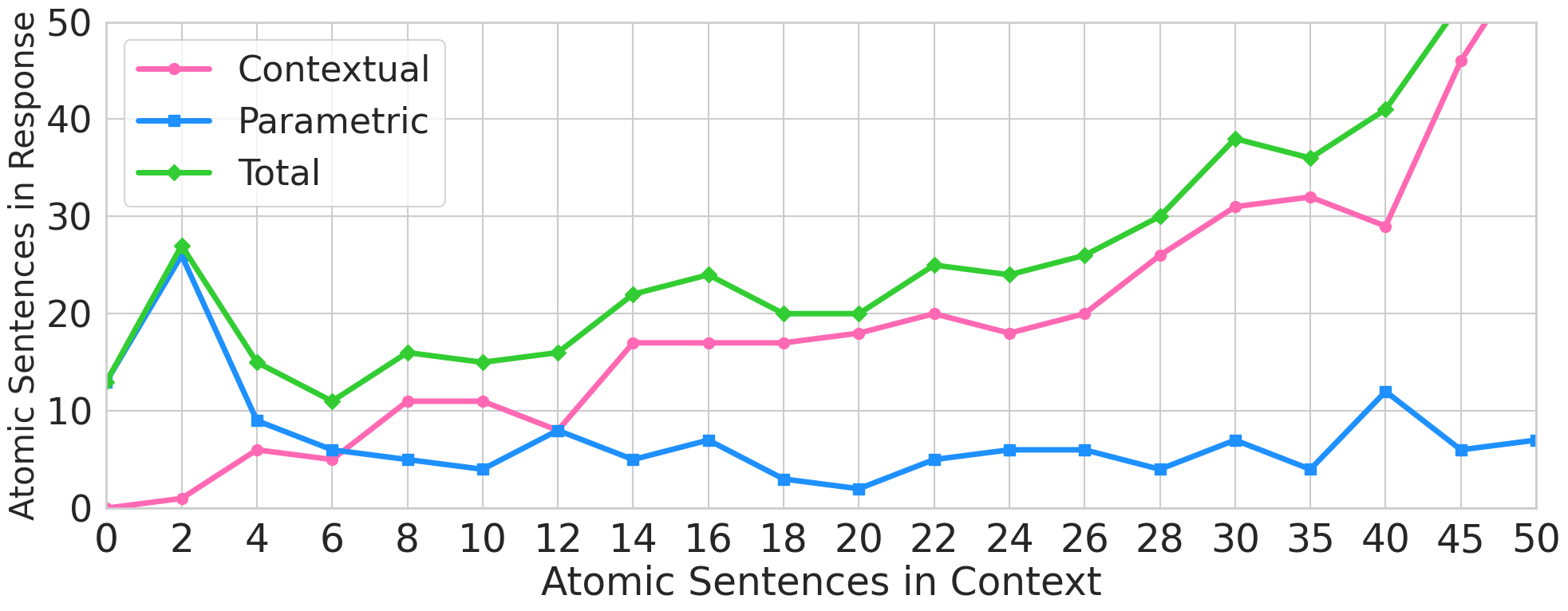}
    \caption{Unseen knowledge results (GPT-4o)}
    \label{fig:new_gpt4o_result}
    \vspace{-0.5cm}
\end{figure}

Without context, most models stated they lacked information on the topic and provided a general background on protests related to Palestine {(see Figure~\ref{fig:new_gpt4o_result}, with additional results in Appendix~\ref{app:new_knowledge}). However, some models, Claude Haiku, Llama 8B, and Mixtral 22x8B, all confidently respond to the prompt as if they have seen this knowledge in their pre-training data. We hypothesize that this may be due to similar events that have occurred in the past that the model has seen in its pretraining data.}
With just two contexts, all models referenced the context but some included facts not in the context. With 50 contexts, they effectively incorporated the contextual knowledge without much additional information. Our findings suggest that care must be taken when querying LLMs on new information as  without context or with minimal context, models may still confidently provide seemingly accurate information not in their parametric knowledge. %This issue is mitigated by providing sufficient contextual information, as shown by the high ratio of contextual knowledge in responses.

\medskip
\noindent \textbf{Prompt Sensitivity} \quad 
Our naturalistic {\em semi-restrict} question prompt (“With this information, tell me about [topic].”) encourages the models to consider the provided contexts while still allowing them to use their parametric knowledge. Considering that LLMs are sensitive to prompts \cite{lu2021fantastically}, we tested two alternative phrasings: a {\em no-restrict} prompt which allows the  models complete freedom to draw on their parametric and contextual knowledge as they see fit (“Tell me about [topic]”) and a {\em  strict} prompt which tries to restrict the models to use only the contexts provided (“Using the provided context only, tell me about [topic].”).

To conduct this ablation study, we randomly selected 20 topics from our dataset, resulting in a total of 400 instances of varying context sizes, and obtained responses different prompts. While the {\em no-restrict} plots look similar to the {\em semi-restrict} prompts presented earlier, the {\em strict} plots show that the models are entirely focused on contextual knowledge with very little parametric knowledge (supporting graphs in Appendix \ref{app:ablation}). These results confirm model sensitivity to prompts and highlight that, with simple prompt adjustments, models can be guided to leverage different ratios of contextual and parametric knowledge. 

\medskip
{\noindent \textbf{Disregard Ambiguous Sentences} \label{sec:ambiguous} \quad
When we used INFUSE to generate entailment scores, a small number of sentences had scores close to 0.5, indicating lower confidence in categorizing them as entailed or not. We decided to exclude these sentences to assess whether the patterns from Section~\ref{main_plots} remained valid. After ignoring sentences with scores between 0.3 and 0.7 (see Appendix~\ref{app:disregard_infuse}), the patterns persisted, confirming the robustness of our approach and the reliability of the INFUSE framework for this task.}

% Phi being the smallest model and has the smallest context window size of 4k definitely has impact on how it utilizes contextual knowledge and parametric knowledge. \yt{but we can't say it's because of context window? what if it's model size?}

\section{Discussion}
%\mee{discussion section summarizes main findings but then also focuses on what it means, their implications - so a call to action. what do we need to do next? develop models that can better leverage contextual? seamlessly integrate diff. sources of knowledge? focus on latter contexts? etc.}

Understanding how LLMs utilize contextual and parametric knowledge is important for real-world knowledge-consistent use cases. We summarize our key findings and their implications: 

\begin{itemize}
    \item {Models process context very similarly} suggesting a standardized approach to context processing across all selected models.

    \item {Context is never fully utilized and some parametric knowledge is always included}  highlighting the need for developing models that utilize input contexts more deterministically.

    \item {Earlier information is prioritized in longer contexts, with responses following the order of presented information}  highlighting the importance of organizing context effectively.

    \item {Responses tend to include similar contextual knowledge but supplemental parametric knowledge.}  

    \item {Hallucinations decrease as context increases } implying that more context helps models generate more accurate responses, reducing reliance on potentially incorrect parametric knowledge.

\end{itemize}

\section{Related Work}

\textbf{Contextual Grounding} \quad 
%This area examines how LLMs integrate local contextual information with the global knowledge embedded in their parameters. 
A persistent challenge for LLMs lies in reconciling contradictory or superseded information between the provided context and their internal knowledge base \cite{li2022large,zhou2023contextfaithful}, with most recent work exploring this interplay between local and global knowledge in knowledge-conflict scenarios using counterfactual datasets \cite{si2022prompting, qian2023mergeconflictsexploringimpacts, feng2024knowledge, Yang-GiveFacts}. The preference for contextual or parametric knowledge in counterfactual settings is not straightforward. Larger models seem better at adapting to counterfactual context, while smaller ones often prioritize their learned knowledge \cite{si2022prompting}. When faced with counterfactual prompts, LLMs adjust their responses to align with the given context, even if it contradicts their pretrained parametric knowledge \cite{li2022large,feng2024knowledge}. In other cases, however, steering LLMs away from generating outputs aligned with their vast but potentially inaccurate pretrained knowledge, even when contradicted by the context, has been challenging \cite{zhou-etal-2023-context}. Our work contributes to contextual grounding studies by prompting LLMs in a {\em knowledge-consistent} scenario and is the first to systematically analyze how the amount of atomic information in a prompt affects context utilization. %This type of scenario is more applicable to many real-world tasks.

\noindent \textbf{Hallucination} \quad There has been extensive work on hallucination detection with methods falling into two general categories: internal parameter based methods \cite{chen2024inside, Su-UnsupervisedHaluDetection, duan2024llms} and external response based methods \cite{manakul-etal-2023-selfcheckgpt,min-etal-2023-factscore, yu2024kola, manakul-etal-2023-selfcheckgpt,sun2024benchmarking}. Similar to our work, \citet{hu2024does} analyze the tendency of LLMs to generate factual hallucinations in the presence of varying local context in question answering. Unlike our approach, they study the effects of entirely switching out local context, whereas our approach incrementally increases the amount of context while comparing hallucination tendency between these responses. %They find that all models exhibit hallucination under context transfer, particularly in the presence of noisy, irrelevant context. 

\section{Conclusion}
Understanding how LLMs handle different context sizes is crucial for developing robust models. Our evaluation of nine widely used LLMs with our \texttt{WikiAtomic} dataset showed that all models process context similarly, balancing contextual and parametric knowledge, while adjusting response lengths consistently. They never use all provided contexts, always include some parametric knowledge, and prioritize information sequentially. As context increases, while contextual knowledge remains similar, parametric knowledge becomes more aligned with the context, and hallucination rates decrease. These insights highlight the importance of organizing context effectively and developing models that utilize input more deterministically.

%Our findings highlight that slight adjustments to the question prompt can influence the balance between contextual and parametric knowledge, enhancing the models' effectiveness in real-world applications.

%\section*{Ethical Considerations (optional)}

\section*{Limitations}

    % \item larger model variants - limited computation power
    % \item  varibility of LLM's output
   Our work, while yielding some interesting findings, is not without limitations.  Our method of extracting atomic sentences from Wikipedia articles often split the source sentences into multiple, occasionally creating sentences that begin with an indirect reference to a subject. Because of the natural flow of presenting information in such a format, we did not randomize the order. Further work could look at how the order of contexts used by models compared when the provided contexts are shuffled.

    We utilized the INFUSE method to classify model response contexts into `contextual' and `parametric'. We empirically set the threshold for determining the categorization based on manual verification. Further work could look at more sophisticated methods to set this threshold for even better results. A limitation of the metric we adopt for hallucination detection is that it relies on a single knowledge source Wikipedia as its knowledge source. Information that is factually accurate, but not present in the knowledge source could be incorrectly classified as a hallucination.

\section*{Acknowledgments}
We thank the anonymous reviewers as well as the
members of PortNLP lab for their insightful comments that helped improve this paper. This research was supported by the National
Science Foundation grant SAI-P 2228783.

\bibliography{references, custom}

%\clearpage
\renewcommand\thefigure{\thesection.\arabic{figure}}  
\setcounter{figure}{0}   
\appendix

\smallskip
\begin{tcolorbox}[colback=green!10!white, colframe=green!45!black, title=\textit{\textbf{Example Contexts}}, fonttitle=\sffamily]
\textbf{Contexts}:  \texttt{Mariah Carey was born on March 27, 1969. She was born in Huntington, New York. She is a highly celebrated American artist. She is known for her work as a singer. She is known for her work as a songwriter...(continue)}
\end{tcolorbox}
\smallskip

\section{Wikipedia Article Contexts Extraction Prompt} 
\label{app:article_extraction}
Figure~\ref{fig:extraction_prompt} shows the exact prompt we used to extract atomic facts from Wikipedia articles. We have tested multiple versions of prompts to break down sentences into atomic sentences. We found if we only asking the model directly to break down sentences to atomic sentences, or only adding the definition of atomic sentence and asking it to break down sentences. The model performed very poorly, the main problem with these approaches was this version of atomic sentences weren't atomic enough. For example: 

\texttt{April is the fourth month of the year in both the Julian and Gregorian calendars}. 

This sentence could be treated as a single information. But this definitly could be further broke down to smaller pieces: 

\texttt{1. April is the fourth month of the year in the Julian calendar} \\
\texttt{2. April is the fourth month of the year in the Gregorian calendar}.

Figure~\ref{fig:extraction_prompt}'s prompt was the most effective version in our experiments, it could handle the above example very well and at the same time add the appropriate subject to each atomic sentences.  We used this version of prompt also to break down responses from each model into atomic sentences (Figure \ref{fig:response_atomization}).

\begin{figure*}[t!]
    \centering
    \includegraphics[width=\textwidth]{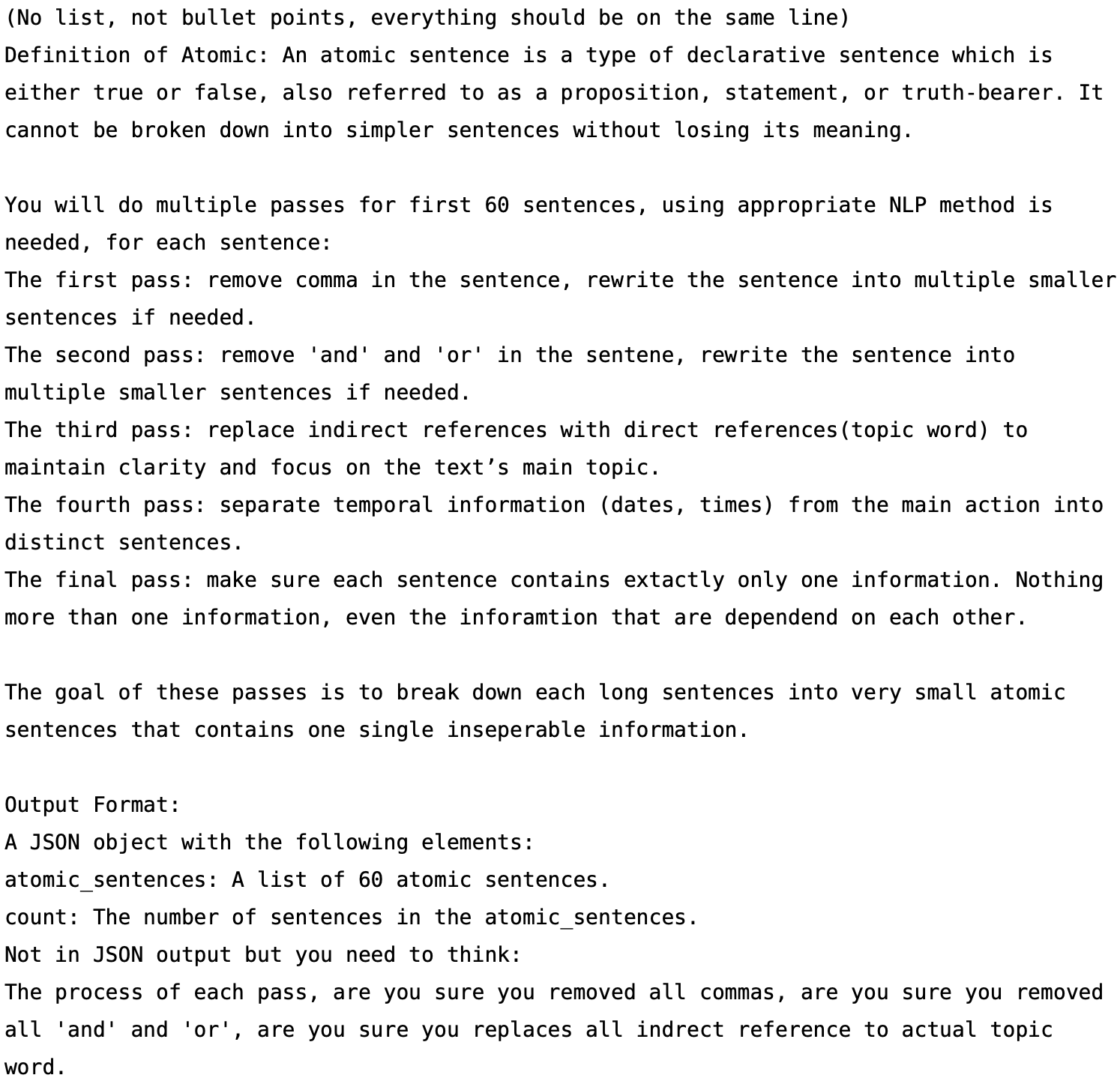}
    \caption{Prompt to extract atomic facts from article}
    \label{fig:extraction_prompt}
\end{figure*}

\begin{figure*}[t!]
    \centering
    \includegraphics[width=\textwidth]{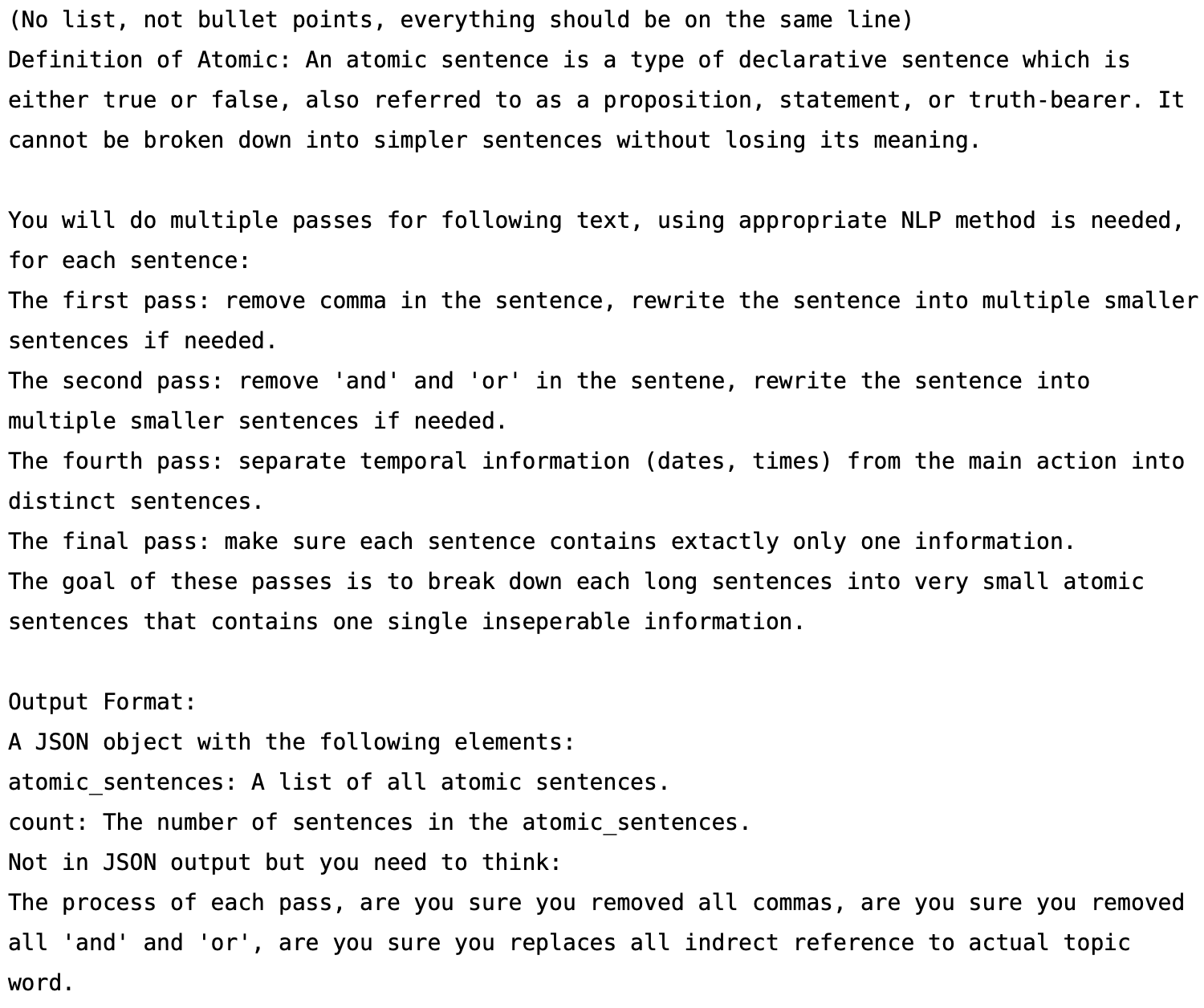}
    \caption{Response Atomization Prompt}
    \label{fig:response_atomization}
\end{figure*}

% \begin{figure*}
%     \centering
%     \begin{subfigure}[t]{0.49\textwidth}
%     \includegraphics[width=\textwidth]{figures/ablation/open_llama370b.png}
%     \caption{No restriction in question (Llama 3 70B)}
%     \label{fig:open_llama370b_main}
%     \end{subfigure}
%     \hfill
%     \begin{subfigure}[t]{0.49\textwidth}
%     \includegraphics[width=\textwidth]{figures/ablation/strict_llama370b.png}
%     \caption{Strict restriction in question (Llama 3 70B)}
%     \label{fig:strict_llama370b_main}
%     \end{subfigure}
%     \caption{Representative patterns when prompt with open/strict question}
%     \label{fig:represent_open_strict}
% \end{figure*}

\section{Alternative Metrics for Contextual/Parametric Knowledge Evaluation}
\label{app:alternative_methods_evaluation}
Figure \ref{fig:alternative_2} shows preliminary results using ROUGE-L, METEOR, and a partial match approach to evaluate GPT-4o model responses. In the figure, local facts and global facts represent contextual knowledge and parametric knowledge, respectively.
\begin{figure*}[t!]
    \centering
    % \begin{subfigure}[t]{0.49\textwidth}
    % \includegraphics[width=\textwidth]{figures/alternative_approach/LLM approach.png}
    % \caption{LLM approach results on GPT-4o}
    % \label{fig:LLM_APPROACH}
    % \end{subfigure}
    % \hfill
    \begin{subfigure}[t]{0.49\textwidth}
    \includegraphics[width=\textwidth]{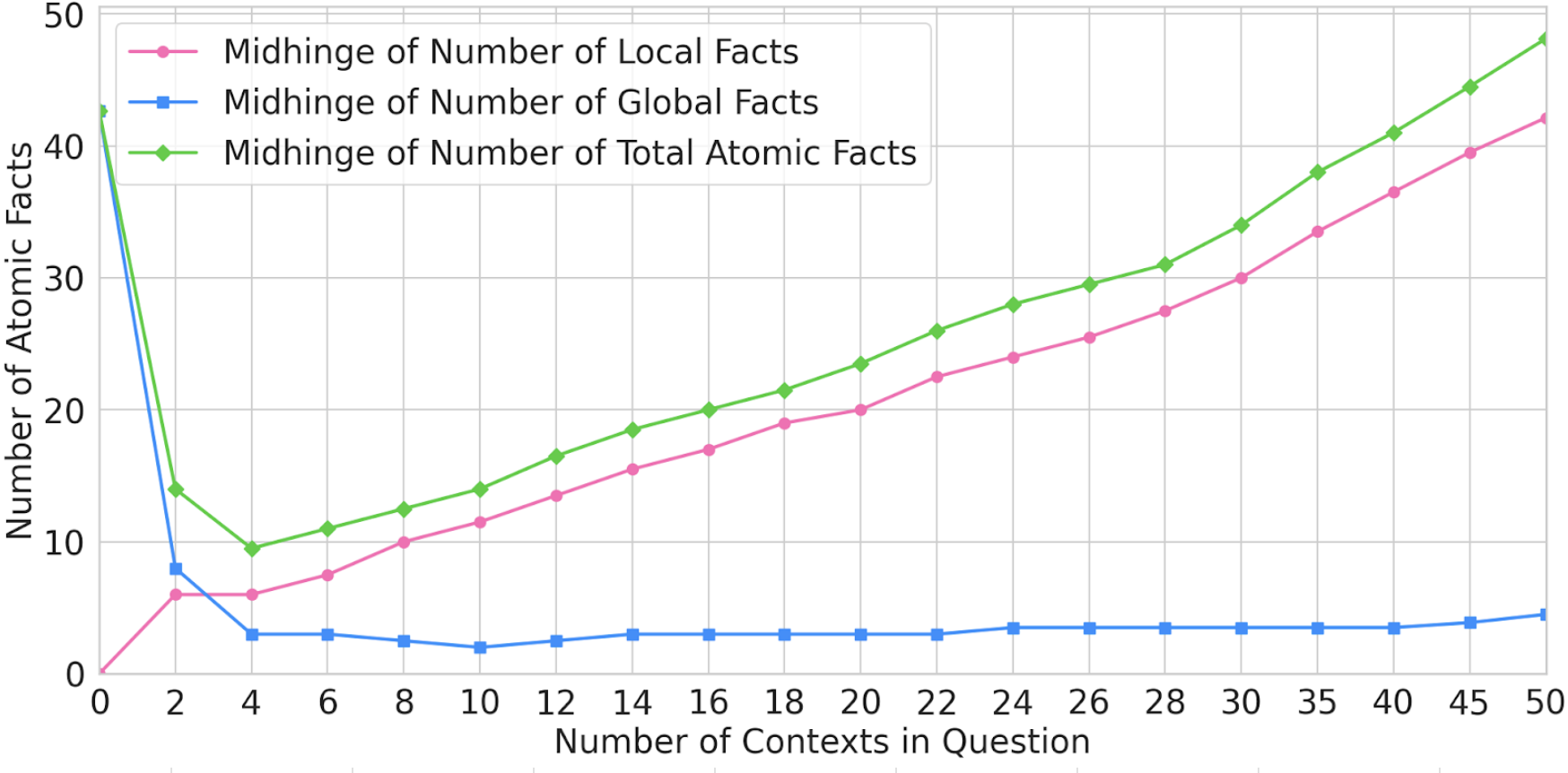}
    \caption{Partial match approach results on GPT-4o}
    \label{fig:partial_match_approach}
    \end{subfigure}
    \begin{subfigure}[t]{0.49\textwidth}
    \includegraphics[width=\textwidth]{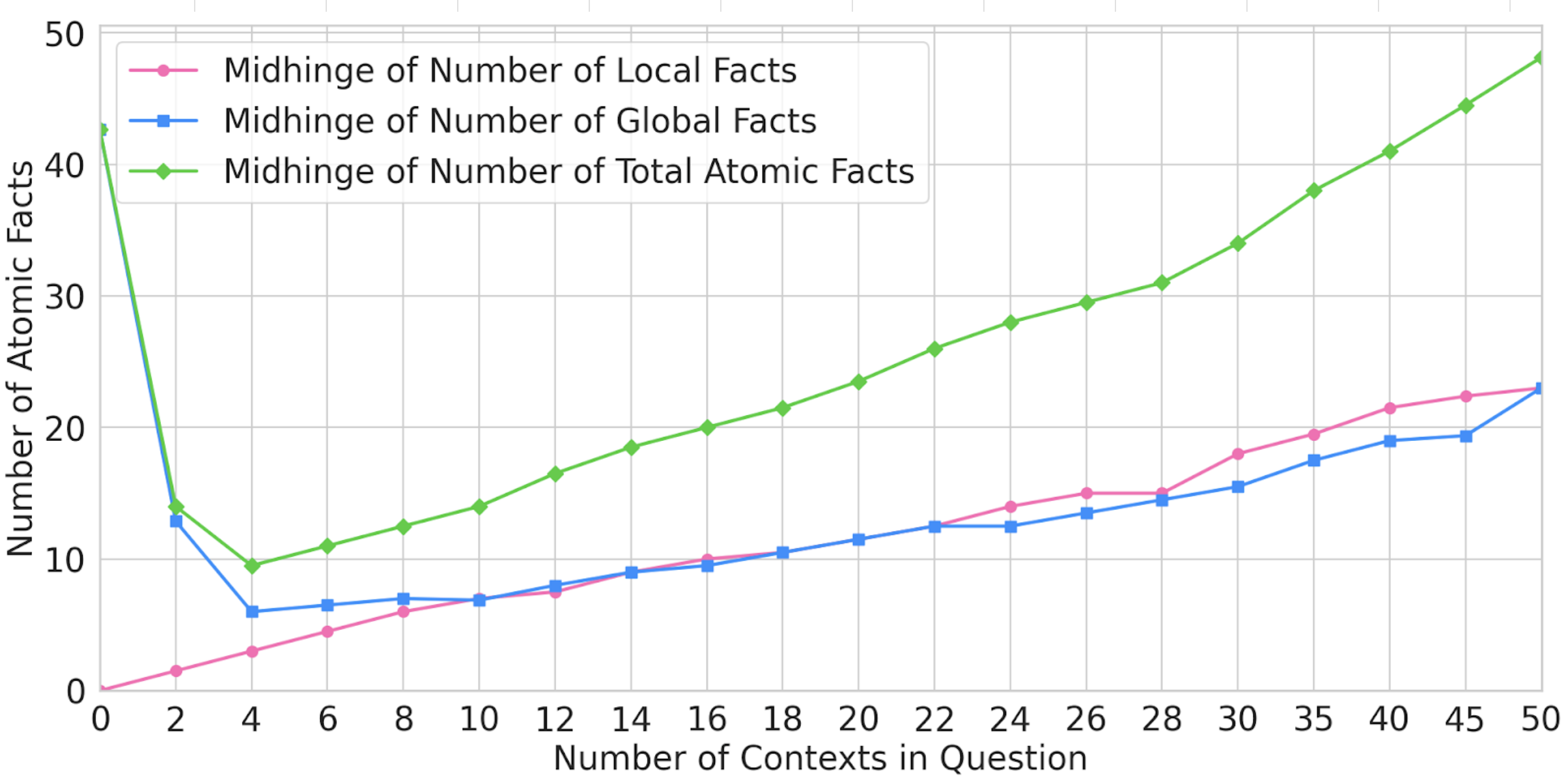}
    \caption{ROUGE-L approach results on GPT-4o}
    \label{fig:rouge_l_approach}
    \end{subfigure}
    \hfill
    \begin{subfigure}[t]{0.49\textwidth}
    \includegraphics[width=\textwidth]{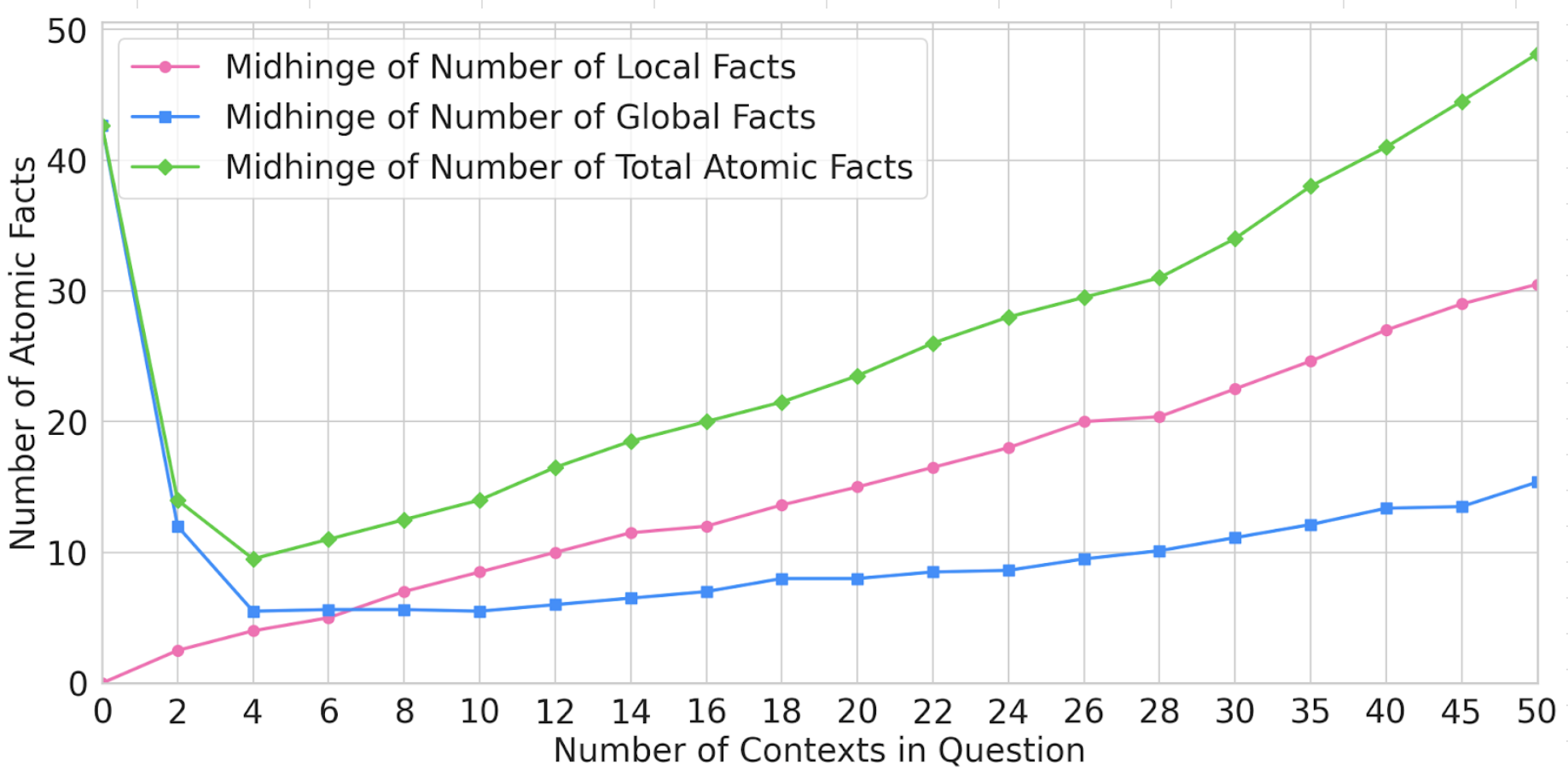}
    \caption{METEOR approach results on GPT-4o}
    \label{fig:meteor_approach}
    \end{subfigure}
    \caption{Alternative methods to evaluate contextual/parametric knowledge}
    \label{fig:alternative_2}
\end{figure*}

\section{Implementation Detail}
\label{app:implementation_detail}
Here we go through implementation details, model parameters. We used default parameters across all models where temperature and top\_p set to 1, presence\_penalty and frequency\_penalty set to 0. For atomic sentence extraction both for Wikipedia articles and models responses, we set GPT4-o's the max\_tokens to 2048 because the output is a JSON object that contains 50 atomic sentences. We think this number is sufficient for this size of output. When generating responses from different models once we have all the questions set up, we set the max\_token to 512 to mimic the real world application setting. 

API calls were made to OpenAI and Anthropic to obtain responses from GPT-4o, Claude 3 Opus, Sonnet, and Haiku, completing in 4 hours. For large open-source models Mixtral 8x22B and Llama 3 70B, we used a third-party service, which took 2 hours. The INFUSE evaluation of responses from 9 LLMs was conducted on 9 A100 GPUs over 10 hours. Inference for smaller models (Llama 3 8B and Phi 3) was performed on an M1 Ultra, taking 5 hours.

\section{Model-specific contextual/parametric evaluation}
\label{app:rest_of_nli}
Figure \ref{fig:mistral_llama3_combine_exp1} shows the rest of models' contextual/parametric evaluation results.

% \begin{figure*}
%     \centering
%     \begin{subfigure}[t]{0.49\textwidth}
%     \includegraphics[width=\textwidth]{figures/line_graphs/sonnet.png}
%     \caption{Claude Sonnet}
%     \label{fig:line_sonnet_exp1}
%     \end{subfigure}
%     \hfill
%     \begin{subfigure}[t]{0.49\textwidth}
%     \includegraphics[width=\textwidth]{figures/line_graphs/haiku.png}
%     \caption{Claude Haiku}
%     \label{fig:line_haiku_exp1}
%     \end{subfigure}
%     \caption{Rest of Claude models combined local, global, and total facts figures}
%     \label{fig:rest_of_claude_exp1}
% \end{figure*}

\begin{figure*}
    \centering
        \begin{subfigure}[t]{0.49\textwidth}
    \includegraphics[width=\textwidth]{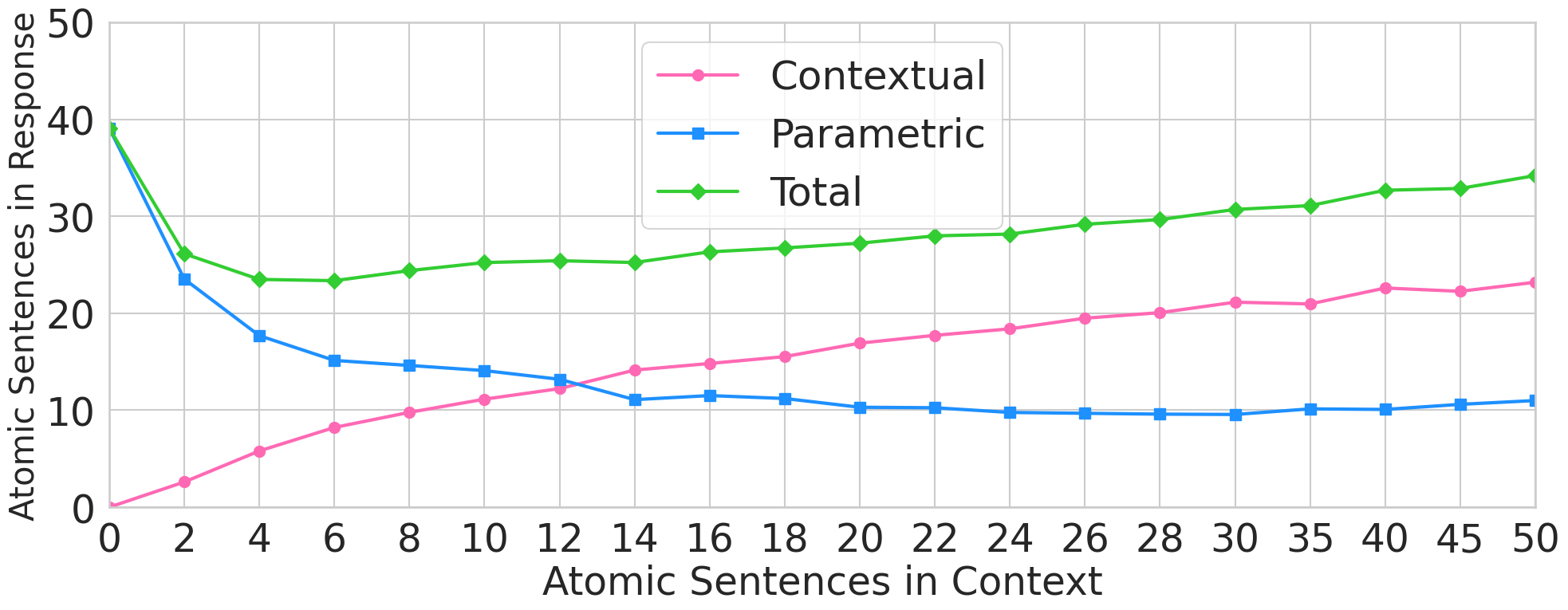}
    \caption{Claude Sonnet}
    \label{fig:line_sonnet_exp1}
    \end{subfigure}
    \hfill
    \begin{subfigure}[t]{0.49\textwidth}
    \includegraphics[width=\textwidth]{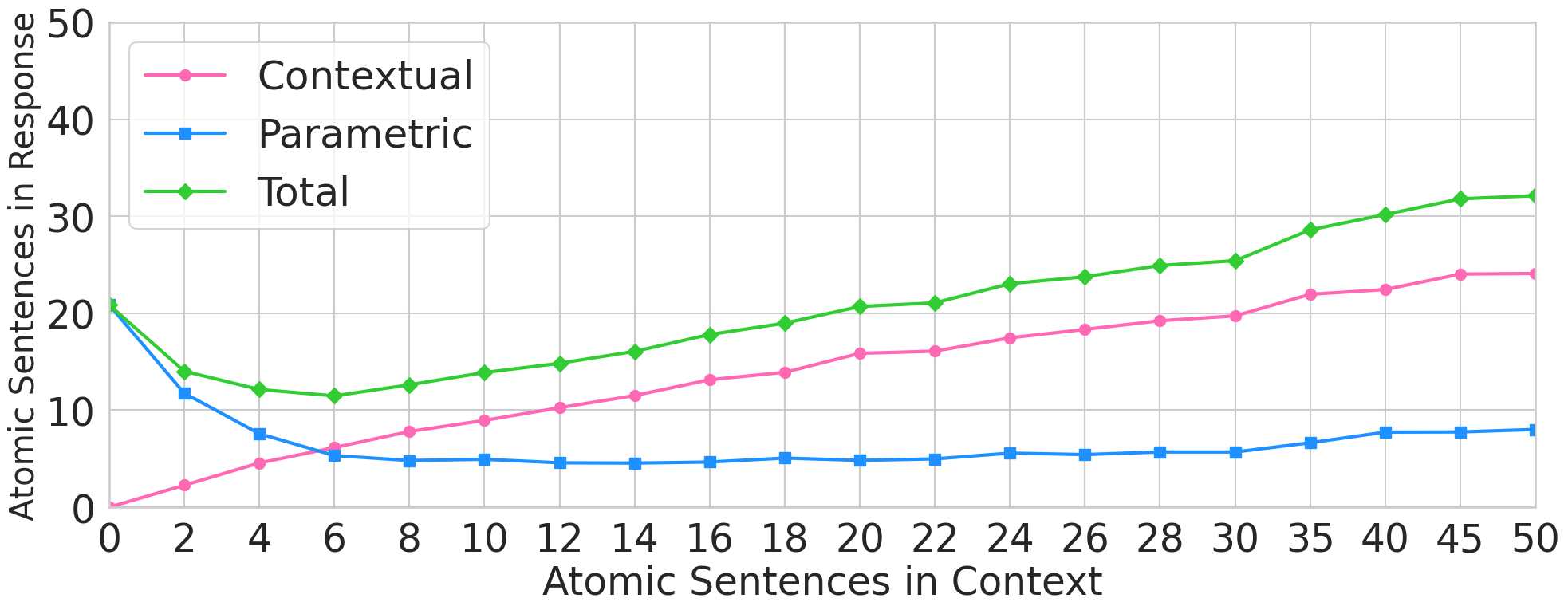}
    \caption{Claude Haiku}
    \label{fig:line_haiku_exp1}
    \end{subfigure}
    \begin{subfigure}[t]{0.49\textwidth}
    \includegraphics[width=\textwidth]{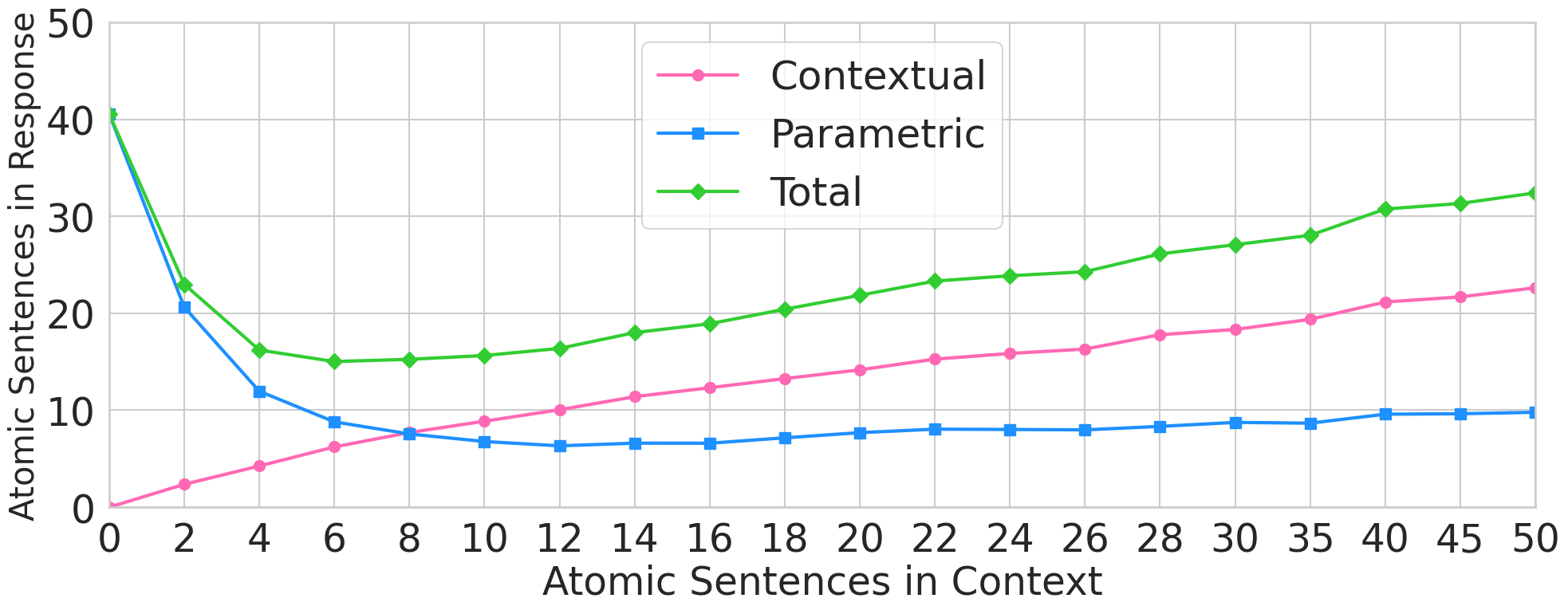}
    \caption{Llama 3 8B}
    \label{fig:llama3_8b_exp1}
    \end{subfigure}
    \hfill
    \begin{subfigure}[t]{0.49\textwidth}
    \includegraphics[width=\textwidth]{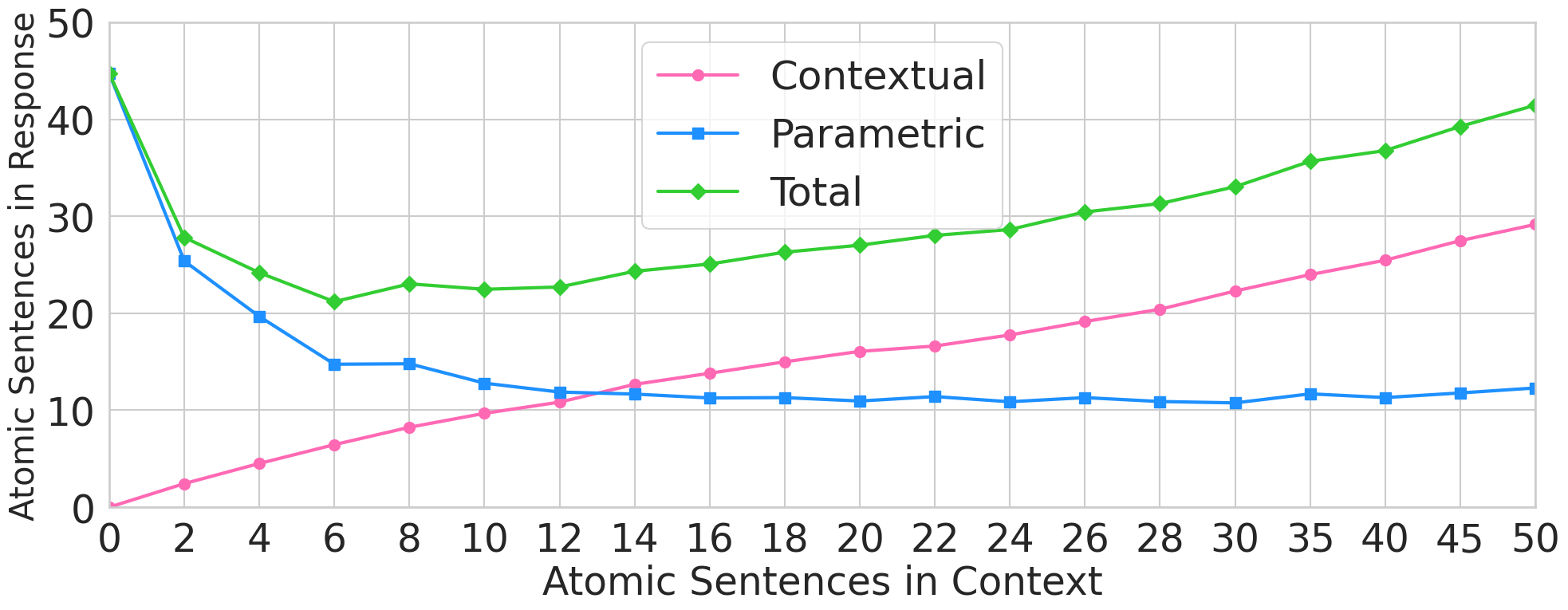}
    \caption{Mistral 7B}
    \label{fig:mixtral_exp1}
    \end{subfigure}
    \hfill
    \begin{subfigure}[t]{0.49\textwidth}
    \includegraphics[width=\textwidth]{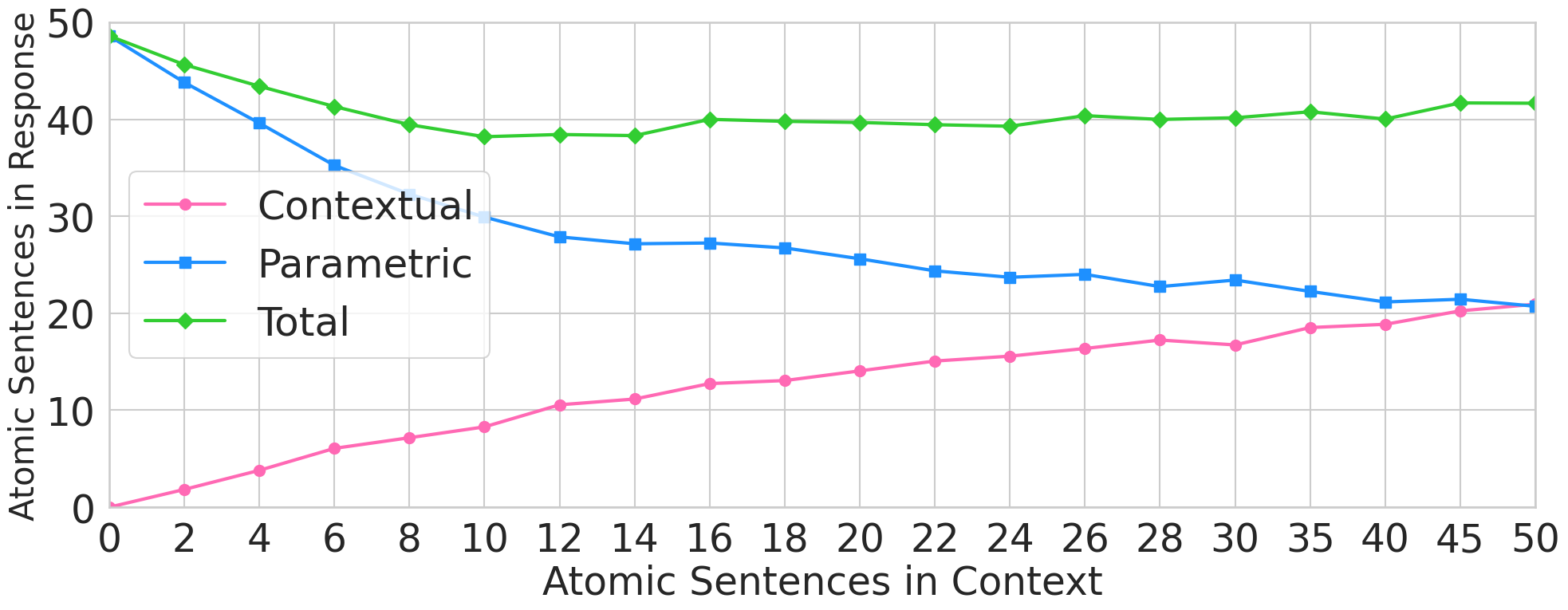}
    \caption{Phi-3 3}
    \label{fig:phi_exp1}
    \end{subfigure}
    \caption{Rest of models' combined local, global, and total facts figures}
    \label{fig:mistral_llama3_combine_exp1}
\end{figure*}

% \begin{figure*}[t!]
%     \centering
%     \includegraphics[width=0.7\textwidth]{figures/line_graphs/phi.png}
%     \caption{Phi-3 3.8B Combined local, global, and total facts figures}
%     \label{fig:phi_exp1}
% \end{figure*}

\section{Context Focus Position Analysis}
\label{app:position_analysis}
Figure \ref{fig:global_context_positions} shows the percentage of contexts were present in responses grouped by quartiles for each models.

\begin{figure*}[t!]
    \centering
    \begin{subfigure}[t]{0.32\textwidth}
    \includegraphics[width=\textwidth]{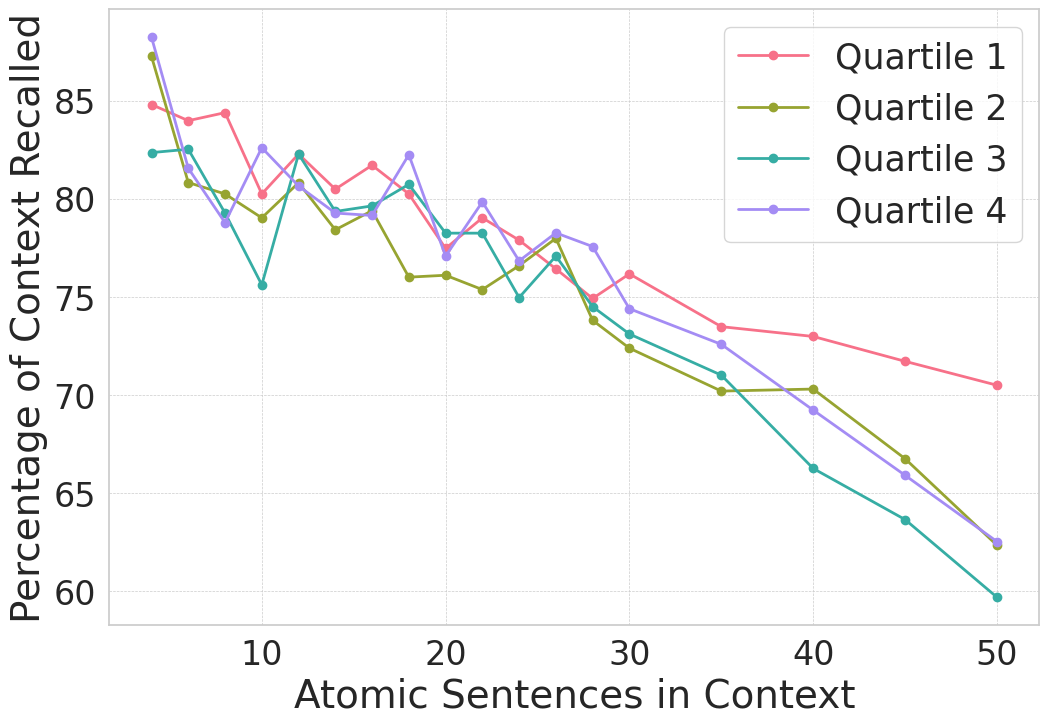}
    \caption{Claude Opus contexts positions}
    \label{fig:opus_position}
    \end{subfigure}
    \hfill
    \begin{subfigure}[t]{0.32\textwidth}
    \includegraphics[width=\textwidth]{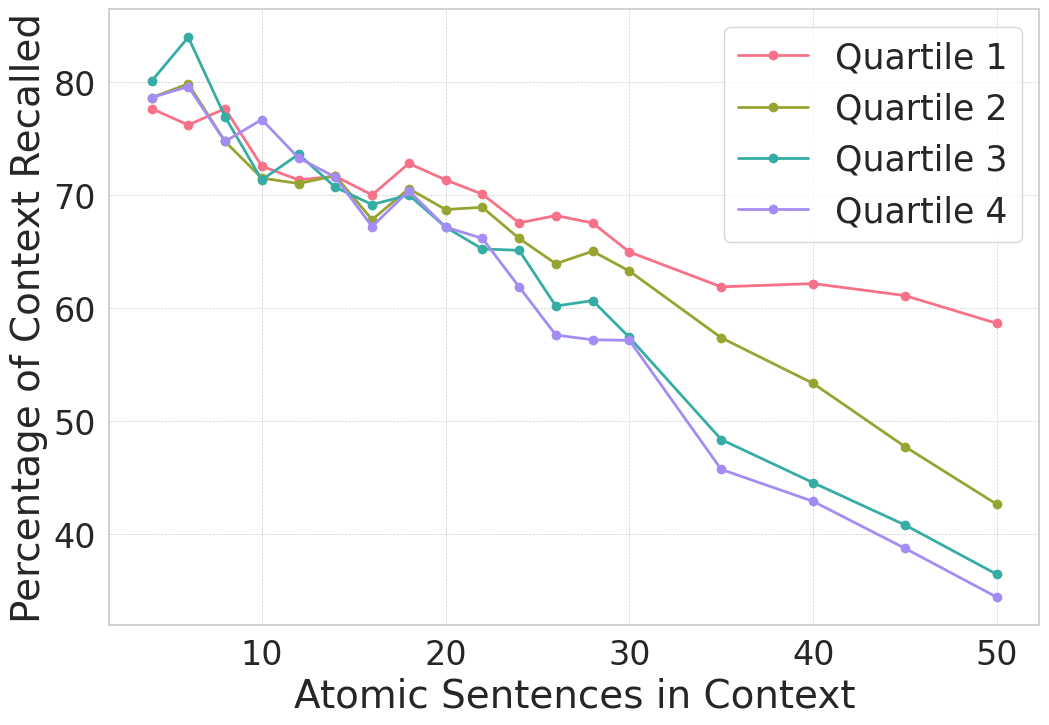}
    \caption{Claude Sonnet contexts positions}
    \label{fig:sonnet_position}
    \end{subfigure}
    \hfill
    \begin{subfigure}[t]{0.32\textwidth}
    \includegraphics[width=\textwidth]{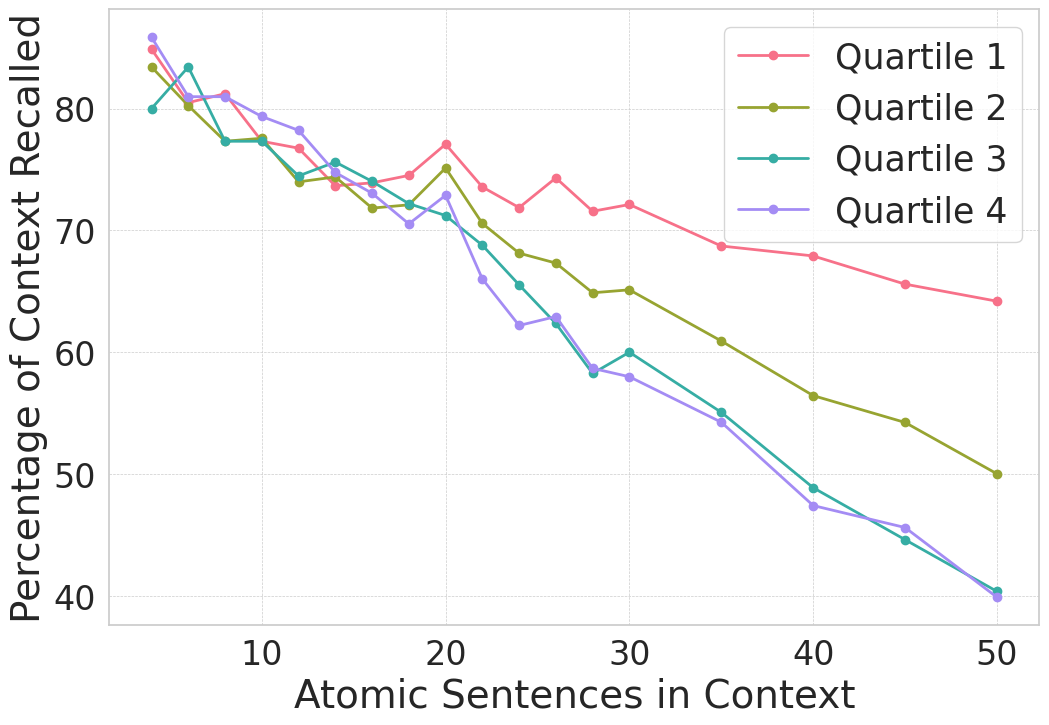}
    \caption{Claude Haiku contexts positions}
    \label{fig:haiku_position}
    \end{subfigure}
    \hfill
    \begin{subfigure}[t]{0.32\textwidth}
    \includegraphics[width=\textwidth]{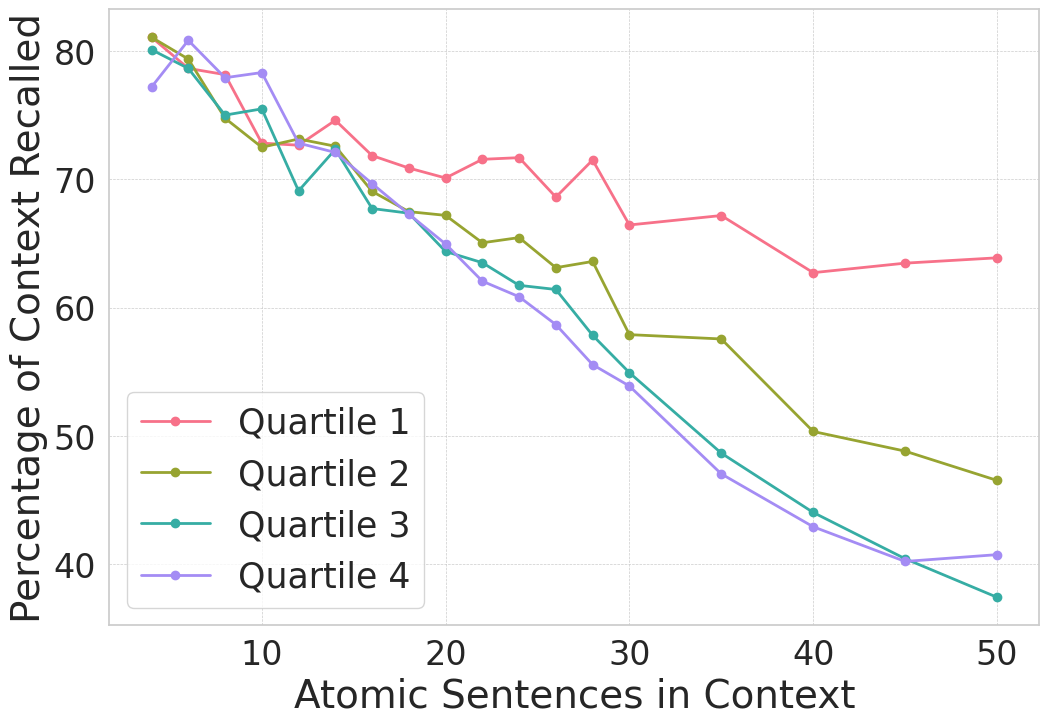}
    \caption{Llama 3 70B contexts positions}
    \label{fig:llama370b_position}
    \end{subfigure}
    \hfill
    \begin{subfigure}[t]{0.32\textwidth}
    \includegraphics[width=\textwidth]{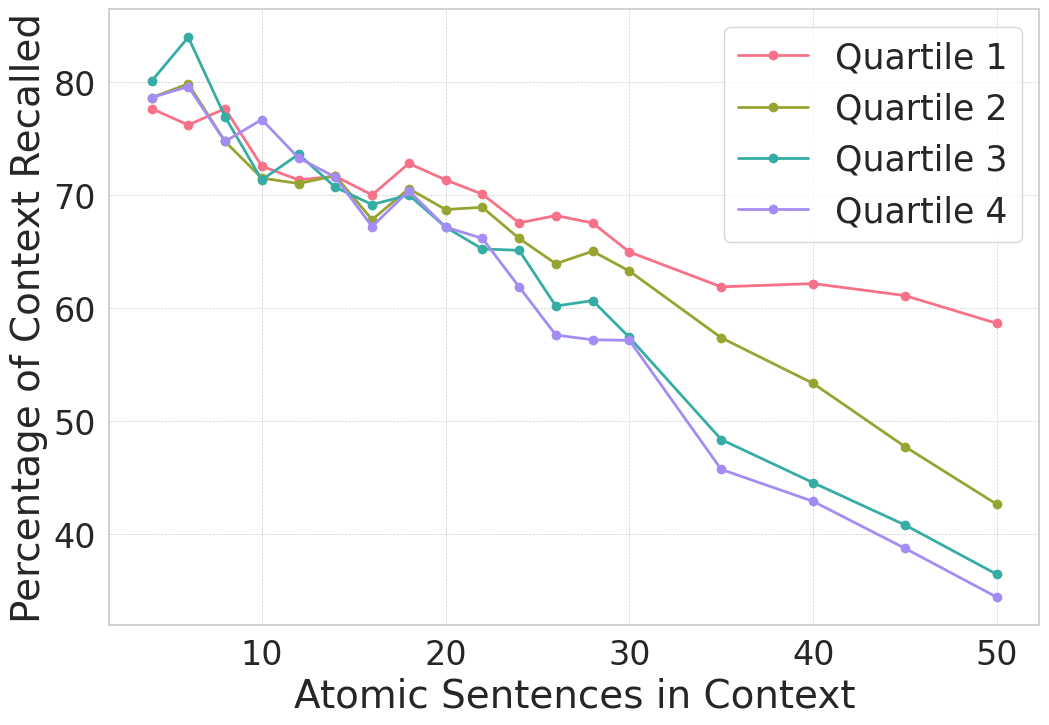}
    \caption{Llama 3 8B contexts positions}
    \label{fig:llama38b_position}
    \end{subfigure}
    \hfill
    \begin{subfigure}[t]{0.32\textwidth}
    \includegraphics[width=\textwidth]{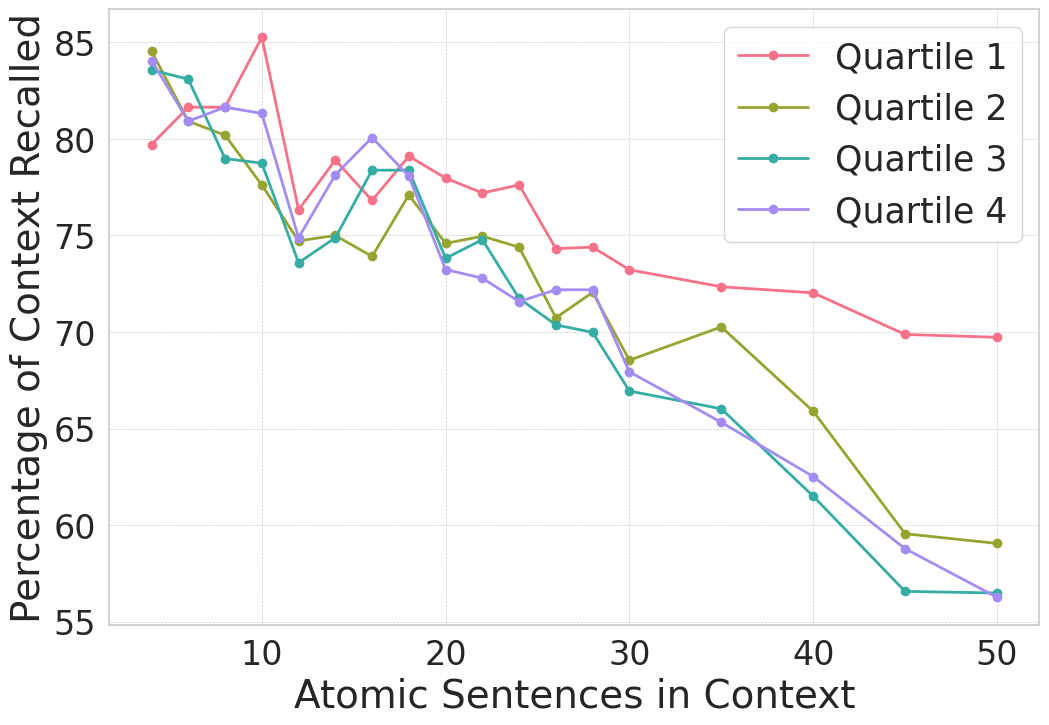}
    \caption{Mixtral 8x22B contexts positions}
    \label{fig:mixtal_position}
    \end{subfigure}
    \hfill
    \begin{subfigure}[t]{0.32\textwidth}
    \includegraphics[width=\textwidth]{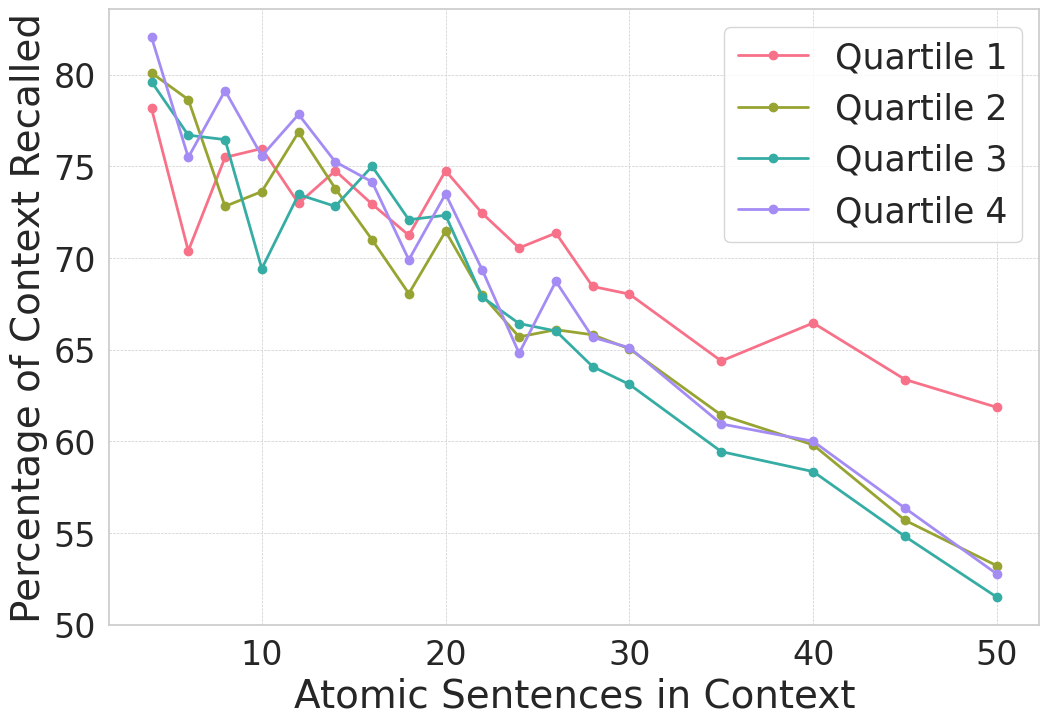}
    \caption{Mixtral 7B contexts positions}
    \label{fig:mistal_position}
    \end{subfigure}
    \hfill
    \begin{subfigure}[t]{0.32\textwidth}
    \includegraphics[width=\textwidth]{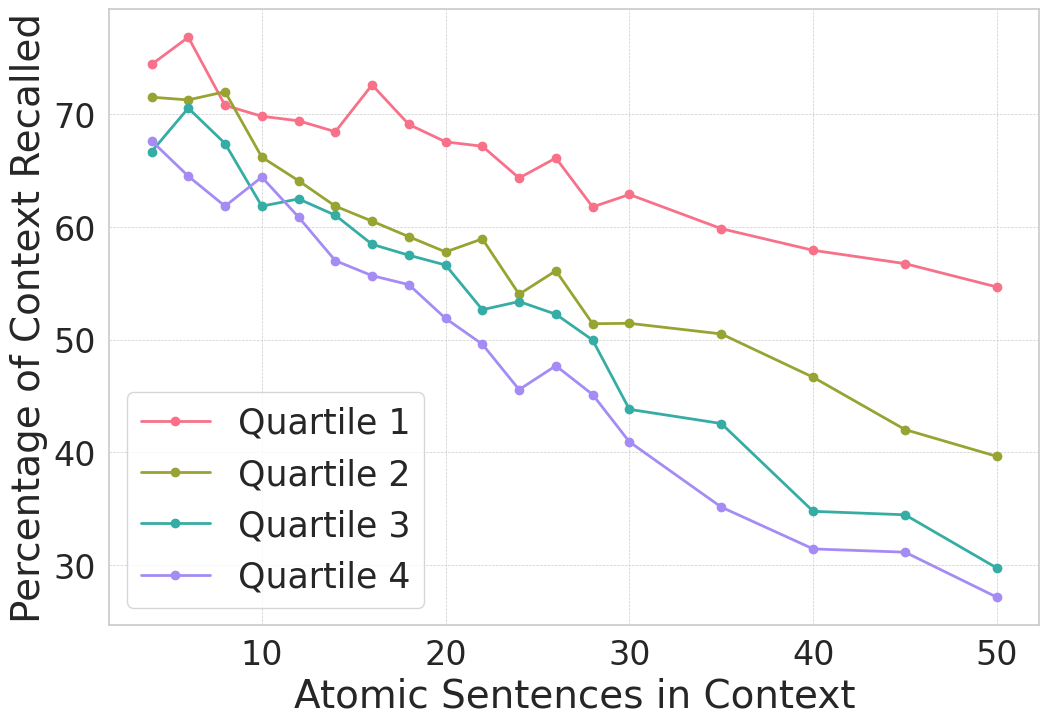}
    \caption{Phi 3 contexts positions}
    \label{fig:phi_position}
    \end{subfigure}
    \caption{Global context positions for various models}
    \label{fig:global_context_positions}
\end{figure*}

\section{Context Response Mapping Analysis}
\label{app:where_contexts_go}
Figure \ref{fig:where_contexts_went_area} shows results of each model's quartile contexts in response areas.

\begin{figure*}[t!]
    \centering
    % \begin{subfigure}[t]{0.49\textwidth}
    %     \centering
    %     \includegraphics[width=\textwidth]{figures/where_went/gpt4o.png}
    %     \caption{GPT-4o quartile contexts in response positions}
    %     \label{fig:gpt4o_where}
    % \end{subfigure}
    \begin{subfigure}[t]{0.49\textwidth}
        \centering
        \includegraphics[width=\textwidth]{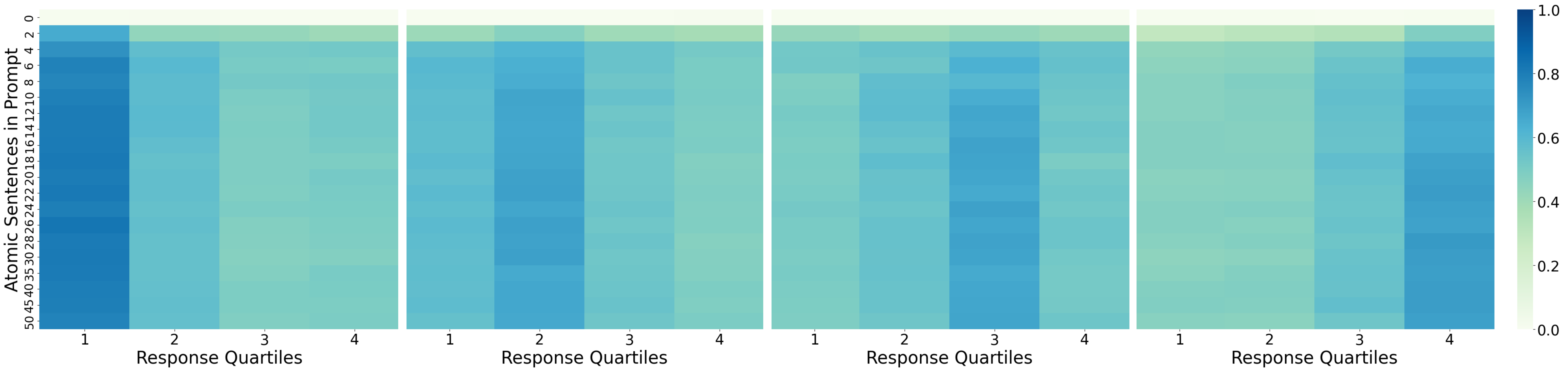}
        \caption{Claude Opus quartile contexts in response positions}
        \label{fig:opus_where}
    \end{subfigure}
    \begin{subfigure}[t]{0.49\textwidth}
        \centering
        \includegraphics[width=\textwidth]{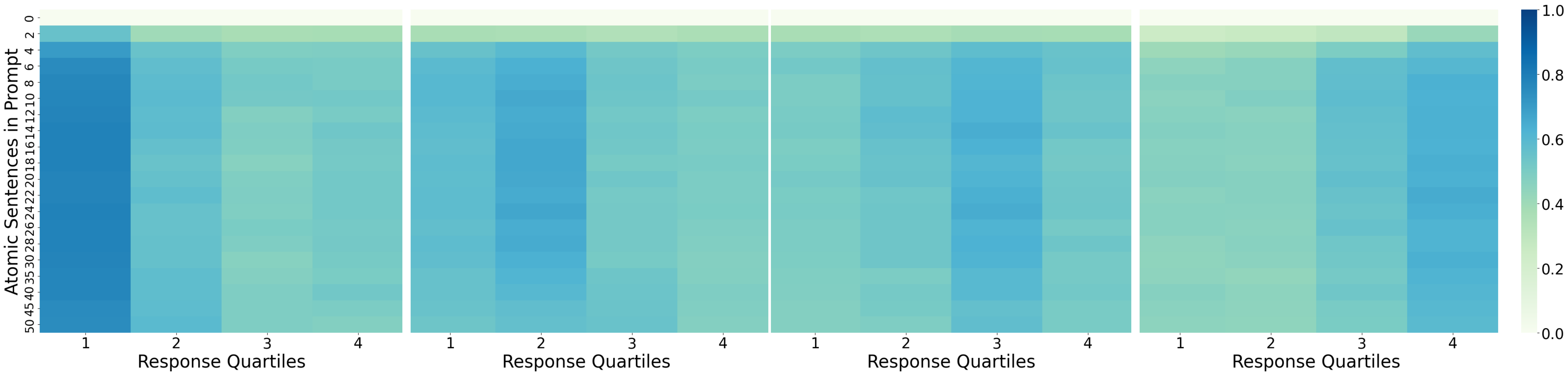}
        \caption{Claude Sonnet contexts in response positions}
        \label{fig:sonnet_where}
    \end{subfigure}
    \begin{subfigure}[t]{0.49\textwidth}
        \centering
        \includegraphics[width=\textwidth]{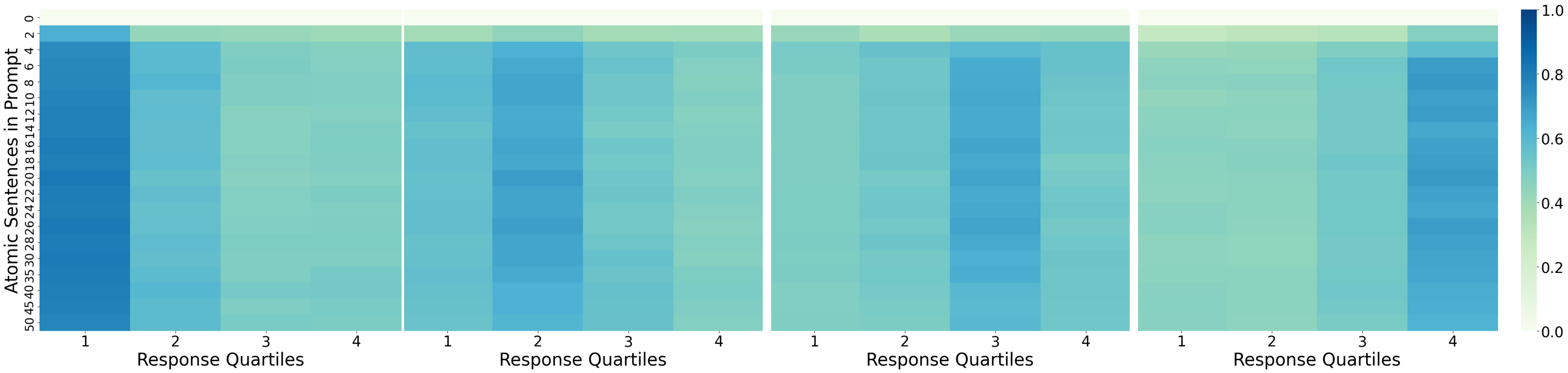}
        \caption{Claude Haiku contexts in response positions}
        \label{fig:haiku_where}
    \end{subfigure}
    \begin{subfigure}[t]{0.49\textwidth}
        \centering
        \includegraphics[width=\textwidth]{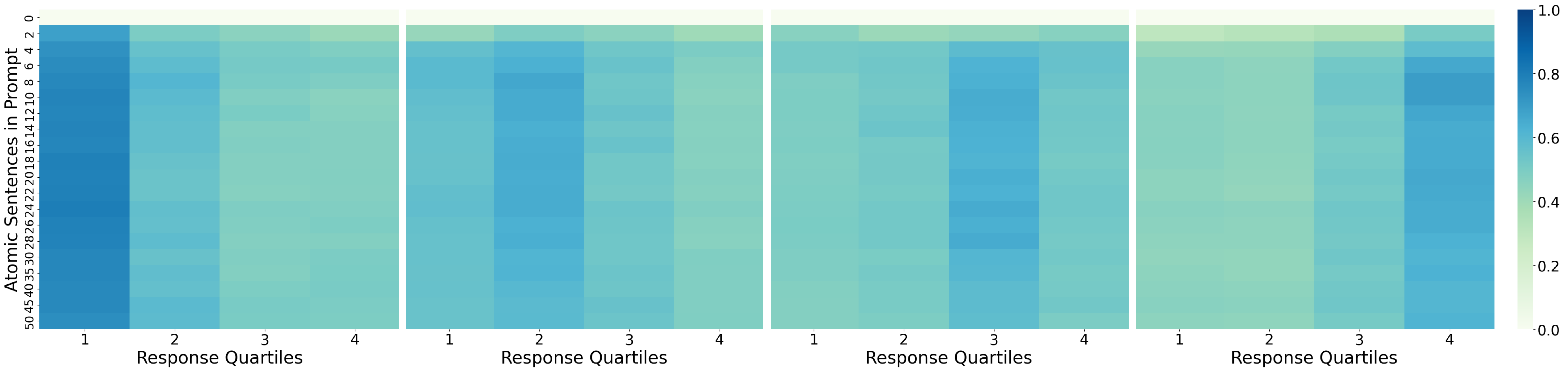}
        \caption{Llama 3 70B contexts in response positions}
        \label{fig:llama370b_where}
    \end{subfigure}
    \begin{subfigure}[t]{0.49\textwidth}
        \centering
        \includegraphics[width=\textwidth]{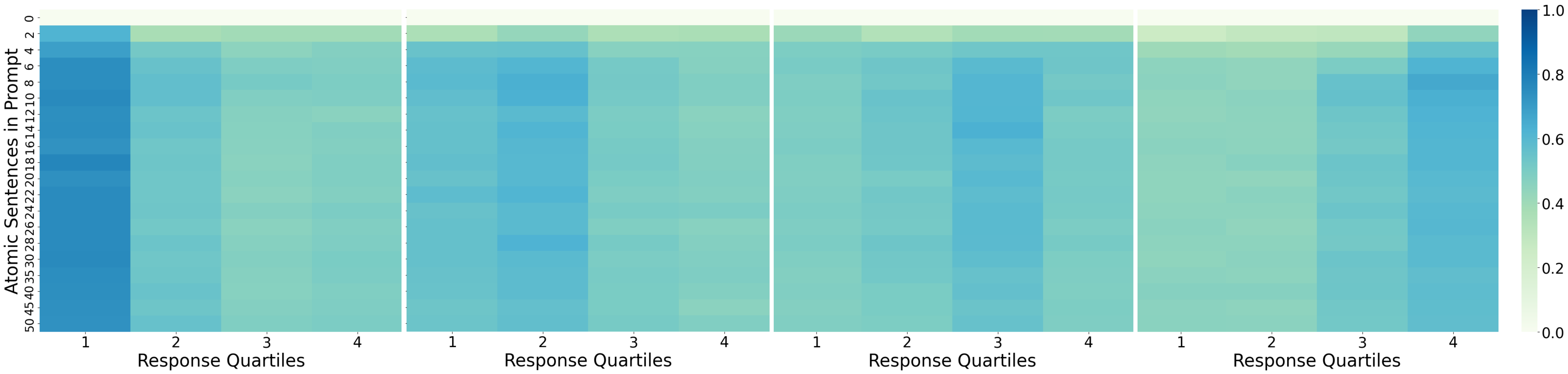}
        \caption{Llama 3 8B quartile contexts in response positions}
        \label{fig:gllama38b_where}
    \end{subfigure}
    \begin{subfigure}[t]{0.49\textwidth}
        \centering
        \includegraphics[width=\textwidth]{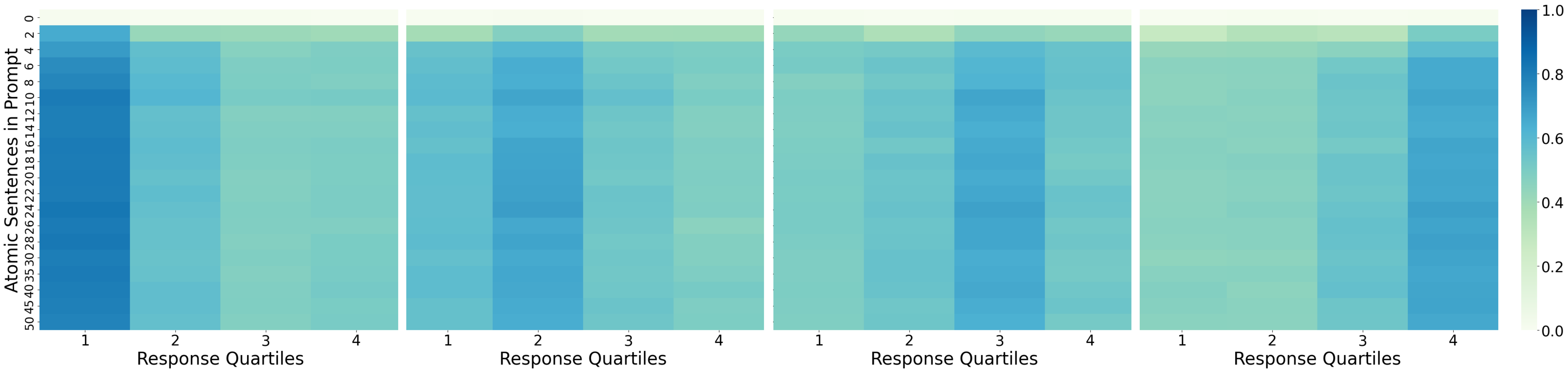}
        \caption{Mixtral 8x22B contexts in response positions}
        \label{fig:mixtral_where}
    \end{subfigure}
    \begin{subfigure}[t]{0.49\textwidth}
        \centering
        \includegraphics[width=\textwidth]{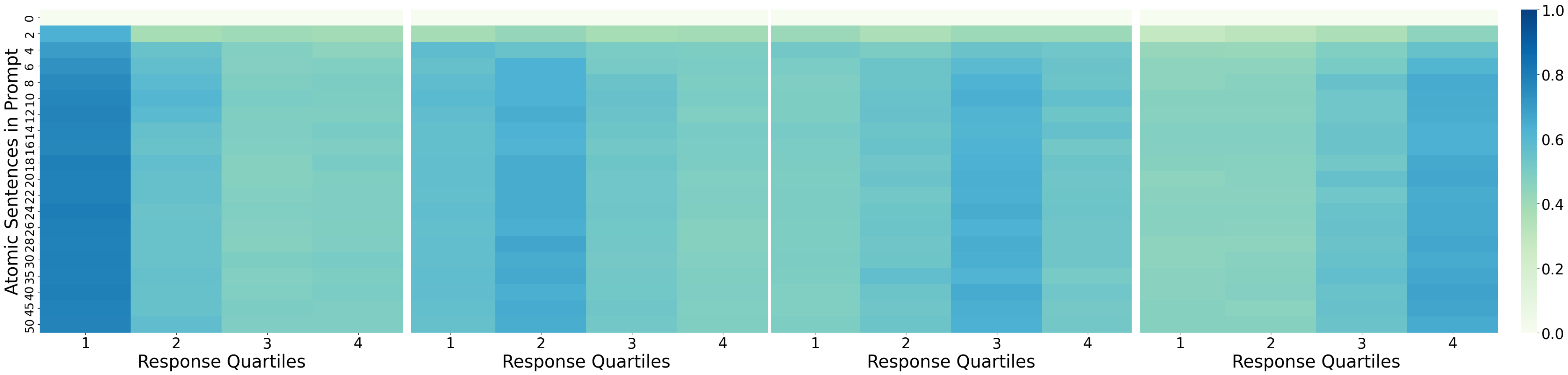}
        \caption{Mistral 7B quartile contexts in response positions}
        \label{fig:mistral_where}
    \end{subfigure}
    \begin{subfigure}[t]{0.49\textwidth}
        \centering
        \includegraphics[width=\textwidth]{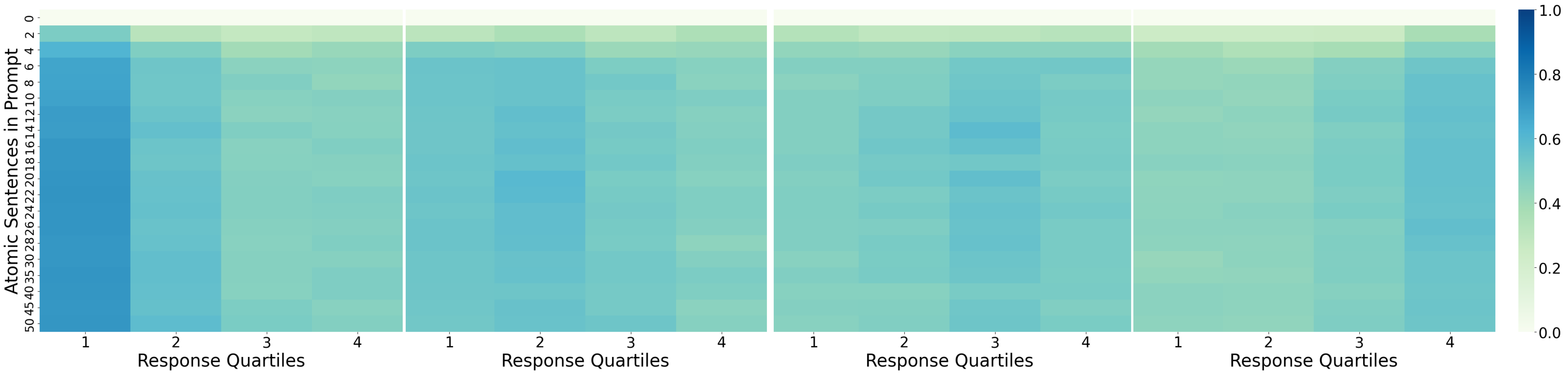}
        \caption{Phi quartile contexts in response positions}
        \label{fig:phi_where}
    \end{subfigure}
    \caption{Quartile contexts in response positions for various models}
    \label{fig:where_contexts_went_area}
\end{figure*}

\section{Local vs. Local Similarity}
\label{app:local_vs_local_simialrity}
Figure \ref{fig:local_local_similarity_each_model} shows contextual knowledge similarity across rest of 9 models other than the Claude Sonnet mentioned in the paper. 

\begin{figure*}[t!]
    \centering
    \begin{subfigure}[t]{0.32\textwidth}
    \includegraphics[width=1\textwidth]{figures/local_vs_local/gpt4o.png}
    \caption{GPT-4o}
    \label{fig:rvr_gpt4o}
    \end{subfigure}
    \hfill
    \begin{subfigure}[t]{0.32\textwidth}
    \includegraphics[width=\textwidth]{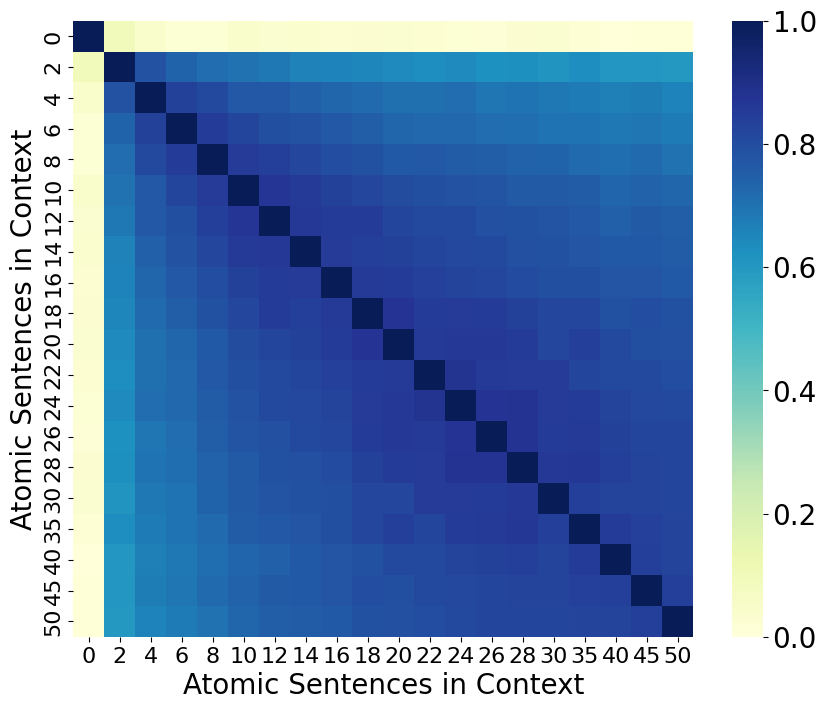}
    \caption{Llama 3 70B}
    \label{fig:rvr_llama70b}
    \end{subfigure}
    \hfill
    \begin{subfigure}[t]{0.32\textwidth}
    \includegraphics[width=\textwidth]{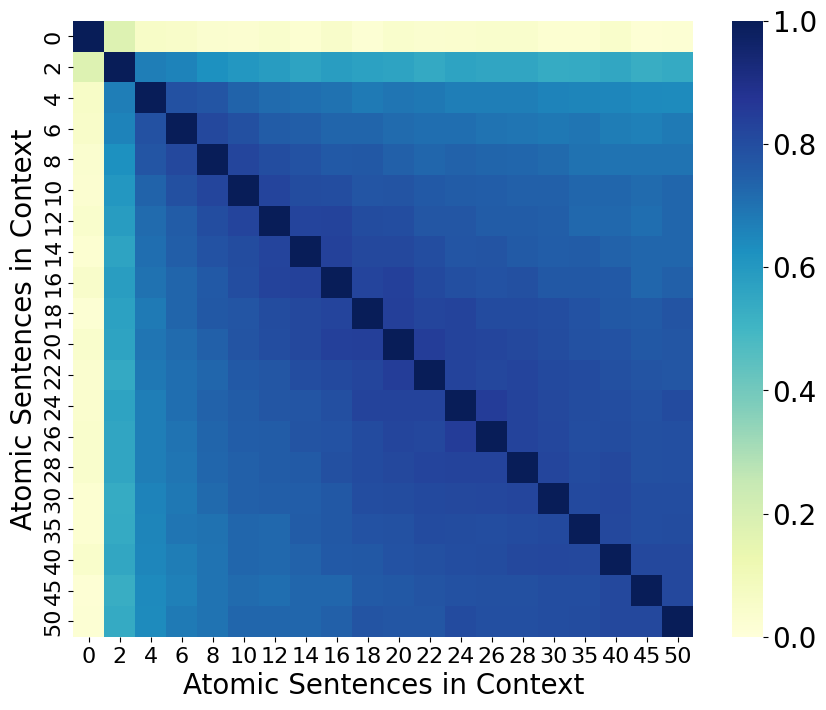}
    \caption{Llama 3 8B}
    \label{fig:rvr_llama7b}
    \end{subfigure}
    \hfill
    \begin{subfigure}[t]{0.32\textwidth}
    \includegraphics[width=\textwidth]{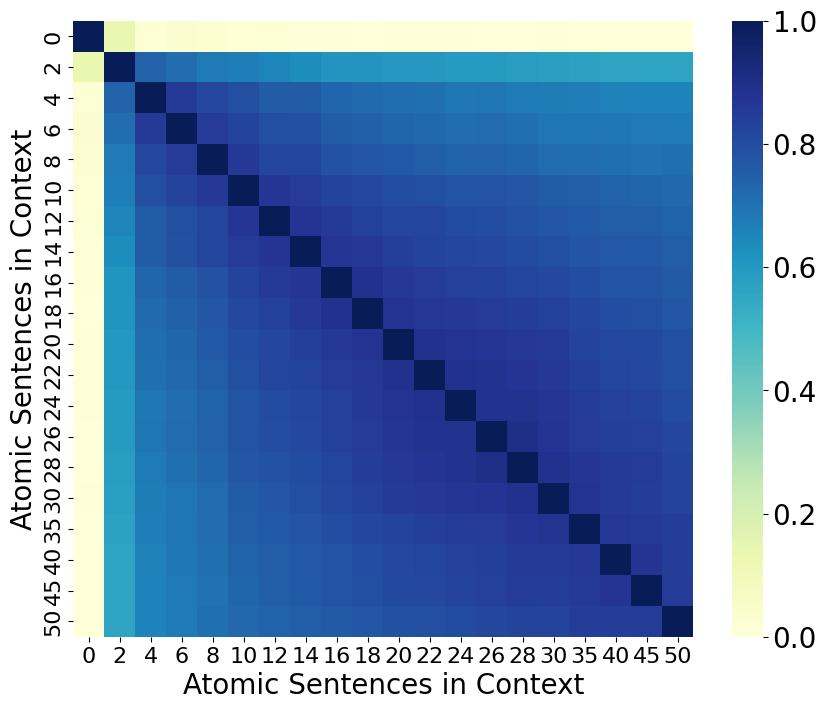}
    \caption{Claude Opus}
    \label{fig:rvr_opus}
    \end{subfigure}
    \hfill
    \begin{subfigure}[t]{0.32\textwidth}
    \includegraphics[width=\textwidth]{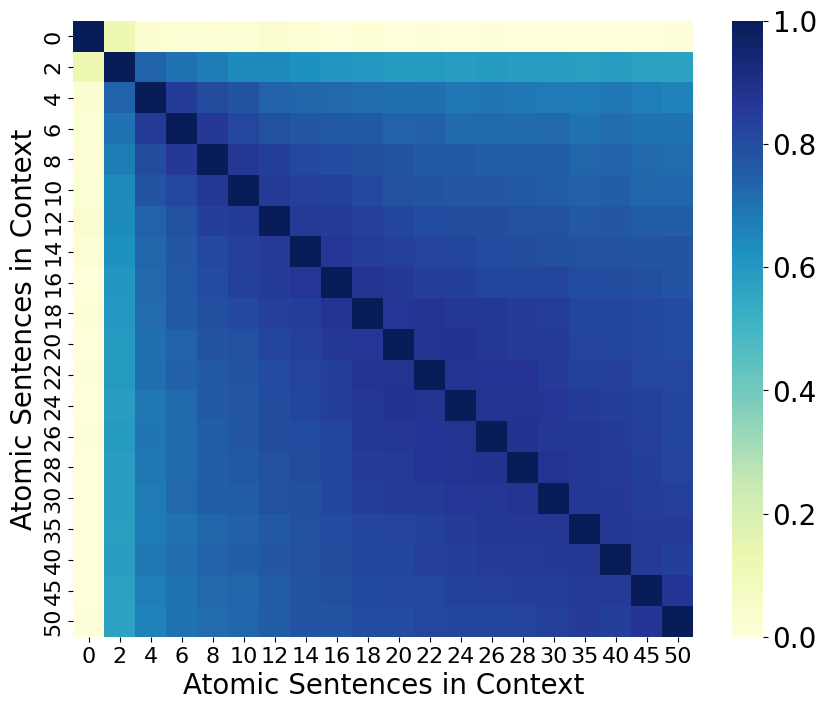}
    \caption{Claude Haiku}
    \label{fig:rvr_haiku}
    \end{subfigure}
    \hfill
    \begin{subfigure}[t]{0.32\textwidth}
    \includegraphics[width=\textwidth]{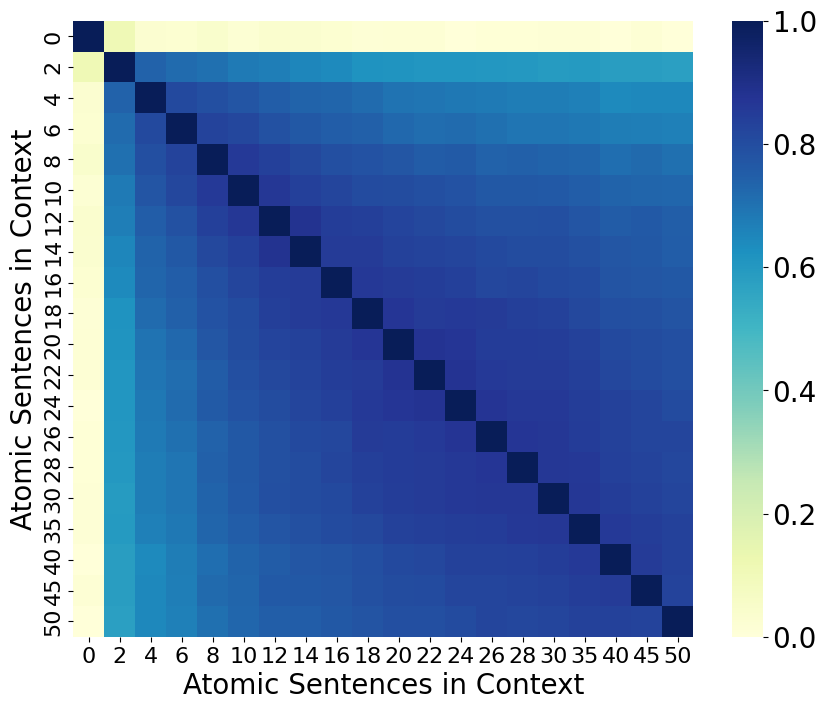}
    \caption{Mixtral 8x22B}
    \label{fig:rvr_mixtral}
    \end{subfigure}
    \hfill
    \begin{subfigure}[t]{0.32\textwidth}
    \includegraphics[width=\textwidth]{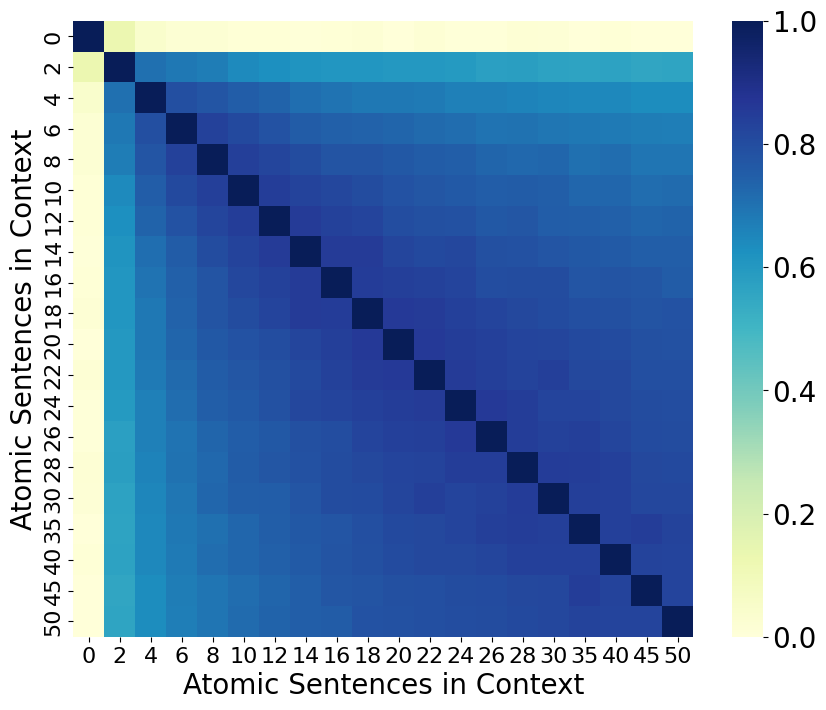}
    \caption{Mixtral 7B}
    \label{fig:rvr_mistral}
    \end{subfigure}
    \hfill
    \begin{subfigure}[t]{0.32\textwidth}
    \includegraphics[width=\textwidth]{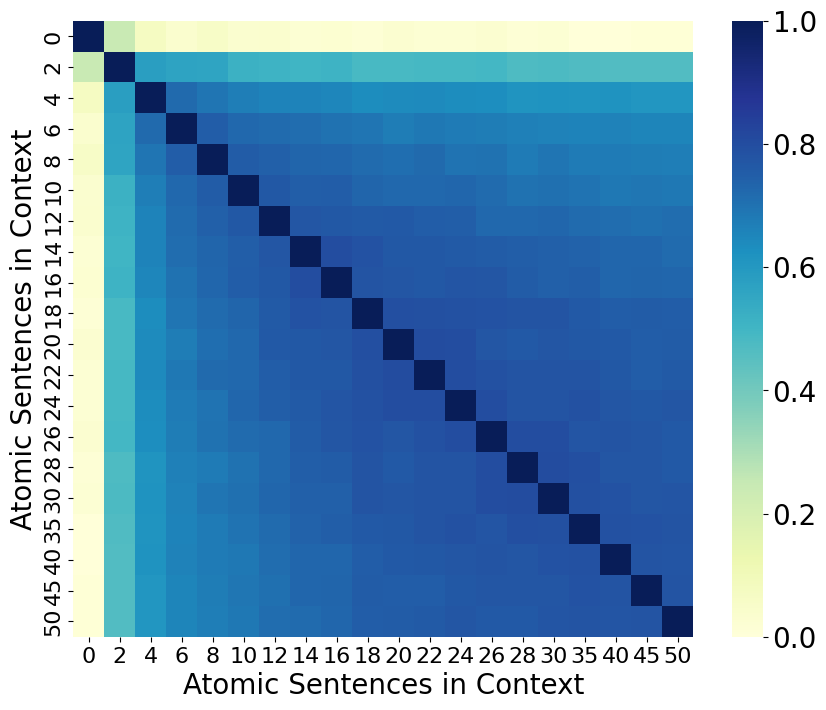}
    \caption{Phi 3}
    \label{fig:rvr_phi}
    \end{subfigure} 
    \caption{Similarity between contextual knowledge within a set of example for each model}
    \label{fig:local_local_similarity_each_model}
\end{figure*}

\section{Parametric vs. Parametric Knowledge in Response}
\label{app:global_global_similarity}
Figure \ref{fig:global_global_similarity_appendix} shows parametric knowledge similarity across each set of example for rest of the models not mentioned in main paper.

\begin{figure*}[t!]
    \centering
    % \begin{subfigure}[t]{0.32\textwidth}
    %     \includegraphics[width=\textwidth]{figures/global_global_similarity/gvg_gpt4o.png}
    %     \caption{GPT-4o global vs global}
    %     \label{fig:gpt4o_gvg}
    % \end{subfigure}
    % \hfill
    \begin{subfigure}[t]{0.32\textwidth}
        \includegraphics[width=\textwidth]{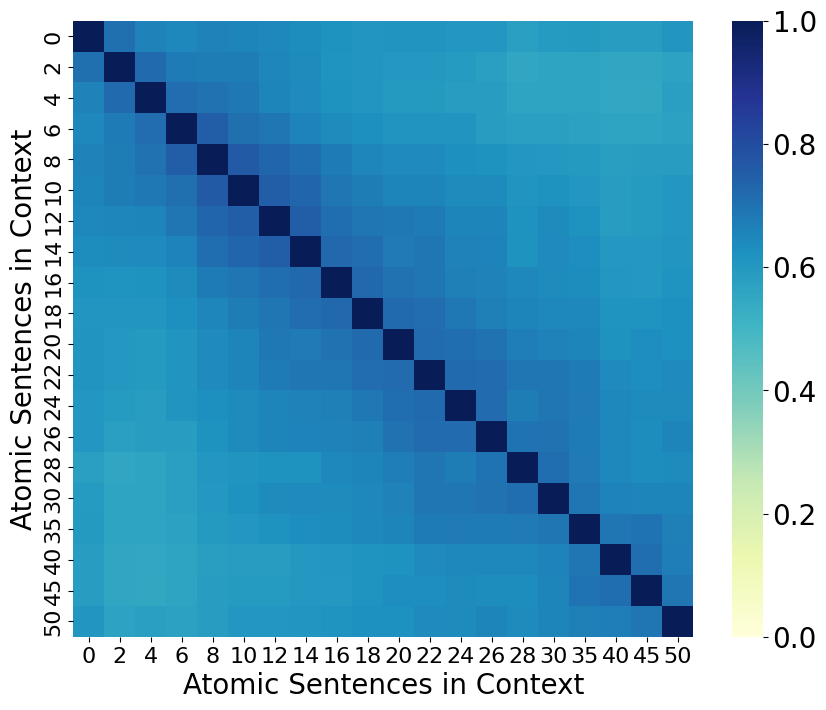}
        \caption{Opus global vs global}
        \label{fig:gvg_opus}
    \end{subfigure}
    \hfill
    \begin{subfigure}[t]{0.32\textwidth}
        \includegraphics[width=\textwidth]{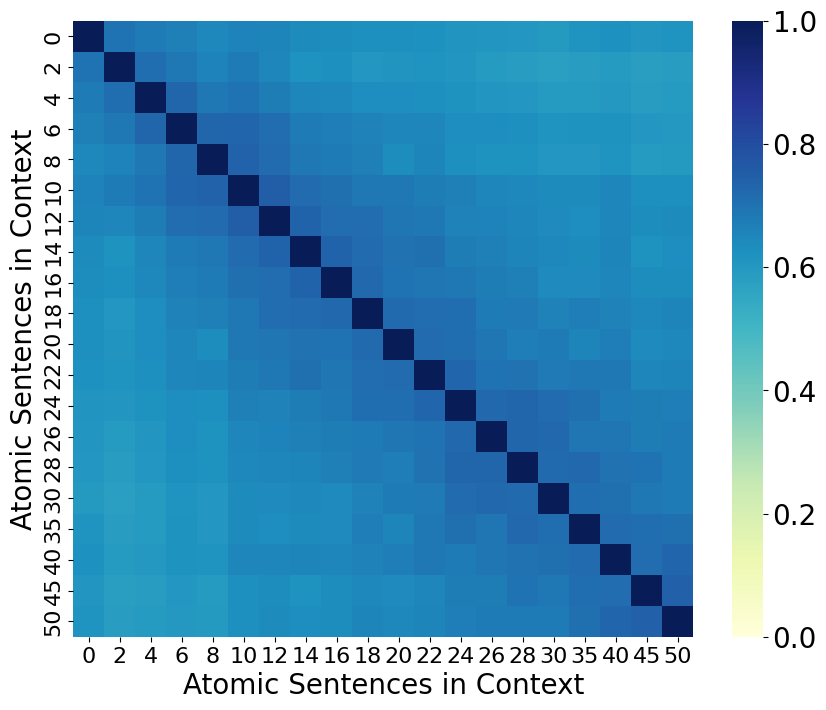}
        \caption{Sonnet global vs global}
        \label{fig:gvg_sonnet}
    \end{subfigure}
    \hfill
    \begin{subfigure}[t]{0.32\textwidth}
        \includegraphics[width=\textwidth]{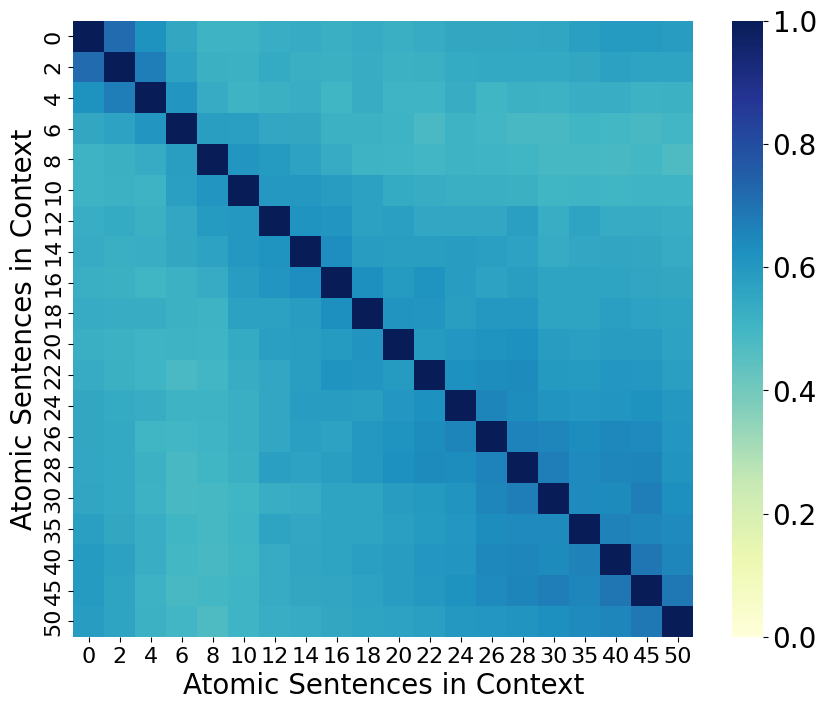}
        \caption{Haiku global vs global}
        \label{fig:gvg_haiku}
    \end{subfigure}
    \hfill
    \begin{subfigure}[t]{0.32\textwidth}
        \includegraphics[width=\textwidth]{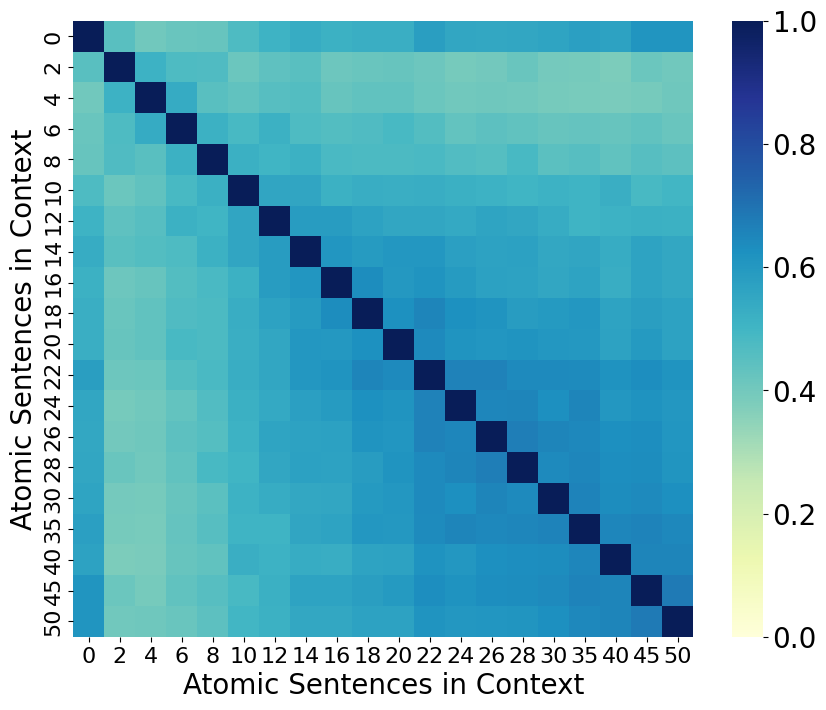}
        \caption{Llama 70B global vs global}
        \label{fig:gvg_llama70b}
    \end{subfigure}
    \hfill
    \begin{subfigure}[t]{0.32\textwidth}
        \includegraphics[width=\textwidth]{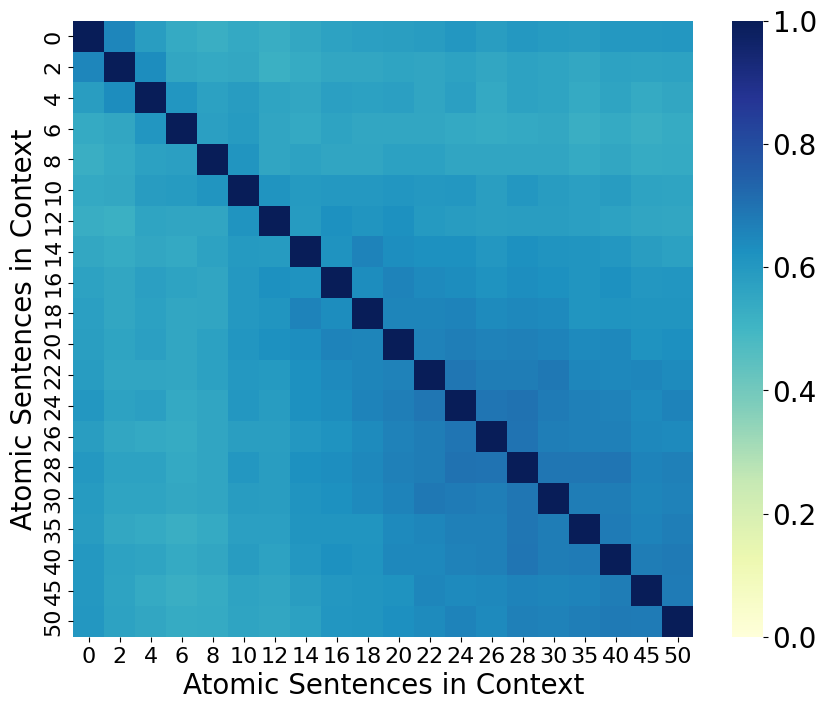}
        \caption{Llama 7B global vs global}
        \label{fig:gvg_llama7b}
    \end{subfigure}
    \hfill
    \begin{subfigure}[t]{0.32\textwidth}
        \includegraphics[width=\textwidth]{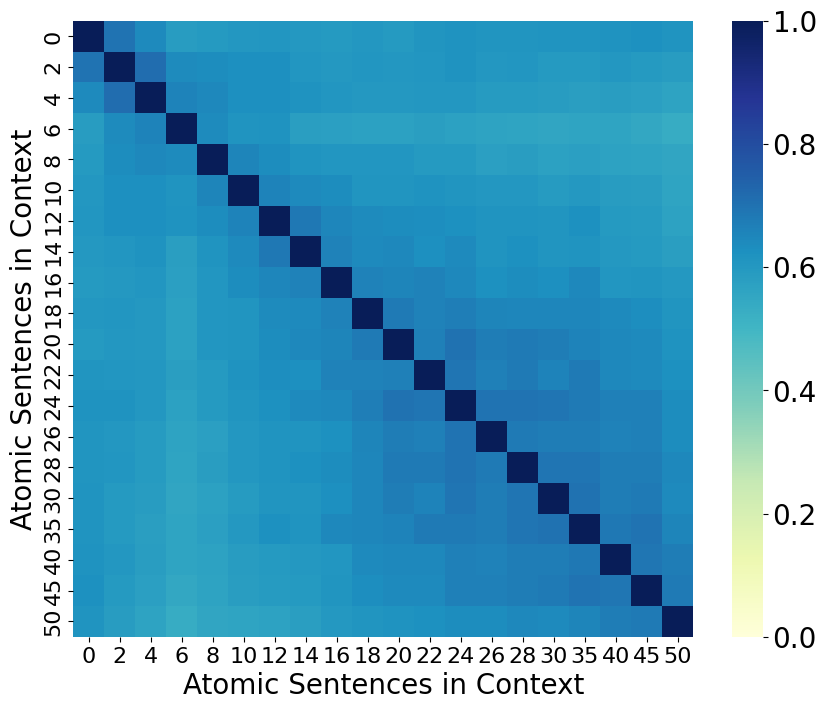}
        \caption{Mistral 8x22B global vs global}
        \label{fig:gvg_mixtral}
    \end{subfigure}
    \hfill
    \begin{subfigure}[t]{0.32\textwidth}
        \includegraphics[width=\textwidth]{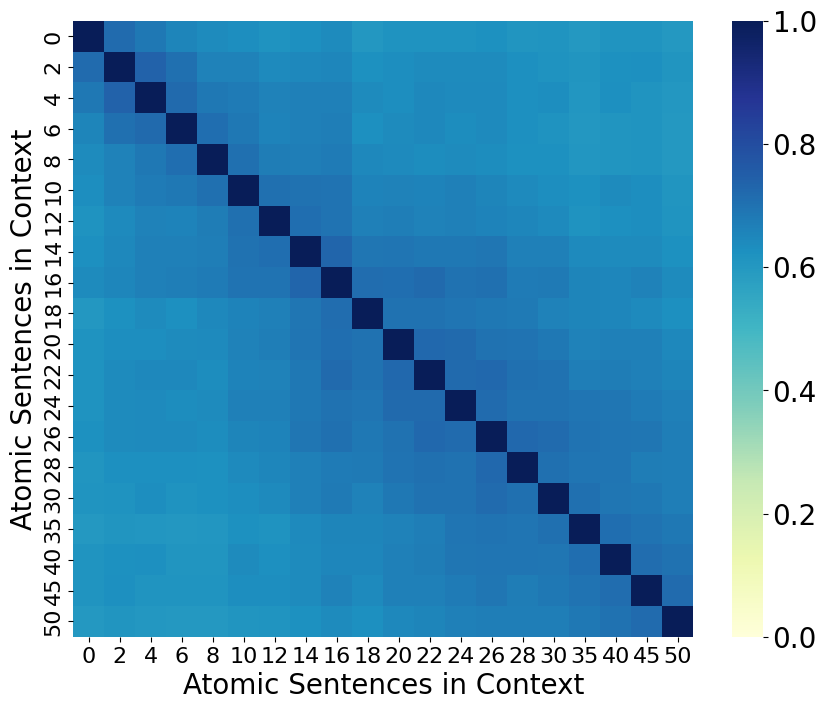}
        \caption{Mistral 7B global vs global}
        \label{fig:gvg_mistral}
    \end{subfigure}
    \hfill
    \begin{subfigure}[t]{0.32\textwidth}
        \includegraphics[width=\textwidth]{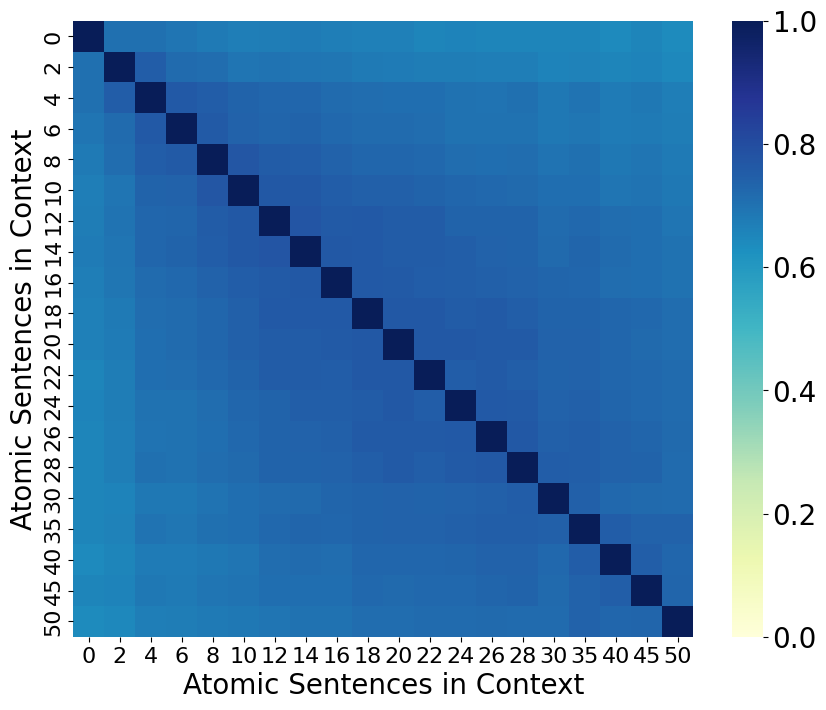}
        \caption{Phi global vs global}
        \label{fig:gvg_phi}
    \end{subfigure}
    \caption{Global vs Global Similarity Analysis for Various Models}
    \label{fig:global_global_similarity_appendix}
\end{figure*}

\section{Contextual vs. Parametric Knowledge in Response}
\label{app:local_global_similarity}
Figure \ref{fig:local_global_similarity_each_model} shows similarity score for contextual and parametric knowledge in each response across different model.

\begin{figure*}[t!]
    \centering
    \begin{subfigure}[t]{0.48\textwidth}
    \includegraphics[width=1\textwidth]{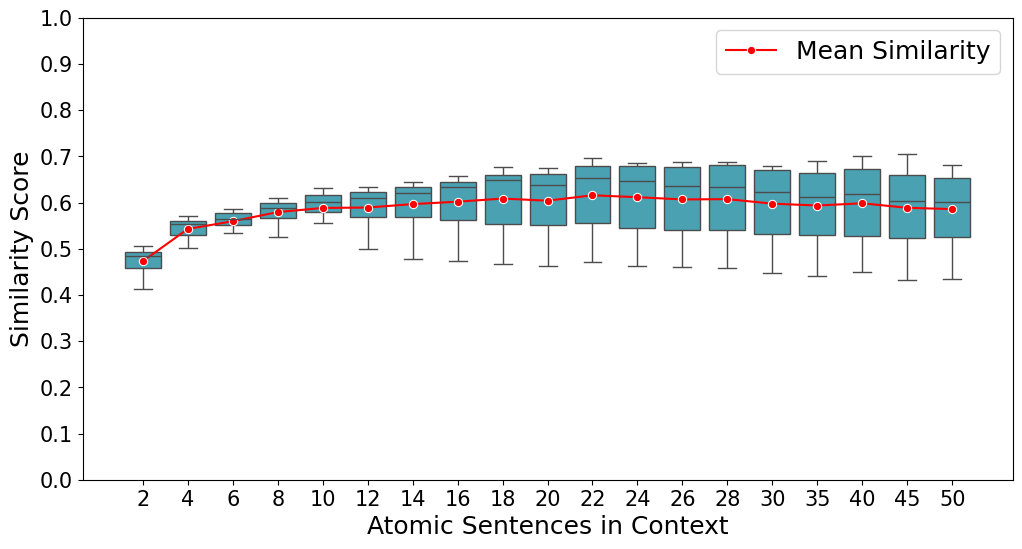}
    \caption{GPT-4o}
    \label{fig:lvg_gpt4o}
    \end{subfigure}
    \hfill
    \begin{subfigure}[t]{0.48\textwidth}
    \includegraphics[width=\textwidth]{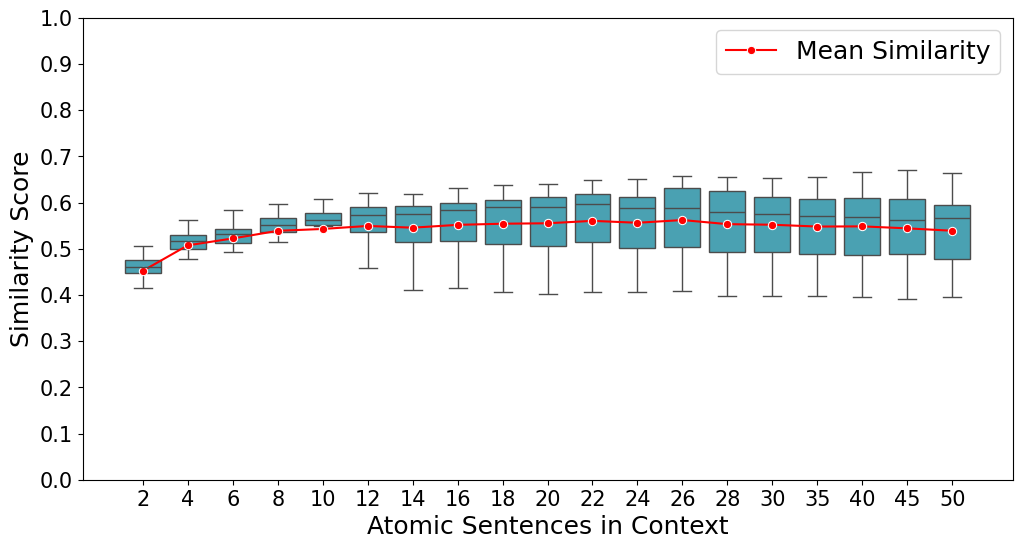}
    \caption{Llama 3 70B}
    \label{fig:lvg_llama70b}
    \end{subfigure}
    \hfill
    \begin{subfigure}[t]{0.48\textwidth}
    \includegraphics[width=\textwidth]{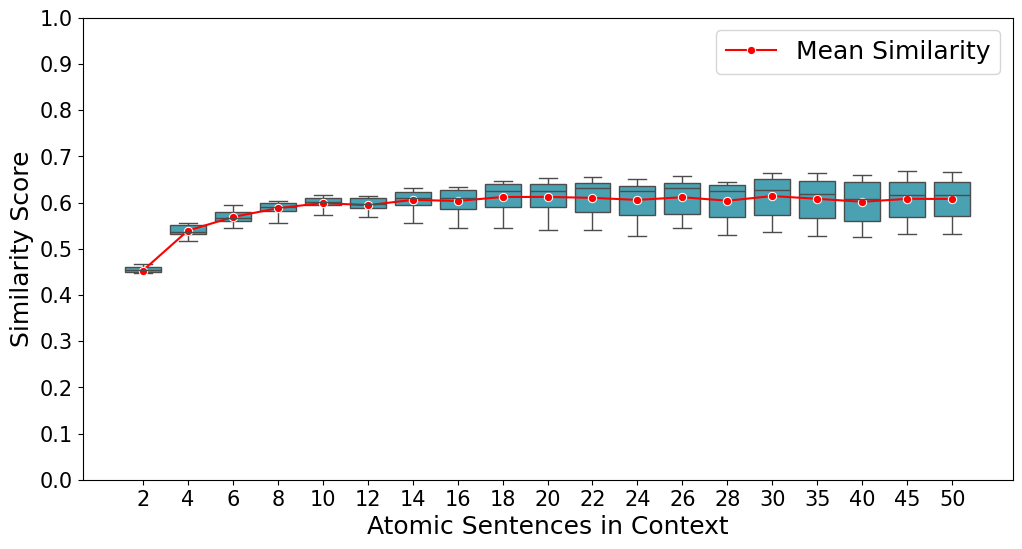}
    \caption{Llama 3 8B}
    \label{fig:lvg_llama7b}
    \end{subfigure}
    \hfill
    \begin{subfigure}[t]{0.48\textwidth}
    \includegraphics[width=\textwidth]{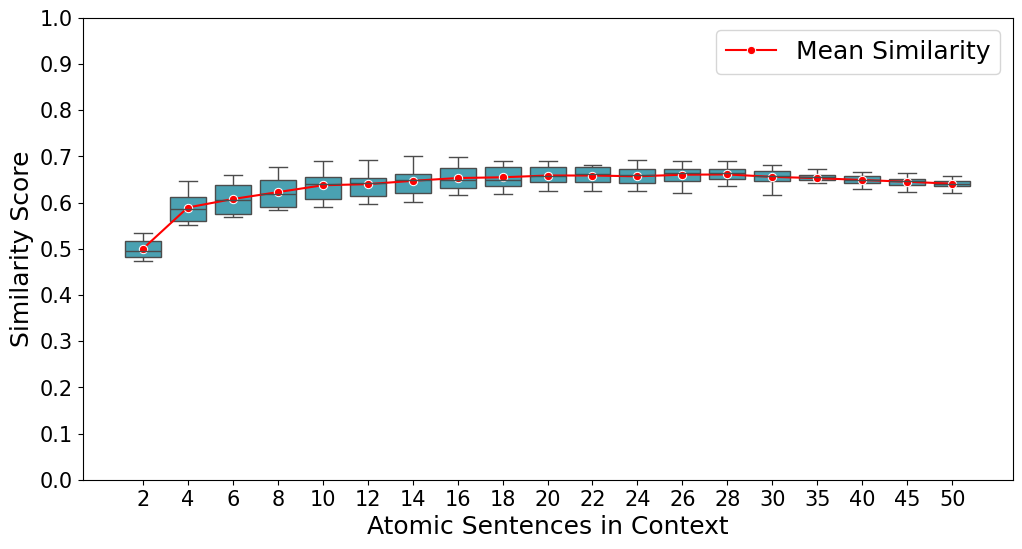}
    \caption{Claude Opus}
    \label{fig:lvg_opus}
    \end{subfigure}
    \hfill
    \begin{subfigure}[t]{0.48\textwidth}
    \includegraphics[width=\textwidth]{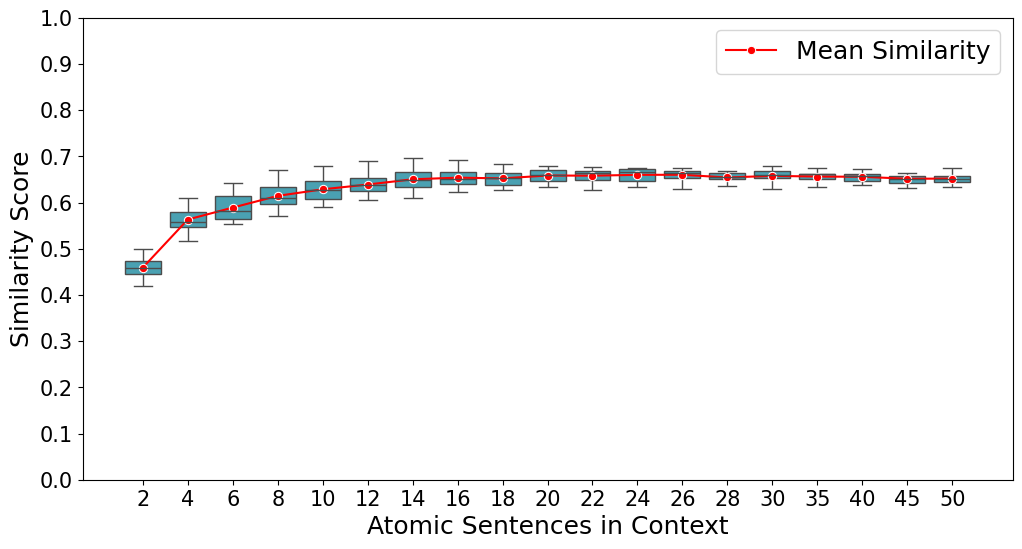}
    \caption{Claude Sonnet}
    \label{fig:lvg_sonnet}
    \end{subfigure}
    \hfill
    \begin{subfigure}[t]{0.48\textwidth}
    \includegraphics[width=\textwidth]{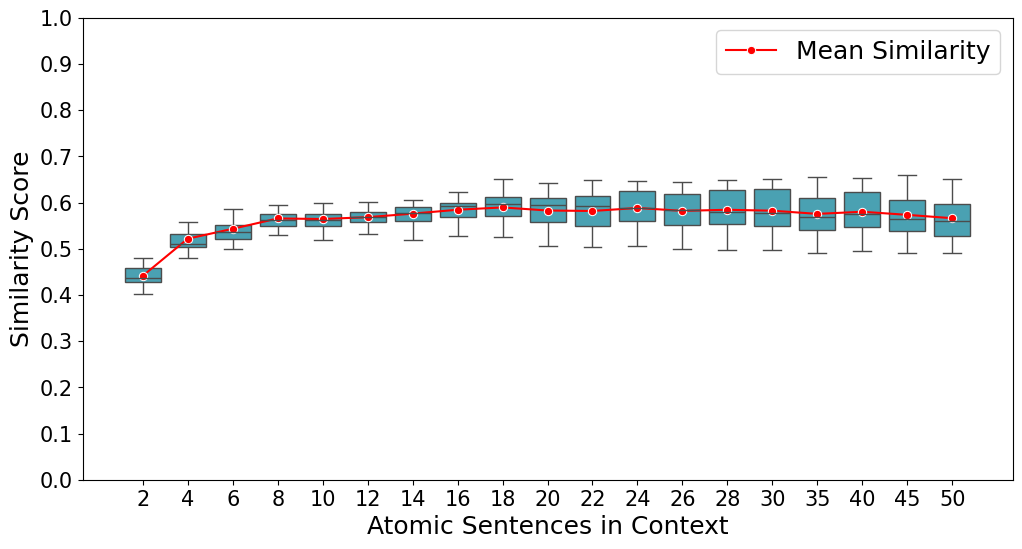}
    \caption{Claude Haiku}
    \label{fig:lvg_haiku}
    \end{subfigure}
    \hfill
    \begin{subfigure}[t]{0.48\textwidth}
    \includegraphics[width=\textwidth]{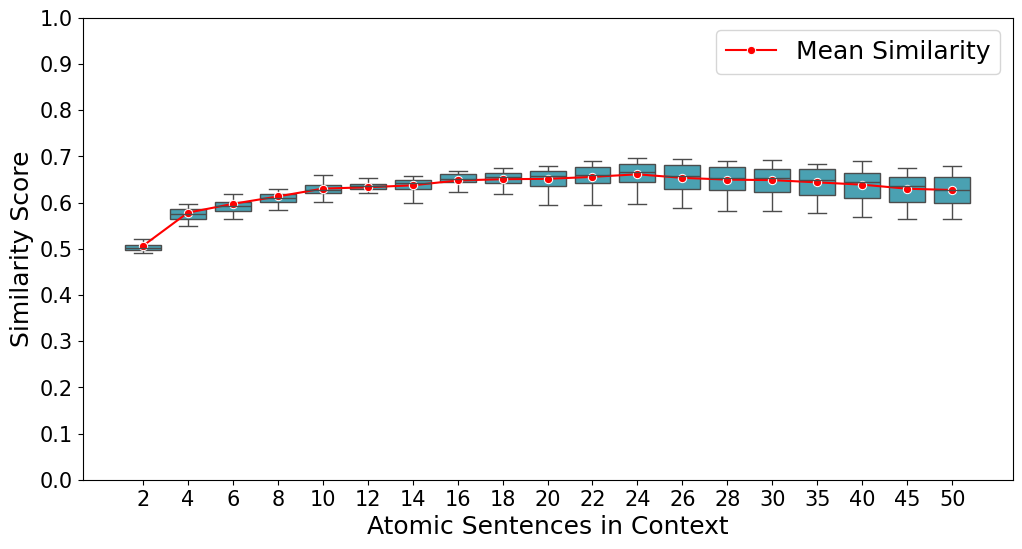}
    \caption{Mixtral 8x22B}
    \label{fig:lvg_mixtral}
    \end{subfigure}
    \hfill
    \begin{subfigure}[t]{0.48\textwidth}
    \includegraphics[width=\textwidth]{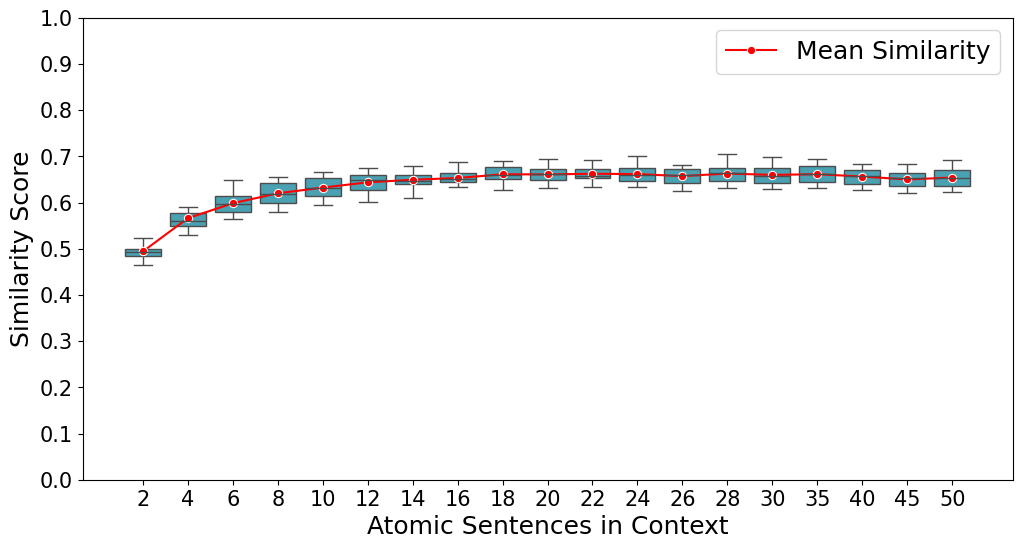}
    \caption{Mixtral 7B}
    \label{fig:lvg_mistral}
    \end{subfigure}
    \hfill
    \begin{subfigure}[t]{0.46\textwidth}
    \includegraphics[width=\textwidth]{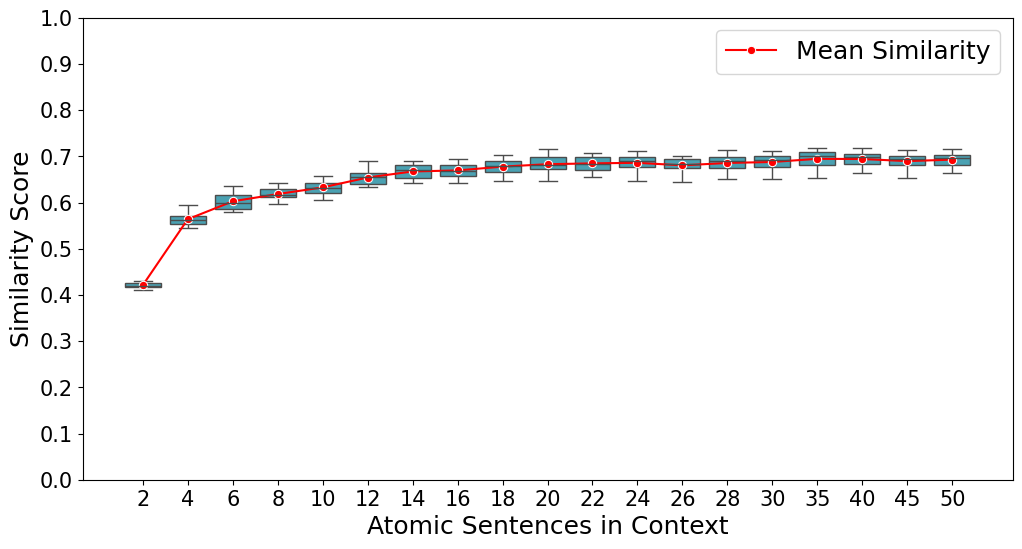}
    \caption{Phi 3}
    \label{fig:lvg_phi}
    \end{subfigure} 
    \caption{Similarity between contextual knowledge and parametric knowledge within a set of example for each model}
    \label{fig:local_global_similarity_each_model}
\end{figure*}

\section{FactScore False Parametric Knowledge}
\label{app:false_claims}
Figure \ref{fig:large_false} show number of false parametric knowledge for all models.

\begin{figure*}[t!]
    \centering
    \begin{subfigure}[t]{0.49\textwidth}
    \includegraphics[width=\textwidth]{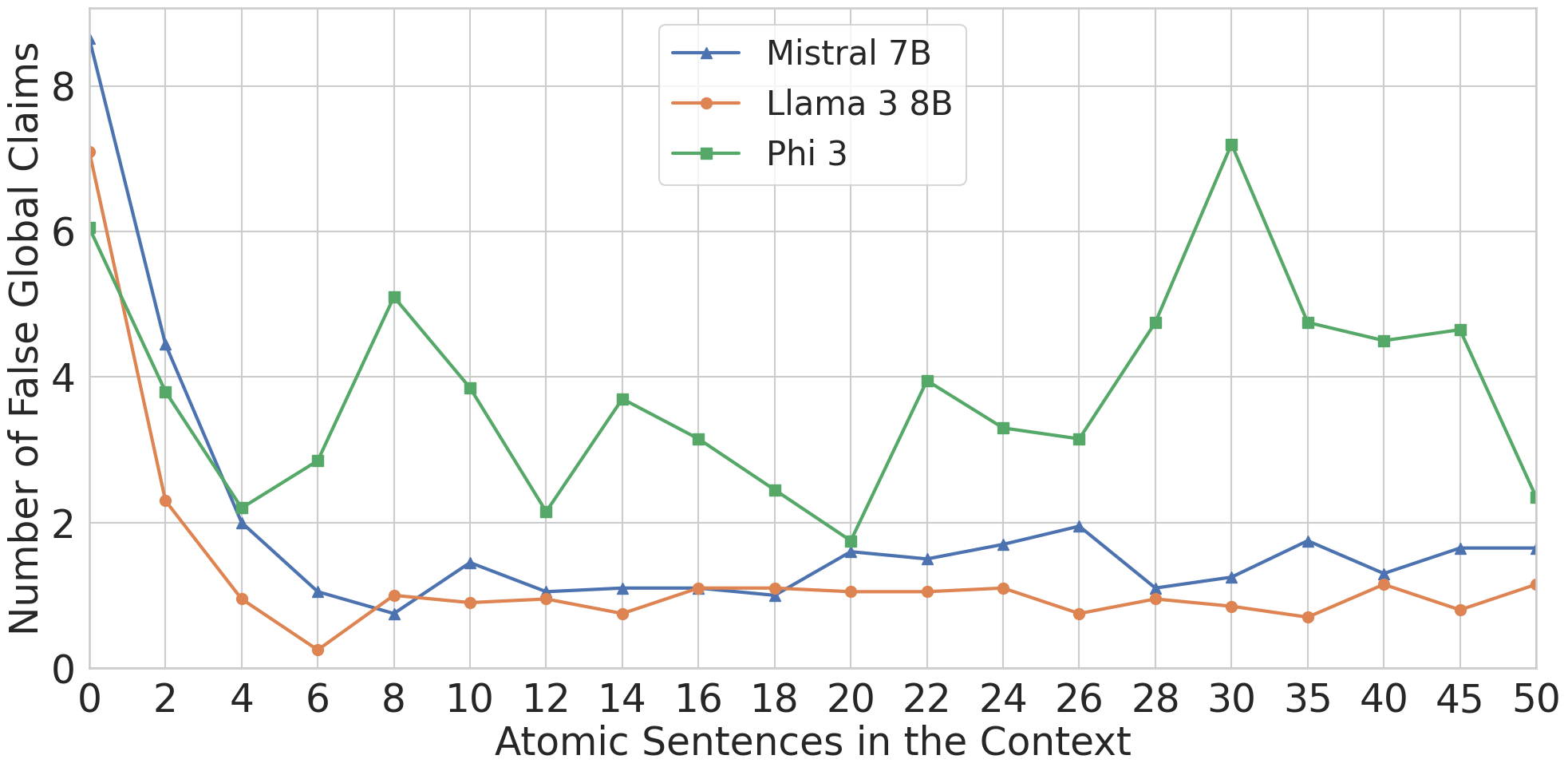}
    \caption{Number of false parametric knowledge for Mistal 7B, Llama 3 8B and Phi 3}
    \label{fig:small_false}
    \end{subfigure}
    \hfill
    \begin{subfigure}[t]{0.49\textwidth}
    \includegraphics[width=\textwidth]{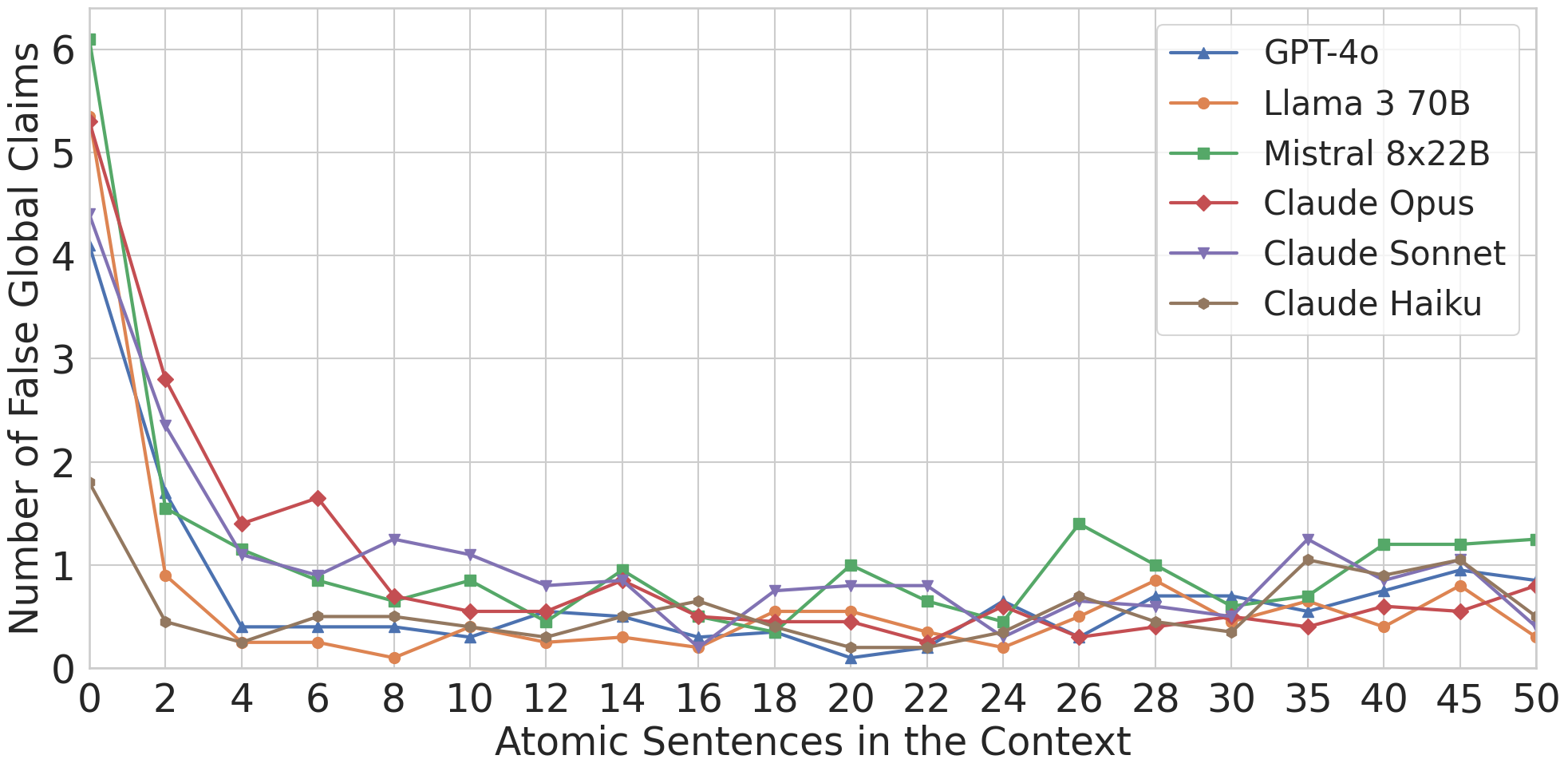}
    \caption{Number of false parametric knowledge for GPT-4o, Llama 3 70B, Mistral 8x22B and Claude Opus, Sonnet and Haiku}
    \label{fig:larg_factscore}
    \end{subfigure}
    \caption{Number of false parametric knowledge across all LLMs}
    \label{fig:large_false}
\end{figure*}

% \section{Response Atomization Prompt}
% \label{app:response_atomization}
% The Figure~\ref{fig:response_atomization} shows the exact prompt we used to extract atomic facts from different LLMs response. 

% \section{Contextual/Parametric Knowledge Evaluation Prompt}
% \label{app:evaluation}
% The Figure~\ref{fig:evaluation_prompt} shows the detailed prompt we used to evaluate local contexts and global facts. These two are two lists. In this version, we found the GPT-4o performs much better when we pass in each of the atomic local context and the global facts list to evaluate compare to pass in the whole two lists. When we passed in two lists at the same time, the model sometimes skipped some data caused the output lists do not add up to total number of atomic facts in the answer. 
% \begin{figure*}[t!]
%     \centering
%     \includegraphics[width=\textwidth]{figures/evaluation_prompt.png}
%     \caption{Local/Global Facts Evaluation Prompt}
%     \label{fig:evaluation_prompt}
% \end{figure*}

\section{New knowledge}
\label{app:new_knowledge}

Figure \ref{fig:new_claude_gpt4o}, \ref{fig:new_llama3_mistralai} show how each behave when asking question about a new information. 

% \begin{figure*}[t!]
%     \centering
%     \includegraphics[width=0.6\textwidth]{figures/new_knowledge/new_gpt4o.png}
%     \caption{New knowledge (GPT-4o)}
%     \label{fig:new_gpt4o}
% \end{figure*} 

\begin{figure*}[t!]
    \centering
    \begin{subfigure}[t]{0.49\textwidth}
    \includegraphics[width=\textwidth]{figures/new_knowledge/new_gpt4o.png}
    \caption{New knowledge (GPT-4o)}
    \label{fig:new_gpt4o}
    \end{subfigure}
    \hfill
    \begin{subfigure}[t]{0.49\textwidth}
    \includegraphics[width=\textwidth]{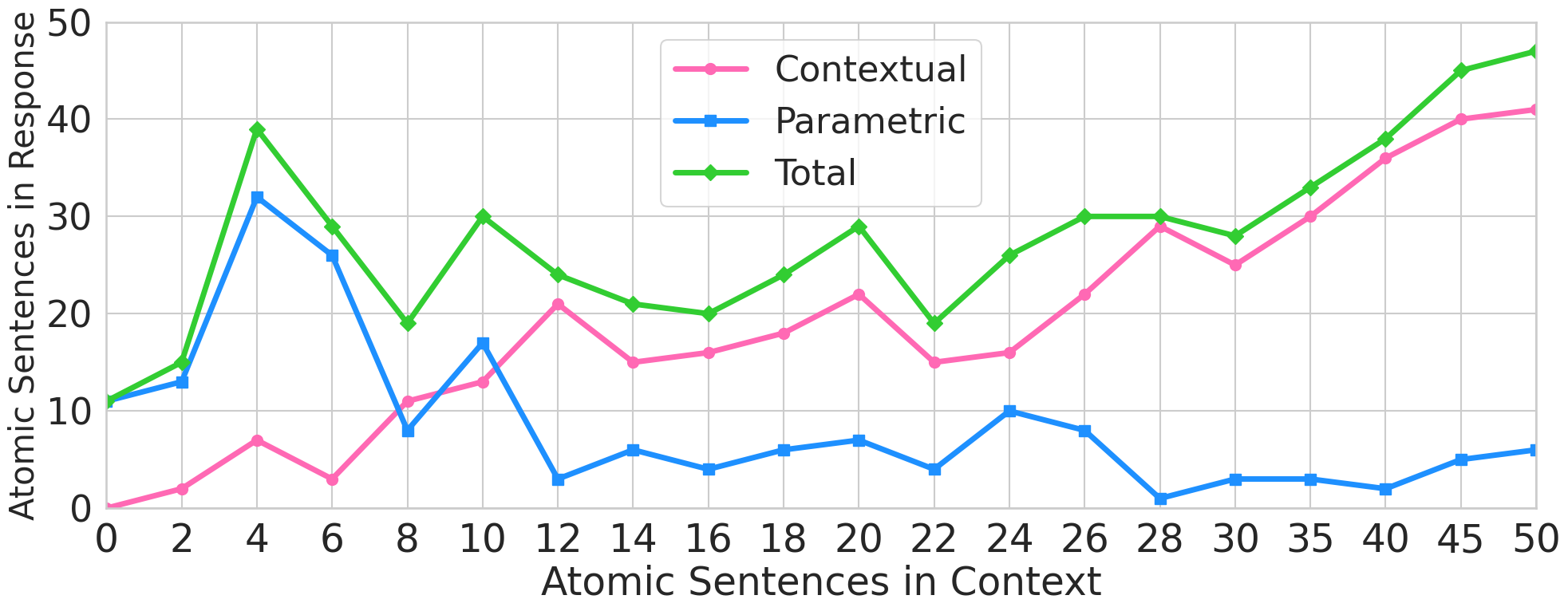}
    \caption{New knowledge (Opus)}
    \label{fig:new_opus}
    \end{subfigure}
    \hfill
    \begin{subfigure}[t]{0.49\textwidth}
    \includegraphics[width=\textwidth]{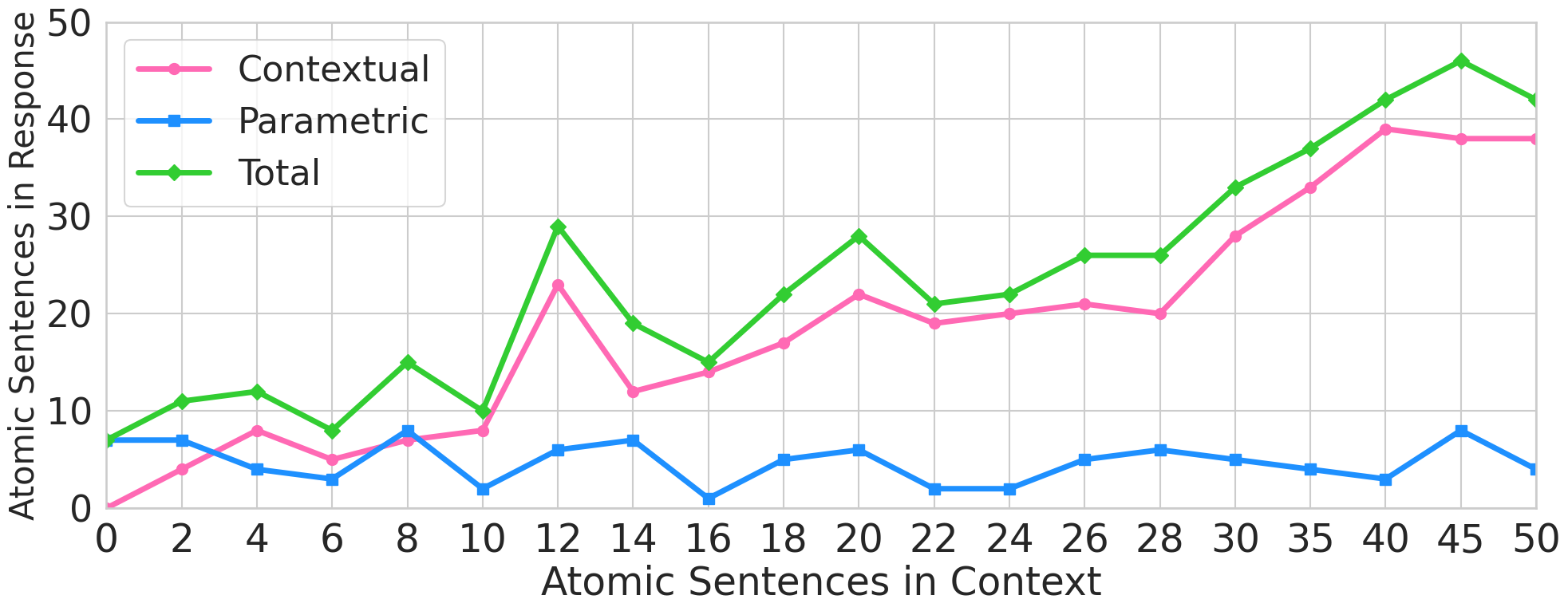}
    \caption{New knowledge (Sonnet)}
    \label{fig:new_sonnet}
    \end{subfigure}
    \hfill
    \begin{subfigure}[t]{0.49\textwidth}
    \includegraphics[width=\textwidth]{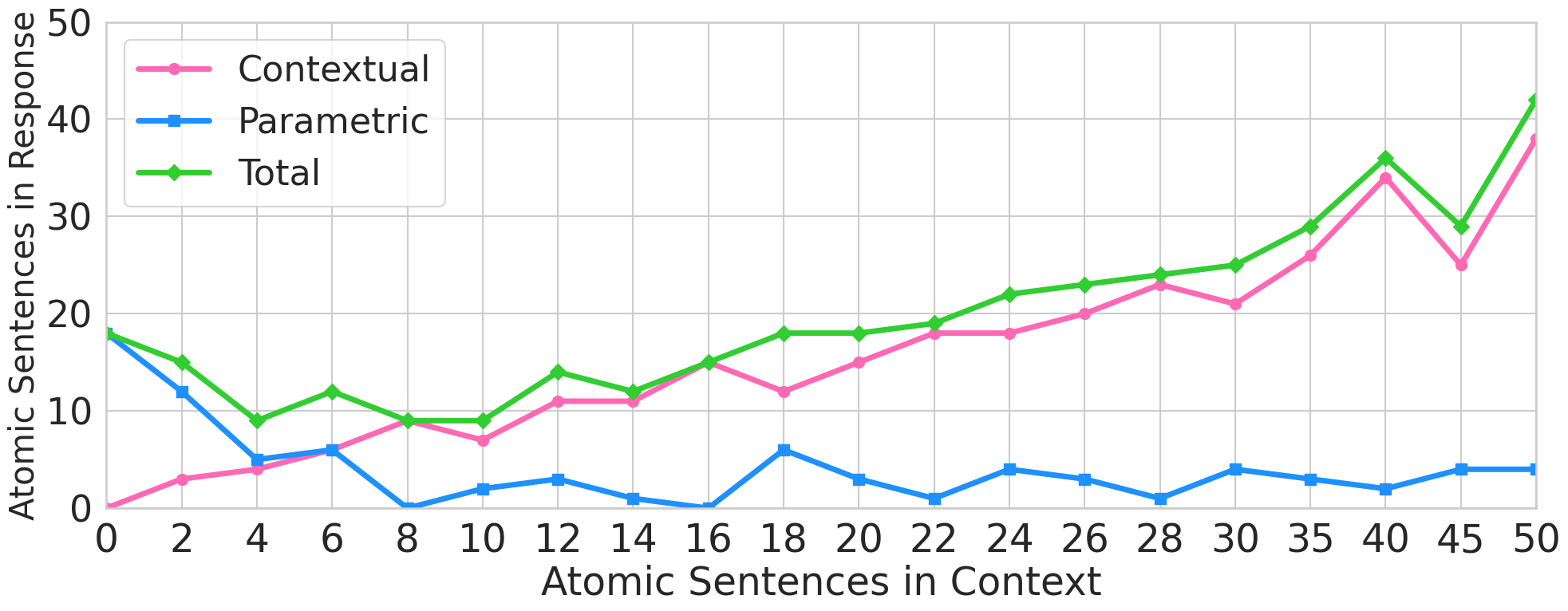}
    \caption{New knowledge (Haiku)}
    \label{fig:new_haiku}
    \end{subfigure}
    \caption{New knowledge (All Claude models)}
    \label{fig:new_claude_gpt4o}
\end{figure*} 

\begin{figure*}[t!]
    \centering
    \begin{subfigure}[t]{0.49\textwidth}
    \includegraphics[width=\textwidth]{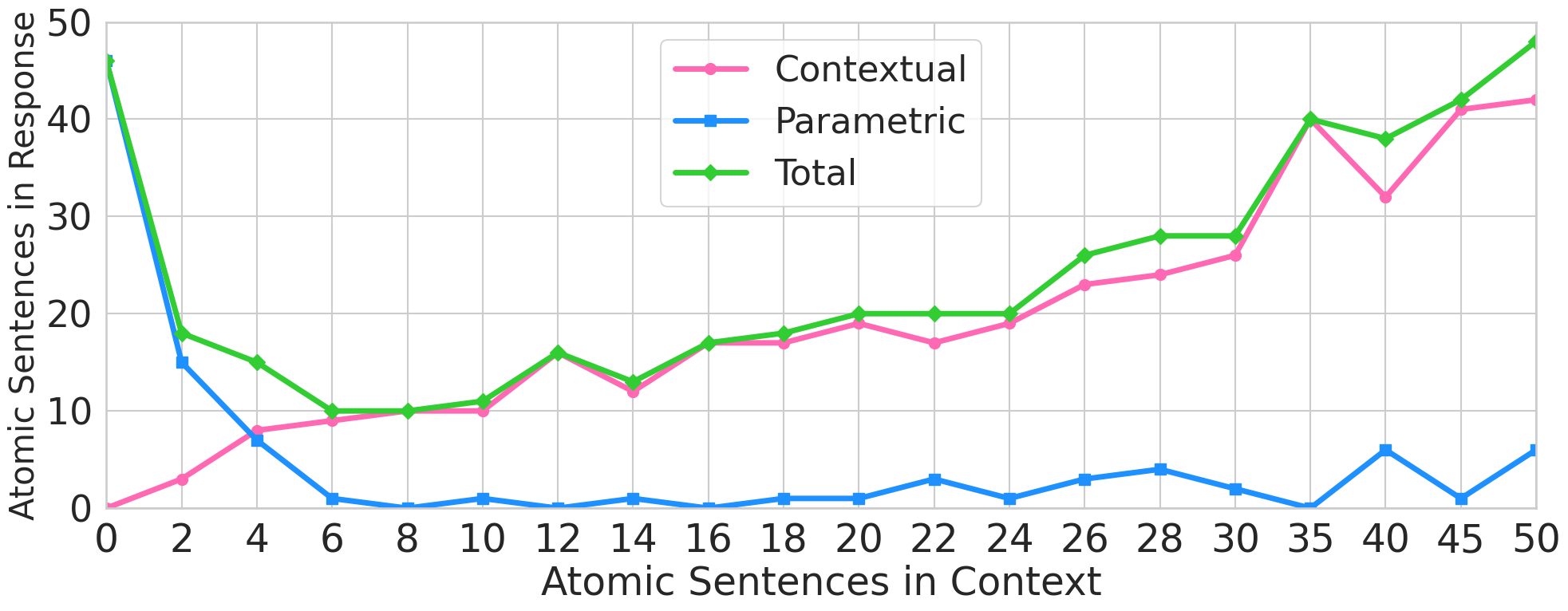}
    \caption{New knowledge (Llama 3 70B)}
    \label{fig:new_llama370b}
    \end{subfigure}
    \hfill
    \begin{subfigure}[t]{0.49\textwidth}
    \includegraphics[width=\textwidth]{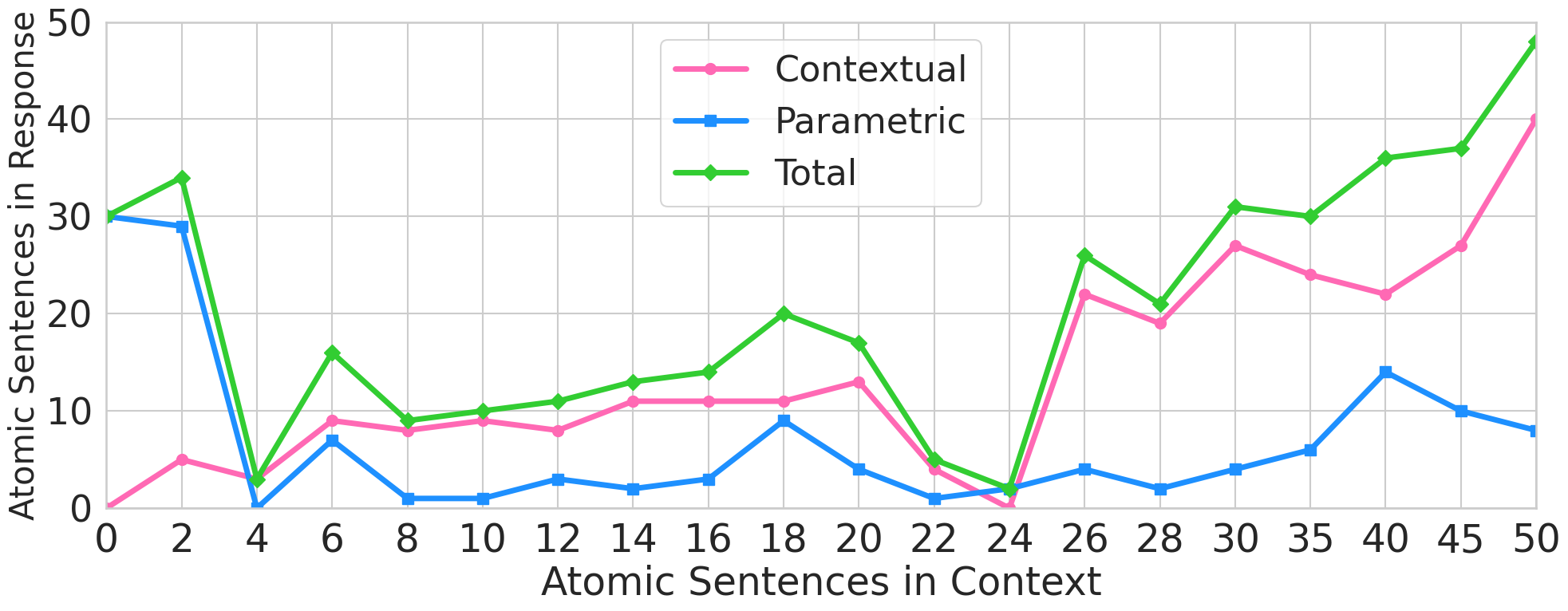}
    \caption{New knowledge (Llama 3 8B)}
    \label{fig:new_llama38b}
    \end{subfigure}
    \begin{subfigure}[t]{0.49\textwidth}
    \includegraphics[width=\textwidth]{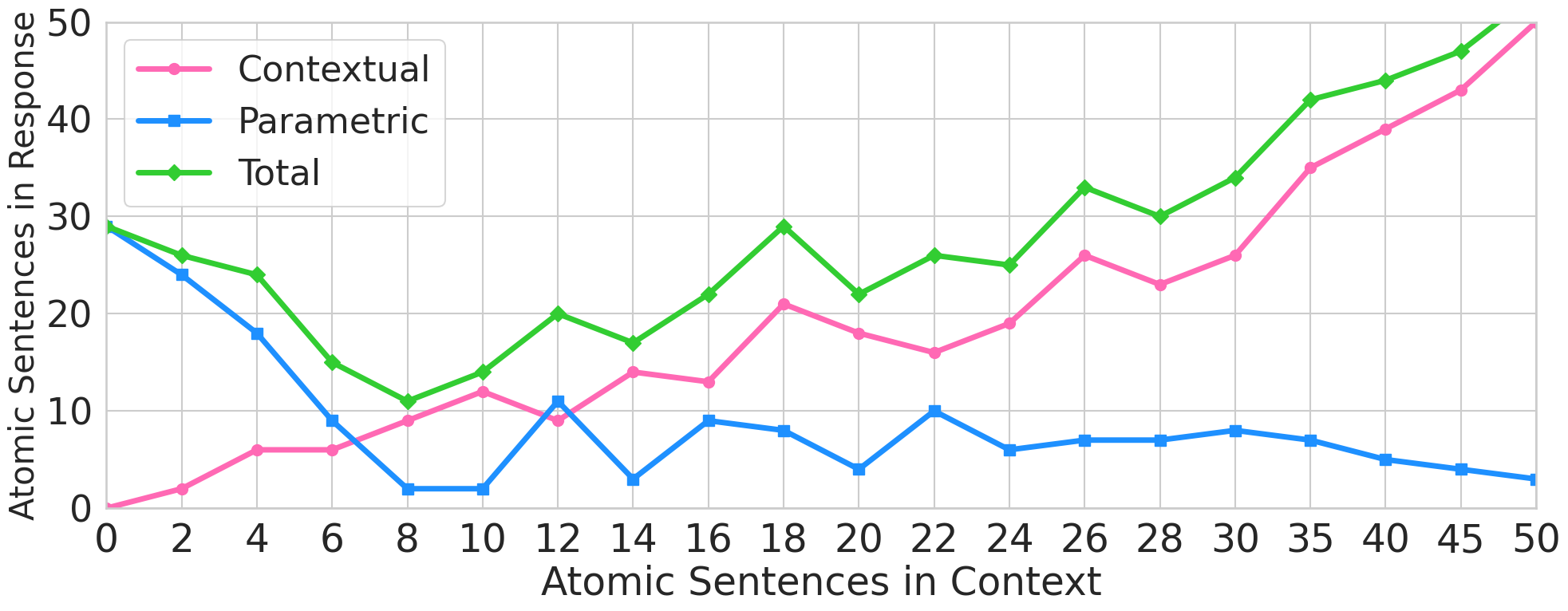}
    \caption{New knowledge (Mixtral 8x22B)}
    \label{fig:new_mixtral}
    \end{subfigure}
    \hfill
    \begin{subfigure}[t]{0.49\textwidth}
    \includegraphics[width=\textwidth]{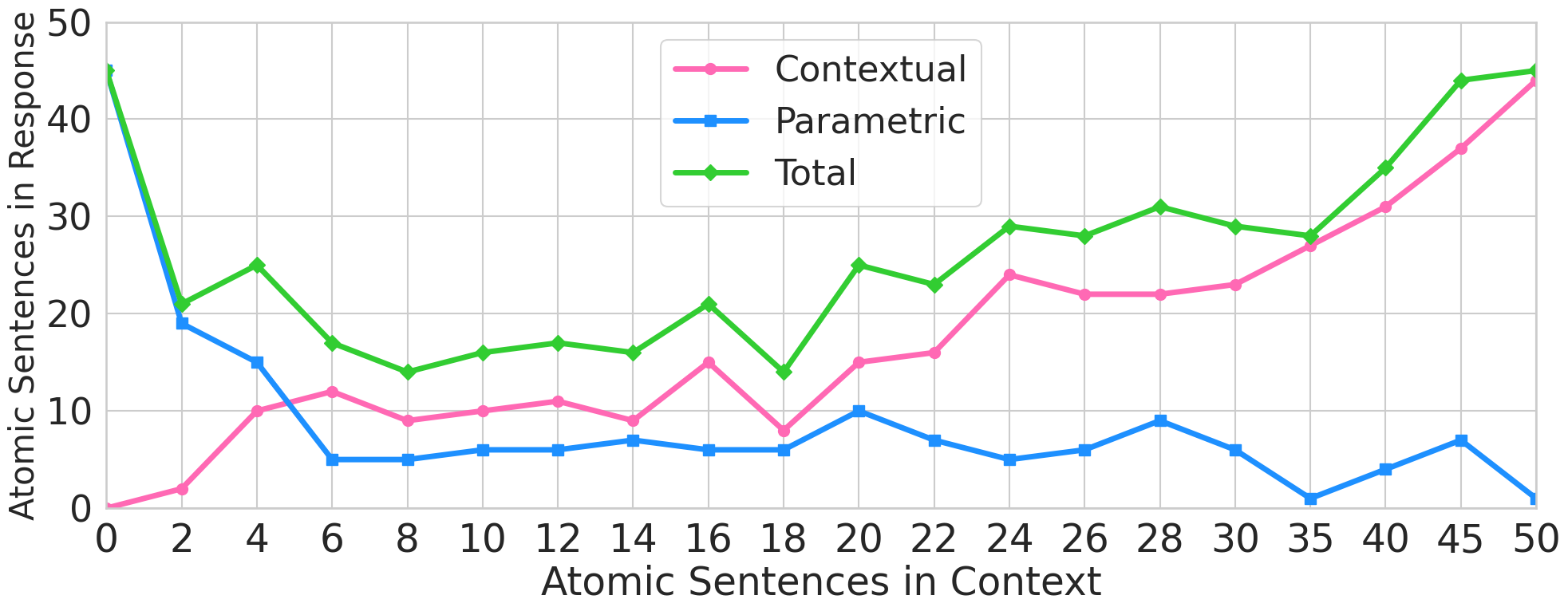}
    \caption{New knowledge (Mistral 7B)}
    \label{fig:new_mistral}
    \end{subfigure}
    \begin{subfigure}[t]{0.6\textwidth}
    \includegraphics[width=\textwidth]{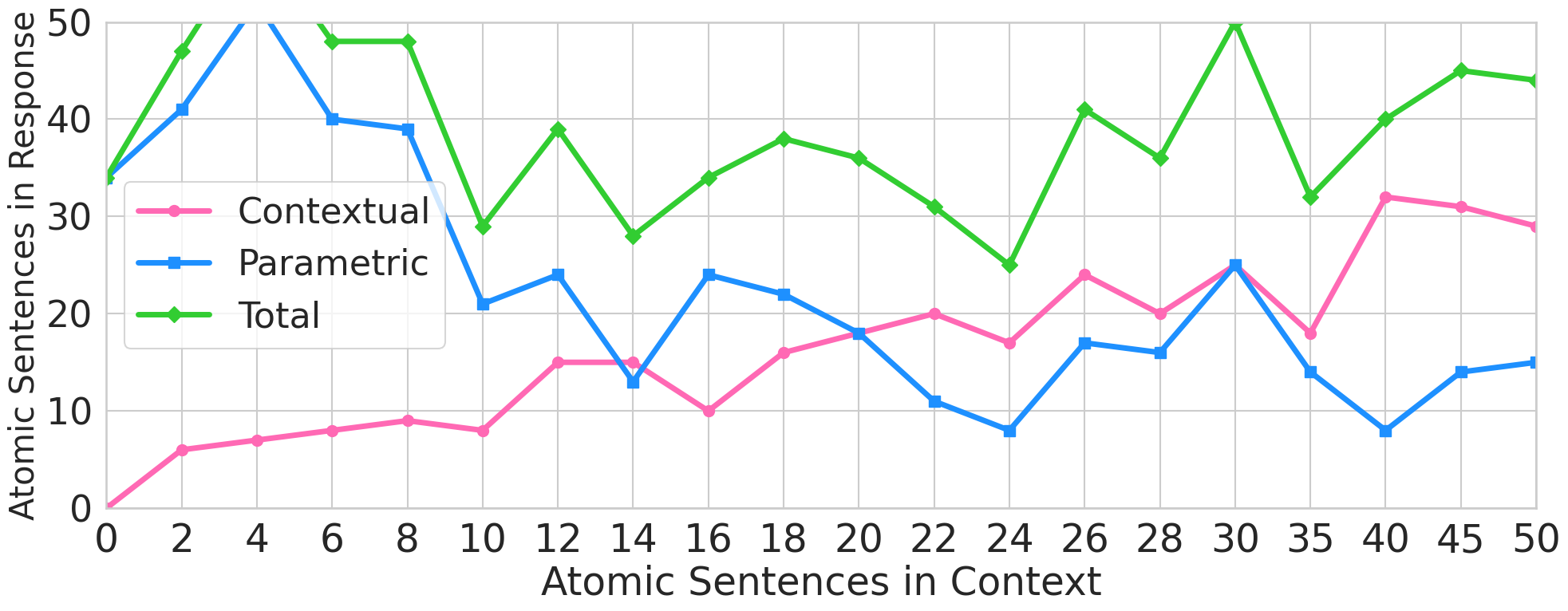}
    \caption{New knowledge (Phi 3)}
    \label{fig:new_phi}
    \end{subfigure}
    \caption{New knowledge (Phi 3, Mistral AI and Llama 3 models)}
    \label{fig:new_llama3_mistralai}
\end{figure*}

\section{Further Analyses}
\label{app:ablation}
This section shows the difference in the model's use of local contexts when using a strict vs unrestricted prompt (Figure~\ref{fig:ablation_gpt4o}, \ref{fig:ablation_claude} and \ref{fig:ablation_mistral}). The strict prompt instructs the model to only use the provided context when answering the question. The unrestricted prompt provides the context but doesn't instruct the model to use it in any way.

%Figure \ref{fig:ablation_gpt4o}, \ref{fig:ablation_claude} and \ref{fig:ablation_mistral} show the results when the models are prompted with the unrestricted prompt which provides the contexts and only instructs: "Tell me about <topic>" 
%Figure... show the results of the strict version when the models are prompted with "Using only provided contexts, tell me about <topic>" 

\begin{figure*}[t!]
    \centering
    \begin{subfigure}[t]{0.49\textwidth}
    \includegraphics[width=\textwidth]{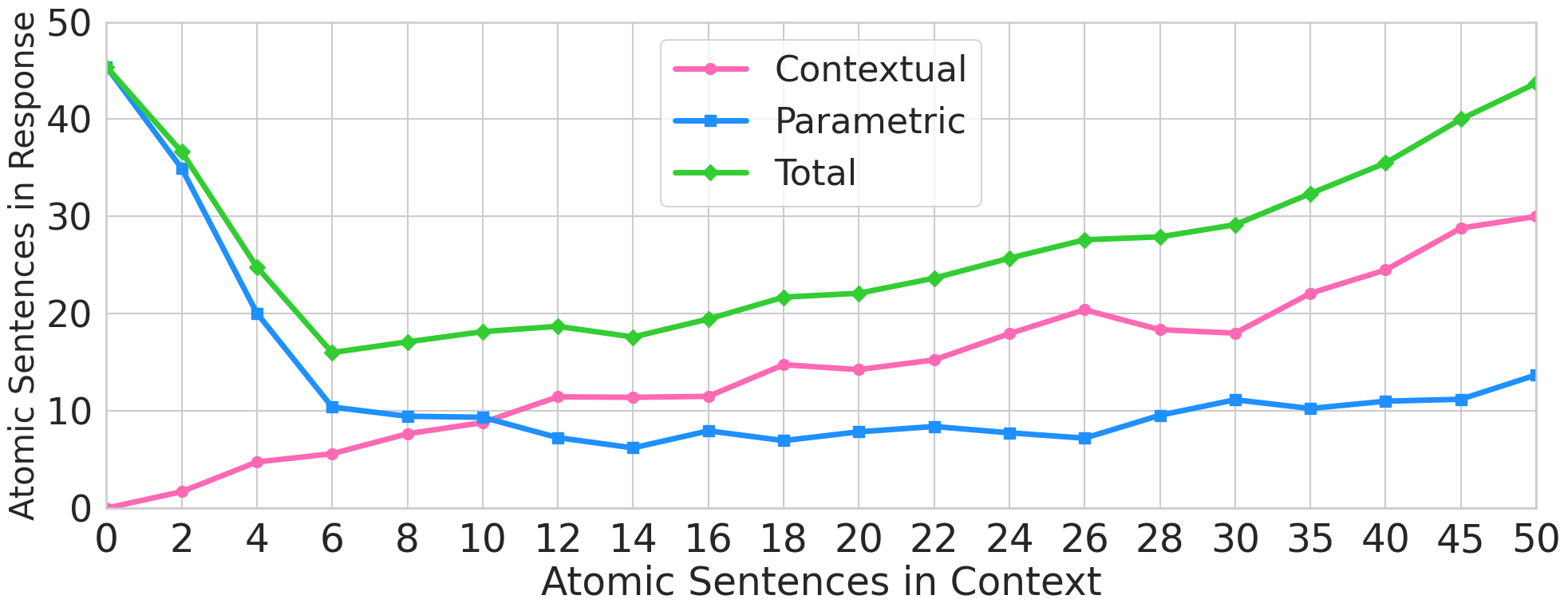}
    \caption{No restriction in question (GPT-4o)}
    \label{fig:open_gpt4o}
    \end{subfigure}
    \hfill
    \begin{subfigure}[t]{0.49\textwidth}
    \includegraphics[width=\textwidth]{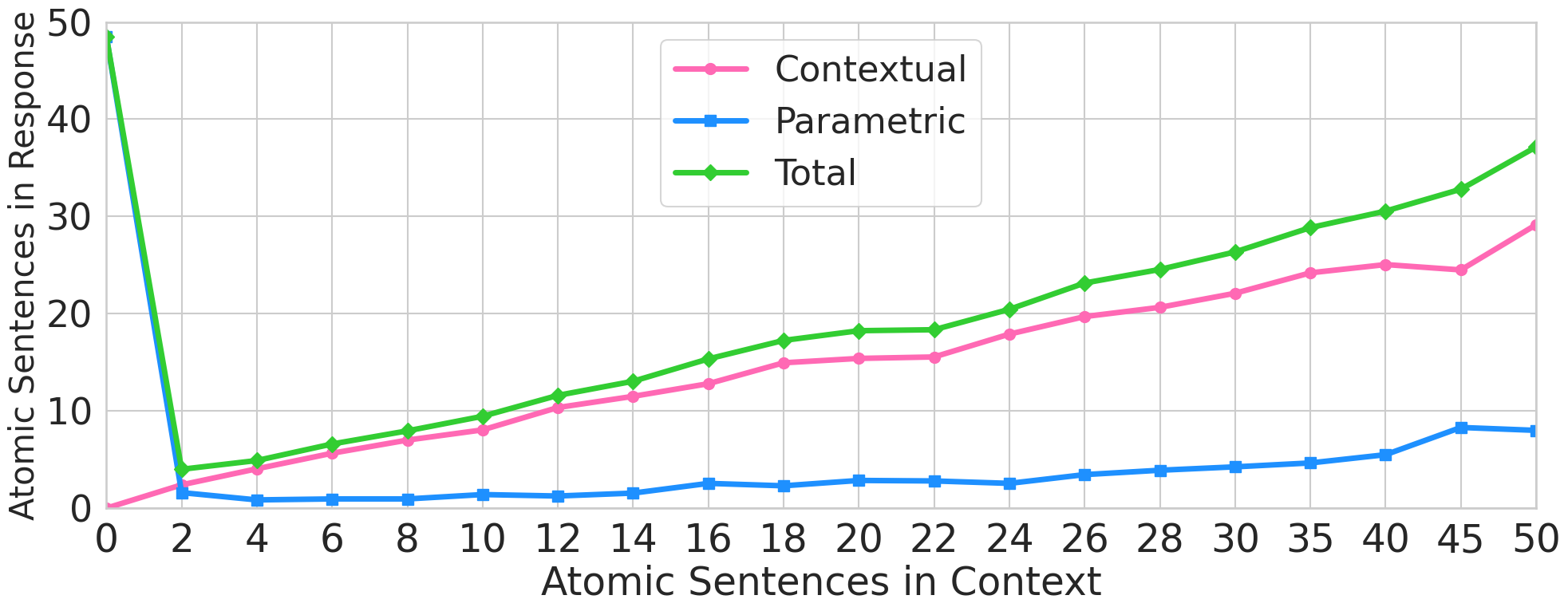}
    \caption{Strict restriction in question (GPT-4o)}
    \label{fig:strict_gpt4o}
    \end{subfigure}
    \caption{Comparing two methods of prompting the model to answer, one instructing to strictly adhere to provided context and the other not imposing any restriction.}
    \label{fig:ablation_gpt4o}
\end{figure*} 

\begin{figure*}[t!]
    \centering
    \begin{subfigure}[t]{0.49\textwidth}
    \includegraphics[width=\textwidth]{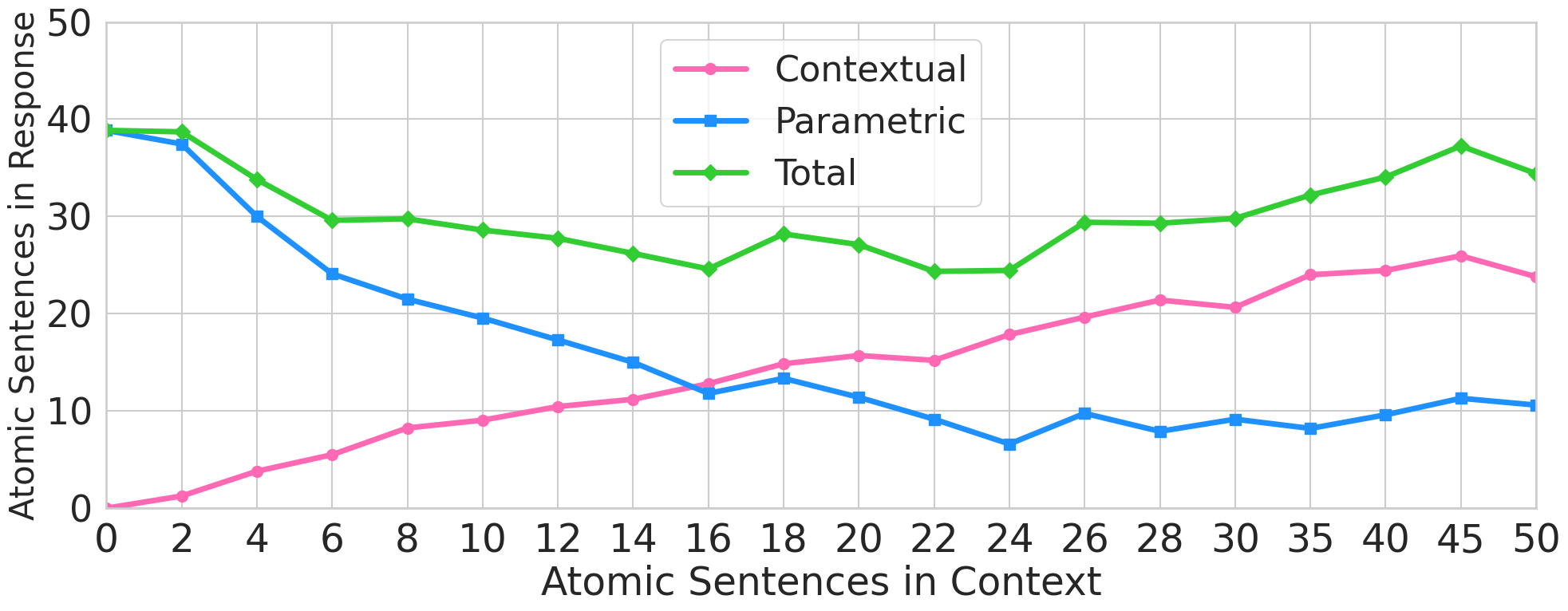}
    \caption{No restriction in question (Opus)}
    \label{fig:open_opus}
    \end{subfigure}
    \hfill
    \begin{subfigure}[t]{0.49\textwidth}
    \includegraphics[width=\textwidth]{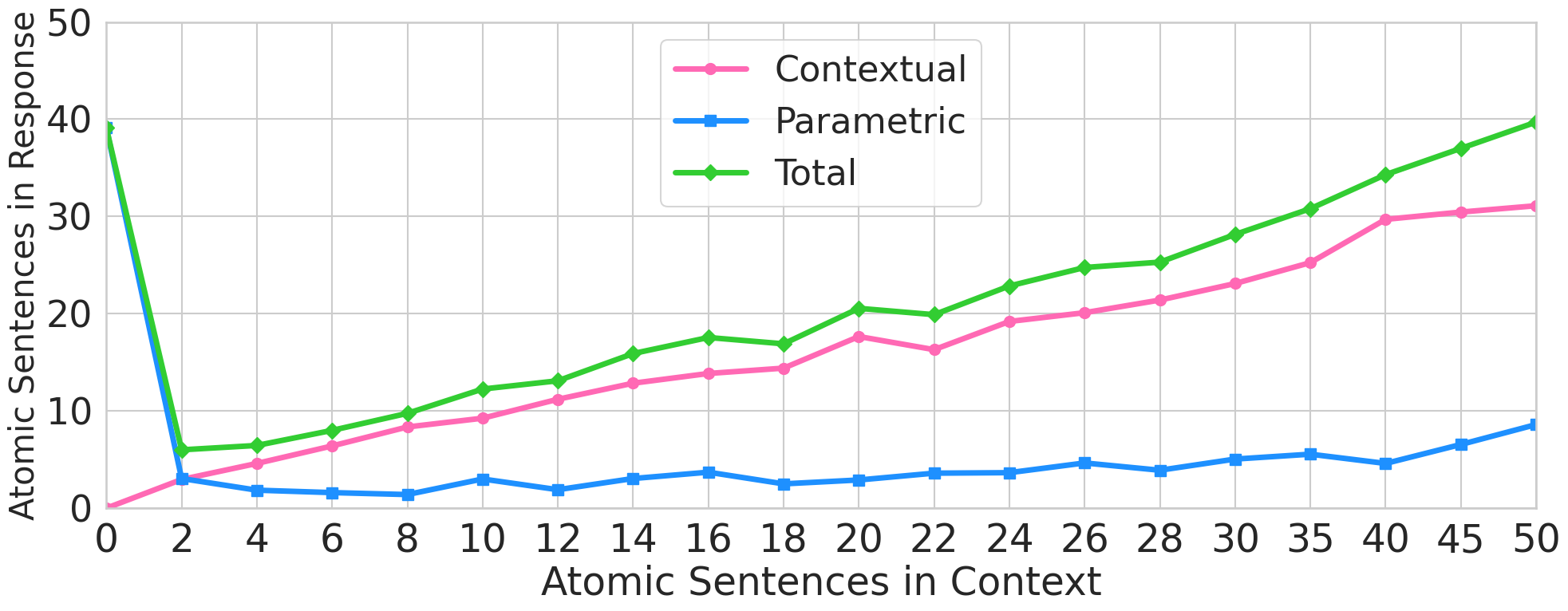}
    \caption{Strict restriction in question (Opus)}
    \label{fig:strict_opus}
    \end{subfigure}
    \begin{subfigure}[t]{0.49\textwidth}
    \includegraphics[width=\textwidth]{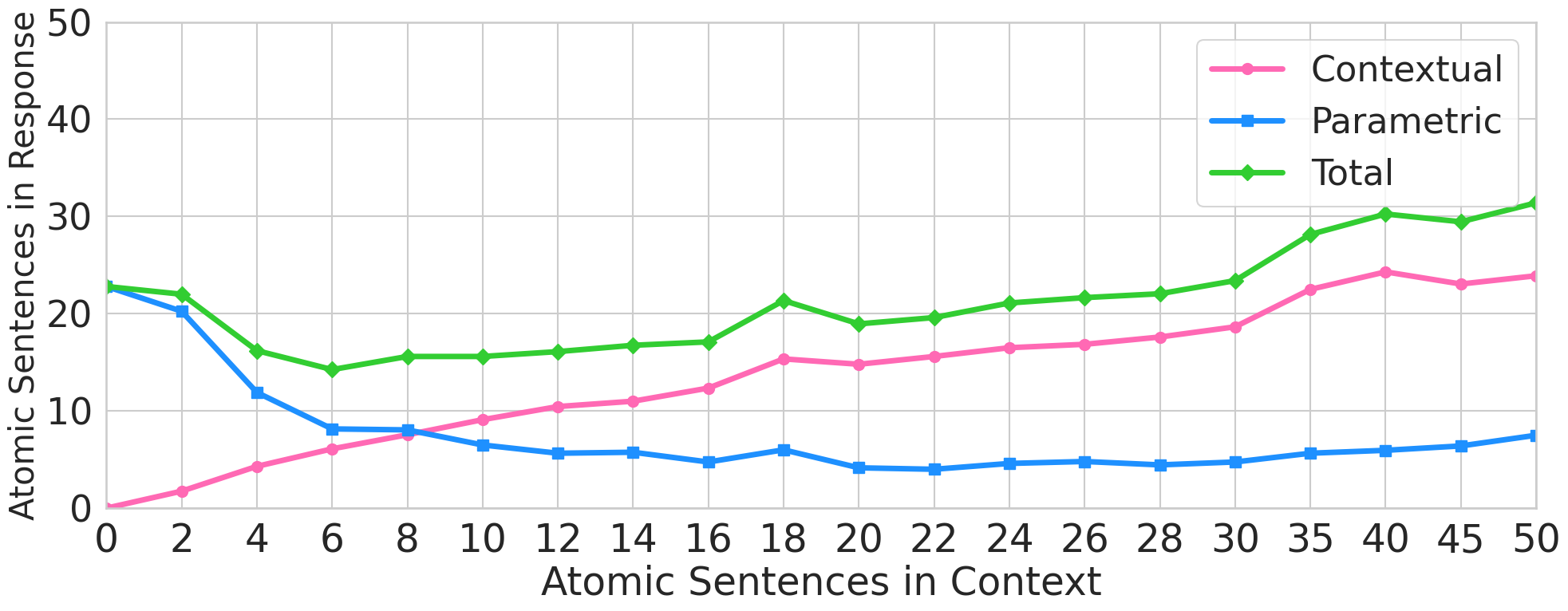}
    \caption{No restriction in question (Sonnet)}
    \label{fig:open_sonnet}
    \end{subfigure}
    \hfill
    \begin{subfigure}[t]{0.49\textwidth}
    \includegraphics[width=\textwidth]{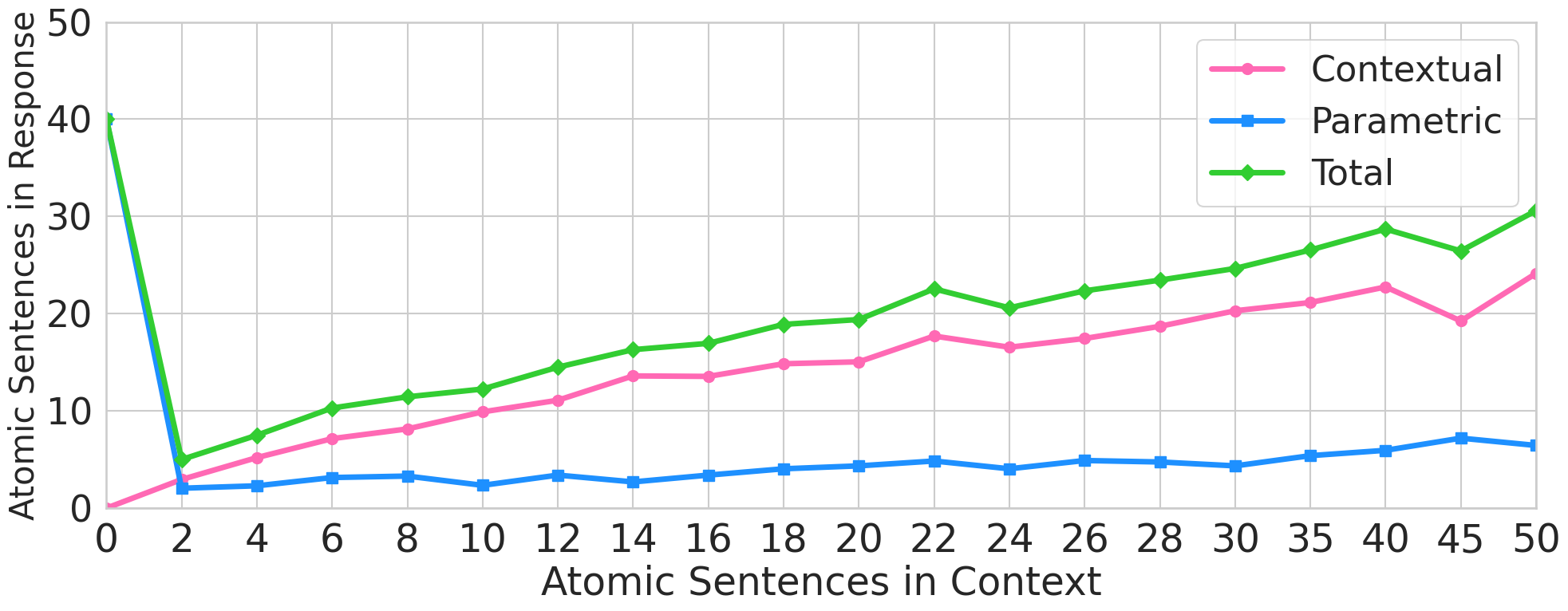}
    \caption{Strict restriction in question (Sonnet)}
    \label{fig:strict_sonnet}
    \end{subfigure}
        \begin{subfigure}[t]{0.49\textwidth}
    \includegraphics[width=\textwidth]{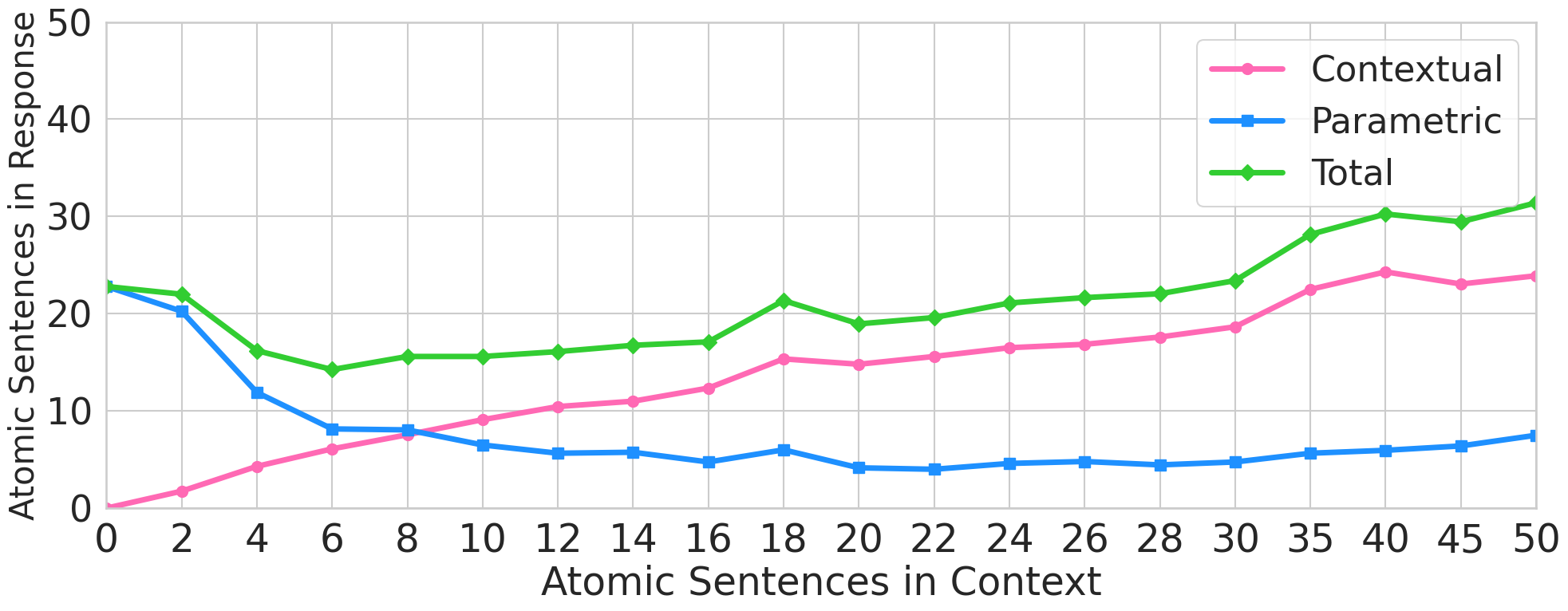}
    \caption{No restriction in question (Haiku)}
    \label{fig:open_haiku}
    \end{subfigure}
    \hfill
    \begin{subfigure}[t]{0.49\textwidth}
    \includegraphics[width=\textwidth]{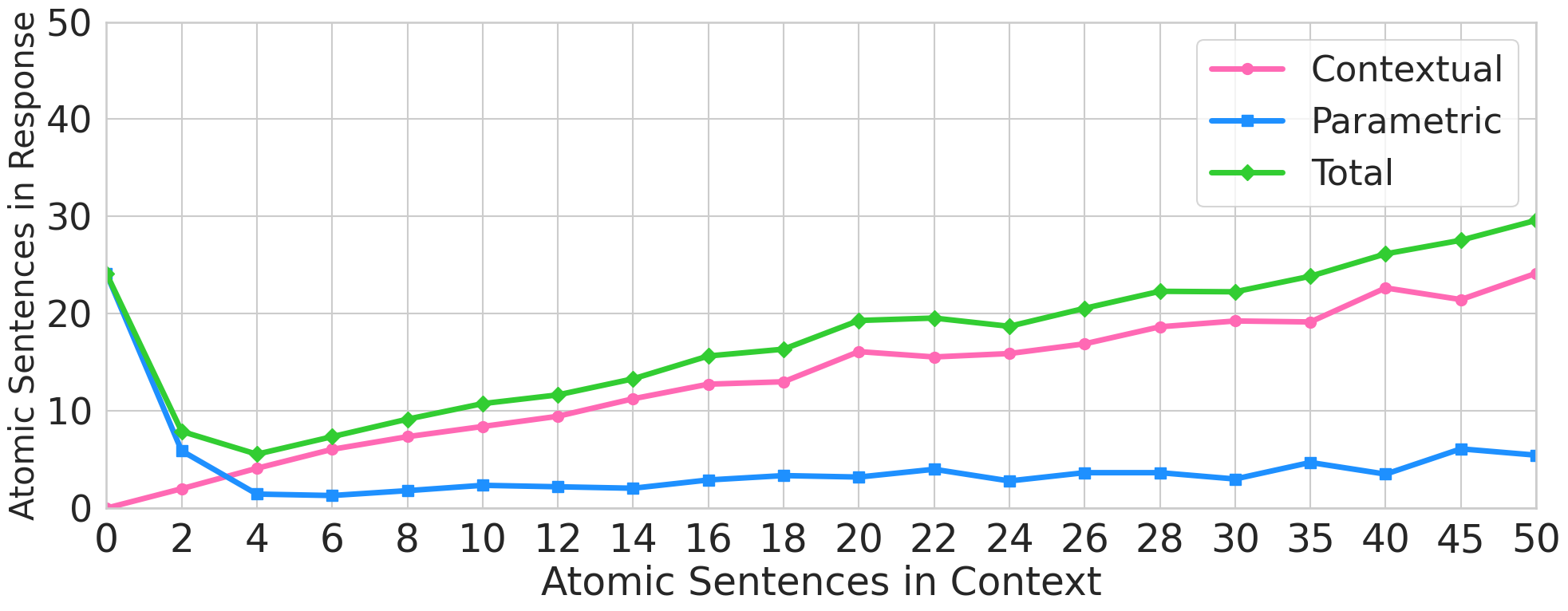}
    \caption{Strict restriction in question (Haiku)}
    \label{fig:strict_haiku}
    \end{subfigure}
    \hfill
    \begin{subfigure}[t]{0.48\textwidth}
    \includegraphics[width=\textwidth]{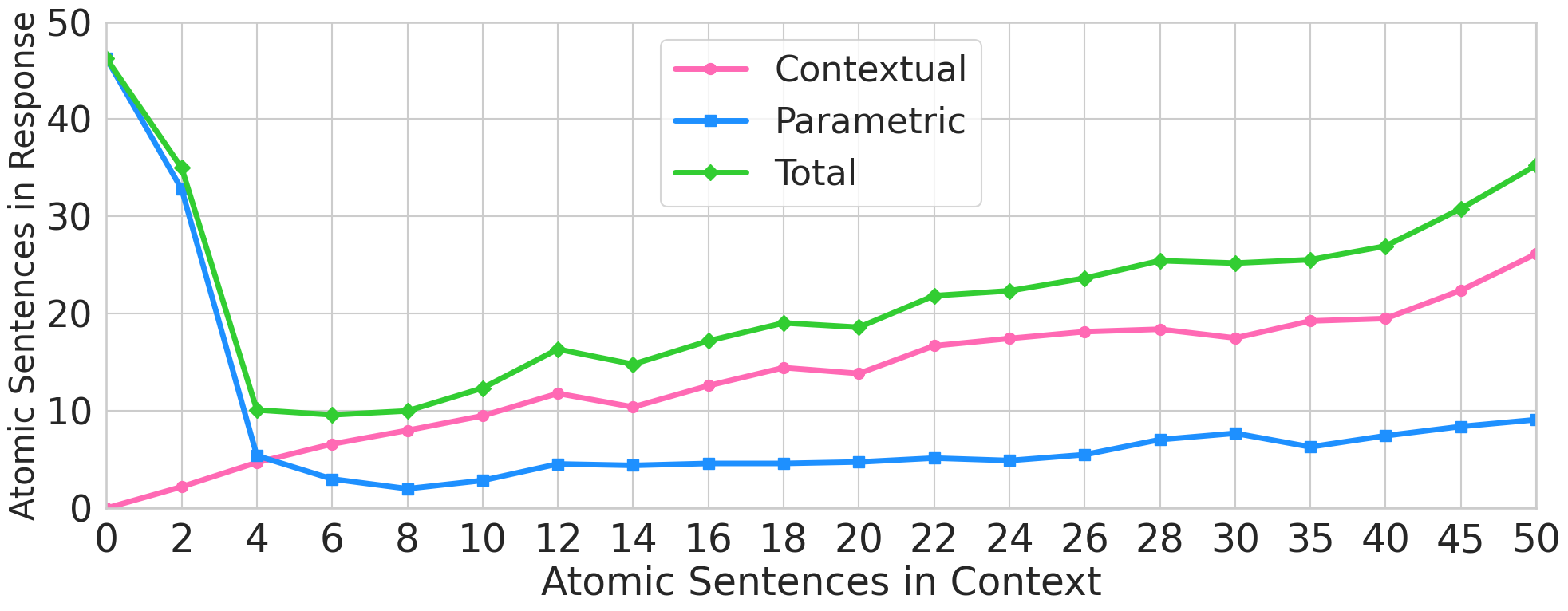}
    \caption{No restriction in question (Llama 3 70B)}
    \label{fig:open_llama370b}
    \end{subfigure}
    \hfill
    \begin{subfigure}[t]{0.48\textwidth}
    \includegraphics[width=\textwidth]{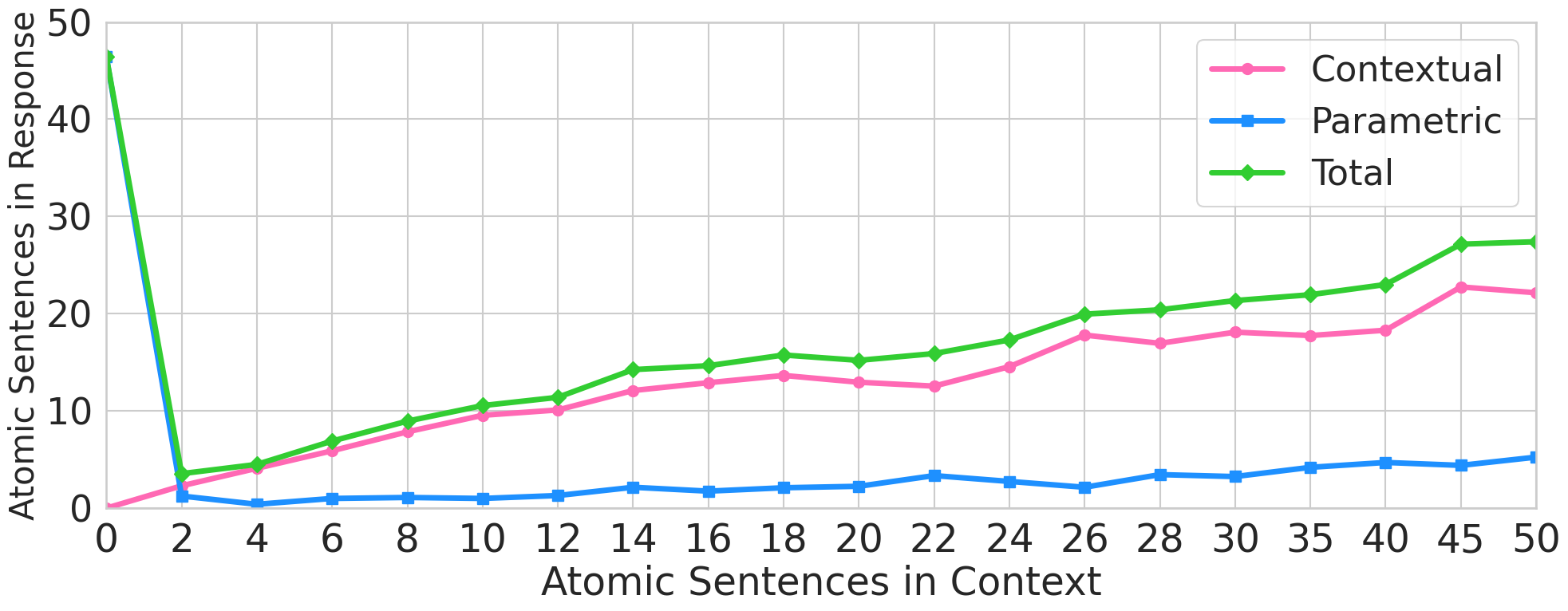}
    \caption{Strict restriction in question (Llama 3 70B)}
    \label{fig:strict_llama370b}
    \end{subfigure}
    \hfill
    \begin{subfigure}[t]{0.49\textwidth}
    \includegraphics[width=\textwidth]{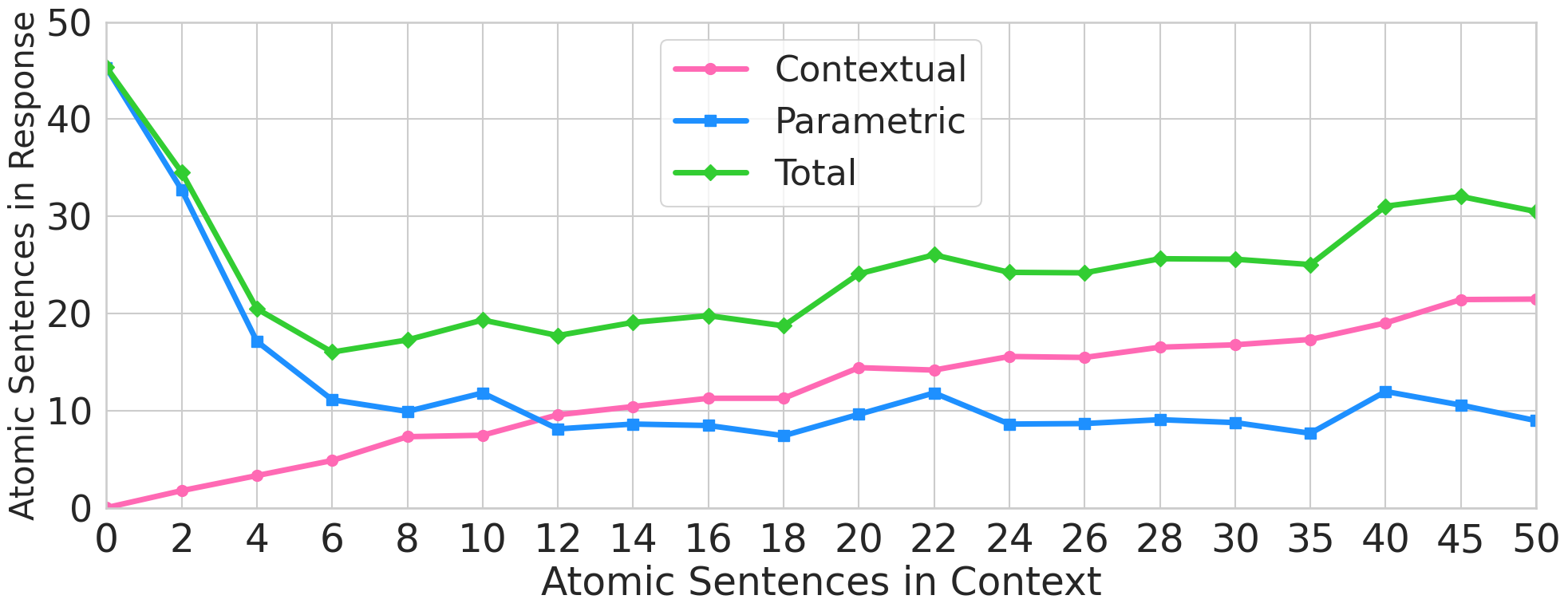}
    \caption{No restriction in question (Llama 3 8B)}
    \label{fig:open_llama38b}
    \end{subfigure}
    \hfill
    \begin{subfigure}[t]{0.49\textwidth}
    \includegraphics[width=\textwidth]{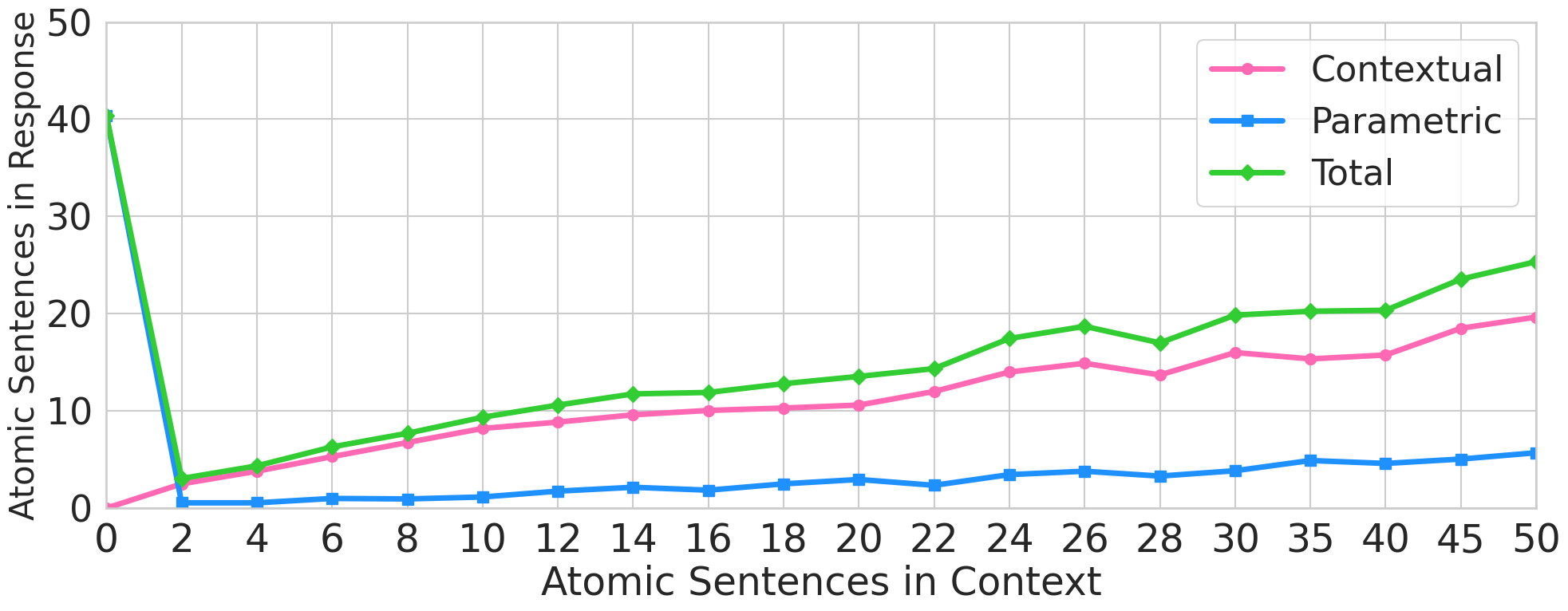}
    \caption{Strict restriction in question (Llama 3 8B)}
    \label{fig:strict_llama38b}
    \end{subfigure}
    \caption{Comparing two methods of prompting the model to answer, one instructing to strictly adhere to provided context and the other not imposing any restriction.}
    \label{fig:ablation_claude}
\end{figure*}

\begin{figure*}[t!]
    \centering
    \begin{subfigure}[t]{0.49\textwidth}
    \includegraphics[width=\textwidth]{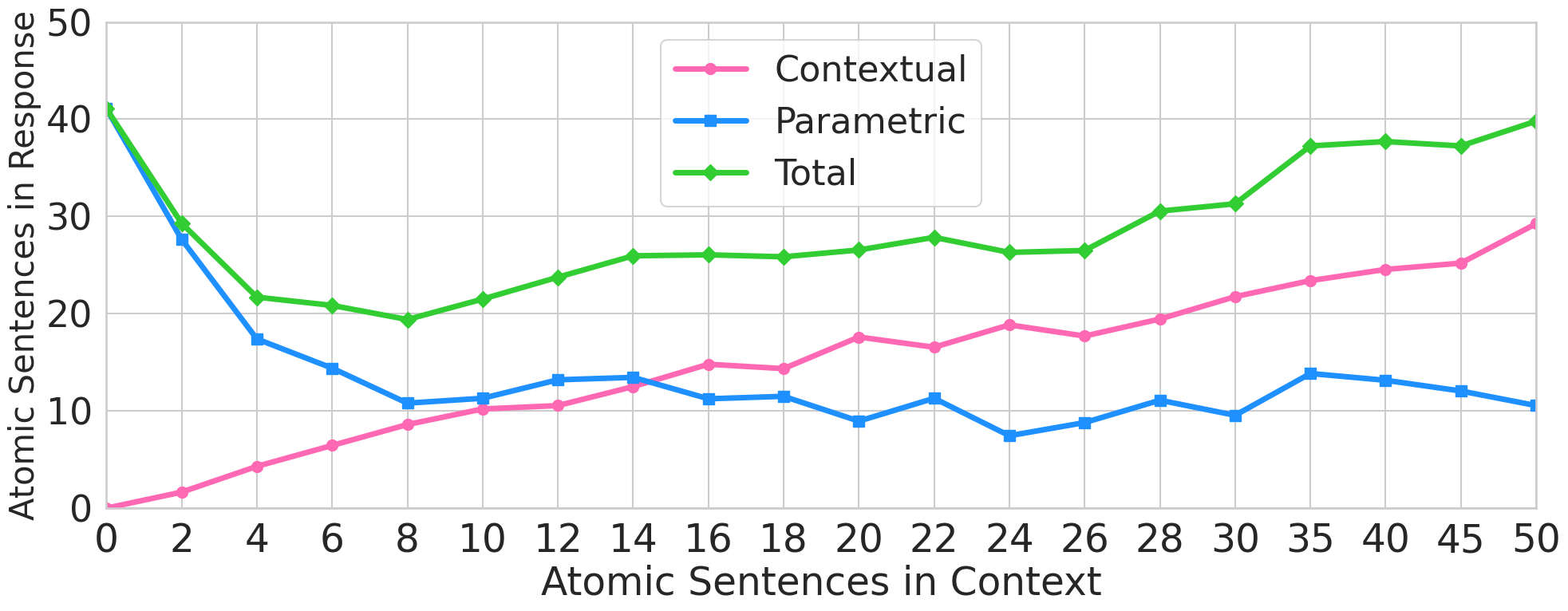}
    \caption{No restriction in question (Mixtral 8x22B)}
    \label{fig:open_mixtral}
    \end{subfigure}
    \hfill
    \begin{subfigure}[t]{0.49\textwidth}
    \includegraphics[width=\textwidth]{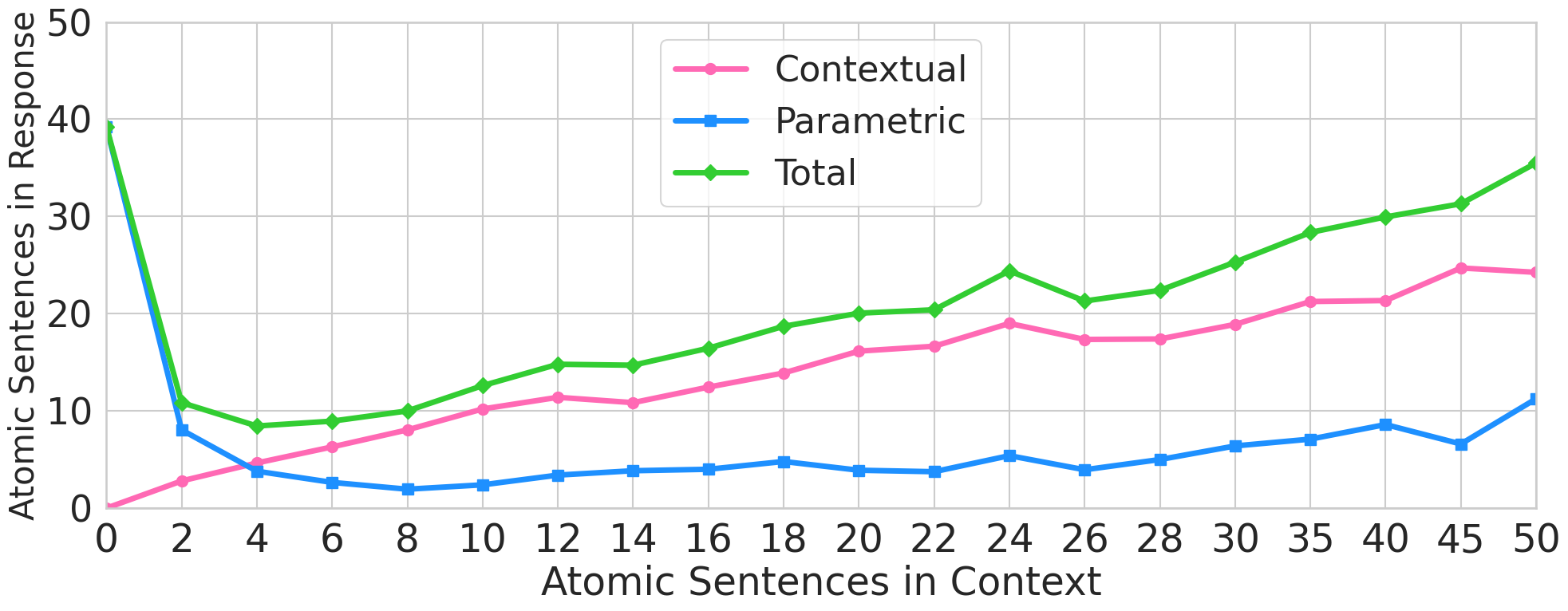}
    \caption{Strict restriction in question (Mixtral 8x22B)}
    \label{fig:strict_mixtral}
    \end{subfigure}
        \begin{subfigure}[t]{0.49\textwidth}
    \includegraphics[width=\textwidth]{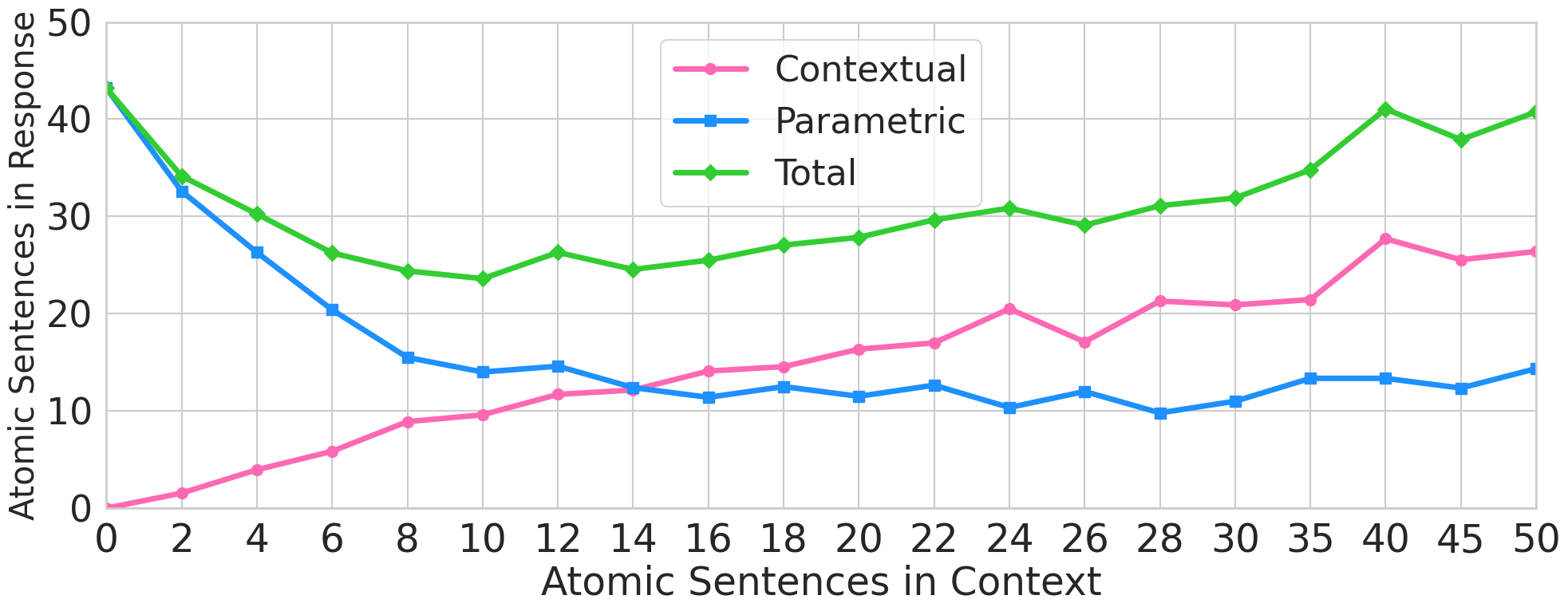}
    \caption{No restriction in question (Mistral 7B)}
    \label{fig:open_mistral}
    \end{subfigure}
    \hfill
    \begin{subfigure}[t]{0.49\textwidth}
    \includegraphics[width=\textwidth]{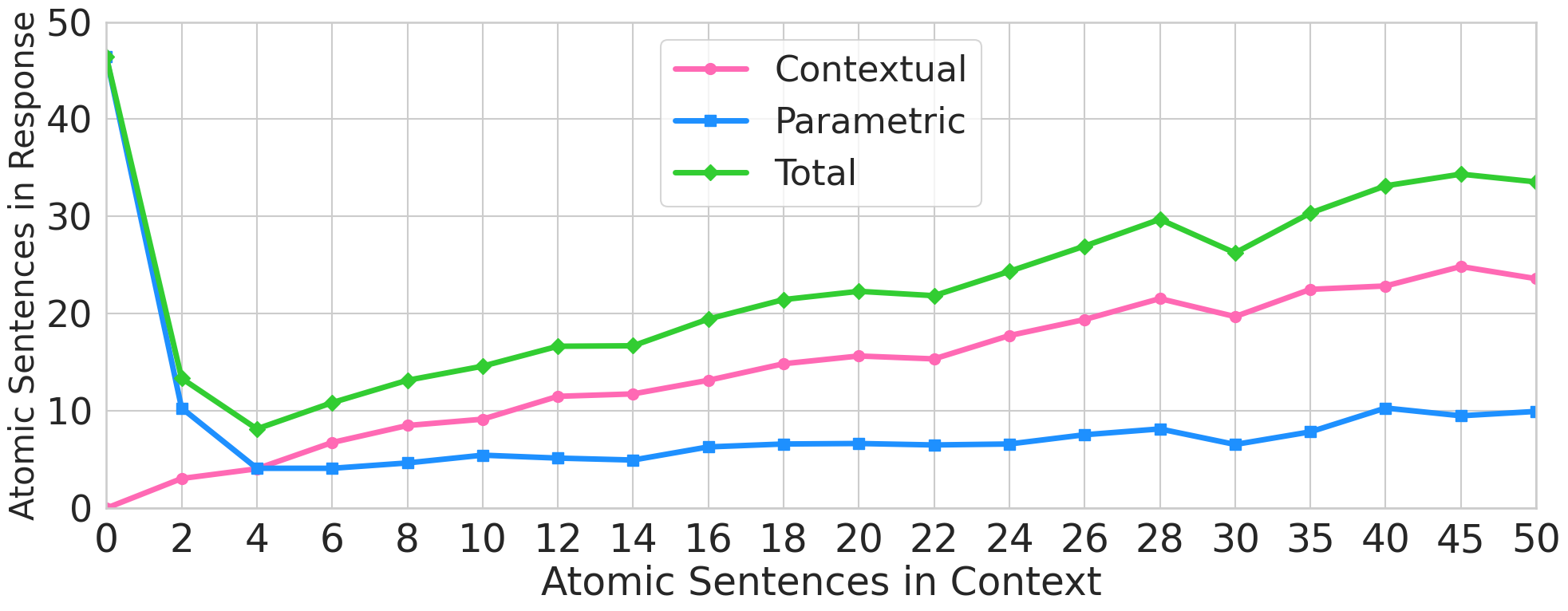}
    \caption{Strict restriction in question (Mistral 7B)}
    \label{fig:strict_msitral}
    \end{subfigure}
    \begin{subfigure}[t]{0.49\textwidth}
    \includegraphics[width=\textwidth]{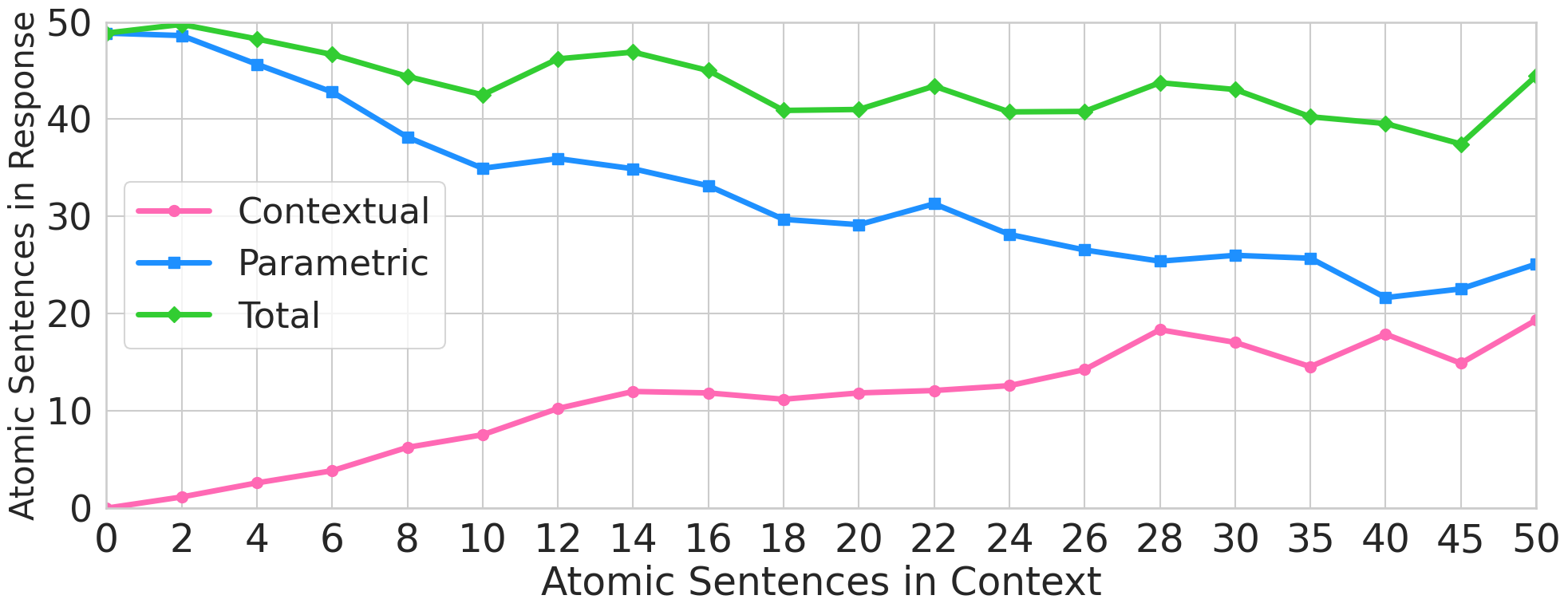}
    \caption{No restriction in question (Phi 3)}
    \label{fig:open_phi}
    \end{subfigure}
    \hfill
    \begin{subfigure}[t]{0.49\textwidth}
    \includegraphics[width=\textwidth]{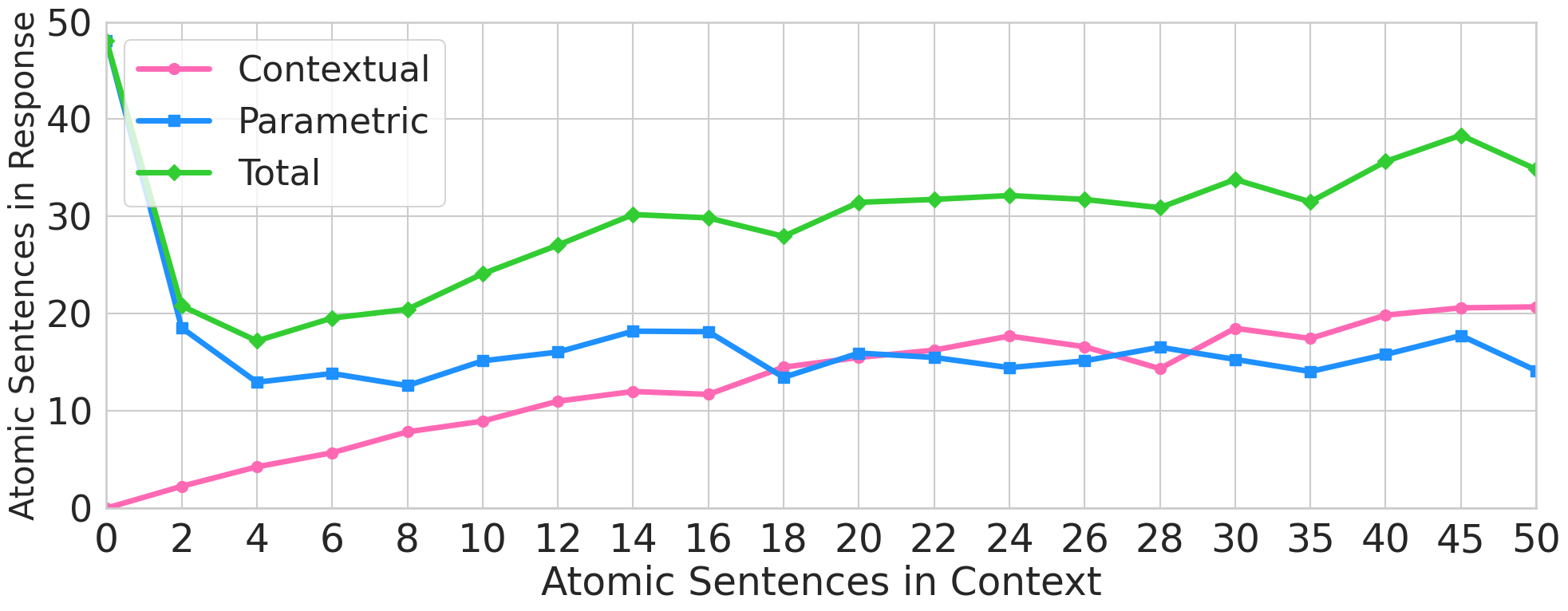}
    \caption{Strict restriction in question (Phi 3)}
    \label{fig:strict_phi}
    \end{subfigure}
    \caption{Comparing two methods of prompting the model to answer, one instructing to strictly adhere to provided context and the other not imposing any restriction.}
    \label{fig:ablation_mistral}
\end{figure*} 

\section{Contextual vs. Parametric while disregarding ambiguous sentences}
\label{app:disregard_infuse}
This section shows the alternative graphs for number of contextual and parametric knowledge in each model (Figure~\ref{fig:ablation_disregarded}) while disregarding sentences received INFUSE score between 0.3 and 0.7.

\begin{figure*}[t!]
    \centering
    \begin{subfigure}[t]{0.49\textwidth}
    \includegraphics[width=\textwidth]{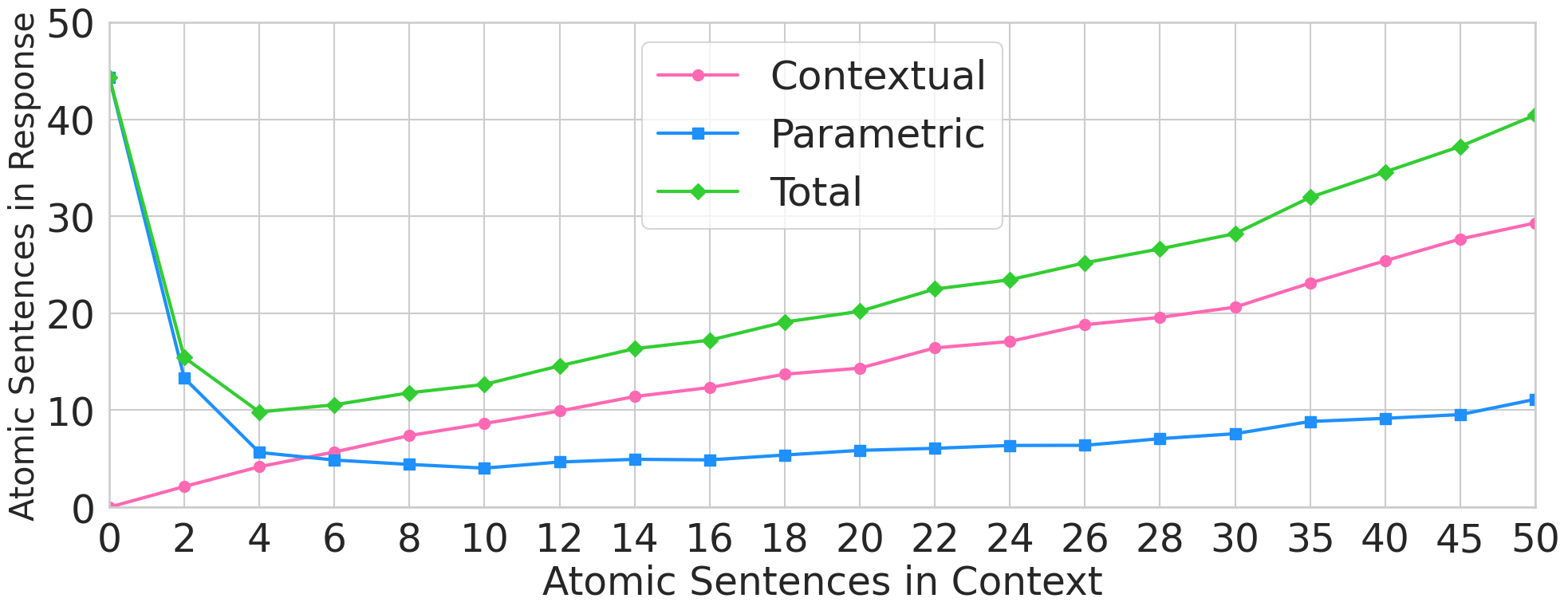}
    \caption{GPT-4o}
    \label{fig:gpt4o_discard}
    \end{subfigure}
    \begin{subfigure}[t]{0.49\textwidth}
    \includegraphics[width=\textwidth]{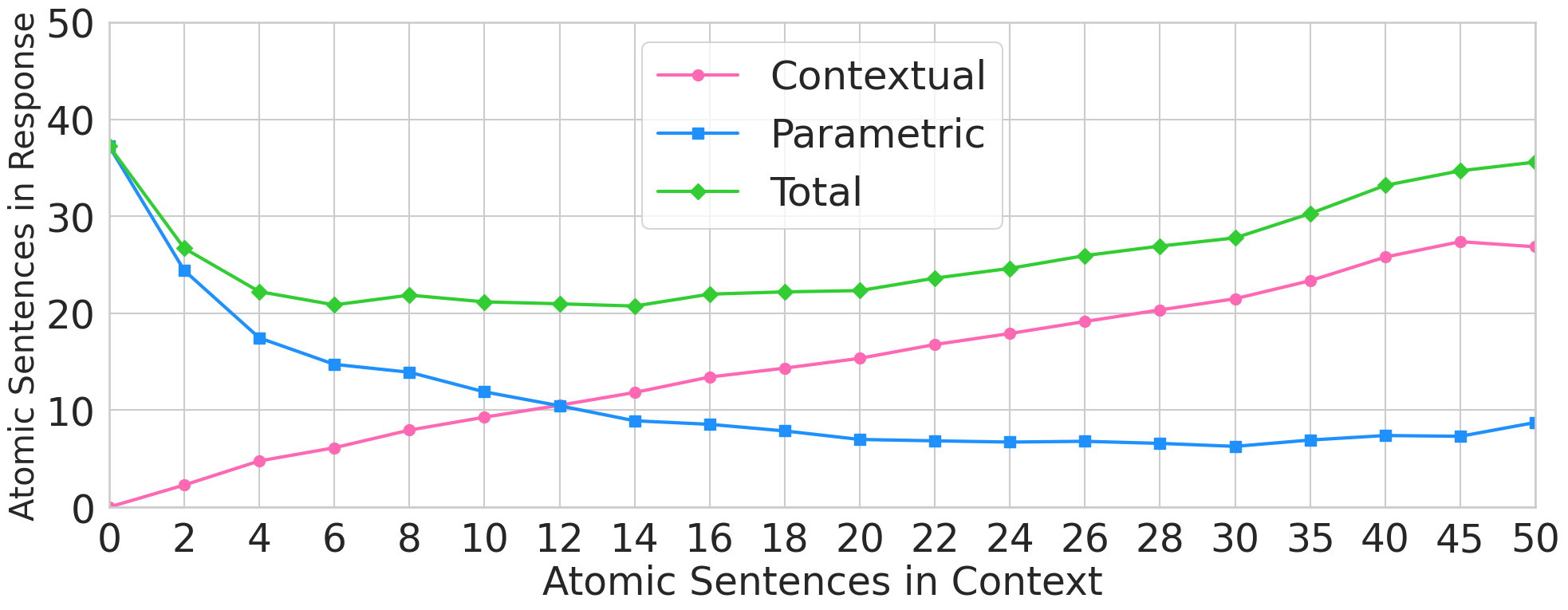}
    \caption{Claude Opus}
    \label{fig:opus_discard}
    \end{subfigure}
    
    \hfill
    
    \begin{subfigure}[t]{0.49\textwidth}
    \includegraphics[width=\textwidth]{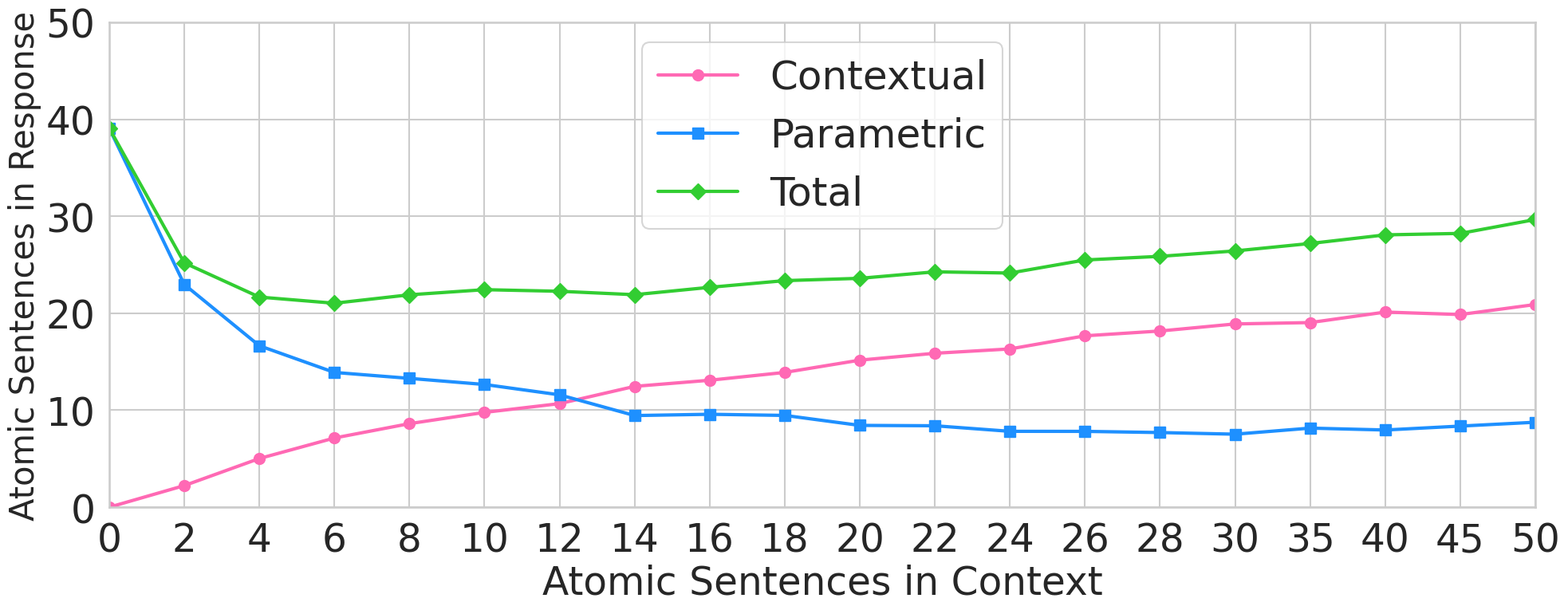}
    \caption{Claude Sonnett}
    \label{fig:sonnett_discard}
    \end{subfigure}
    \begin{subfigure}[t]{0.49\textwidth}
    \includegraphics[width=\textwidth]{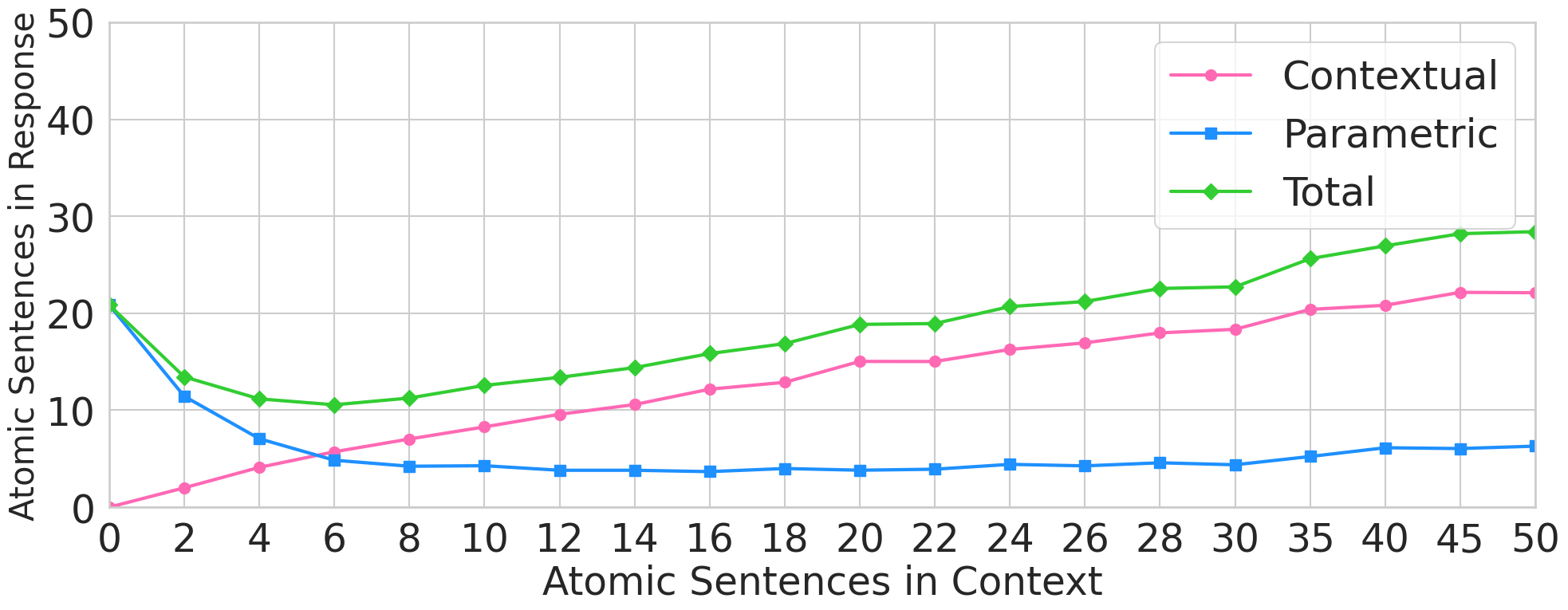}
    \caption{Claude Haiku}
    \label{fig:haiku_discard}
    \end{subfigure}

    \hfill
    
    \begin{subfigure}[t]{0.49\textwidth}
    \includegraphics[width=\textwidth]{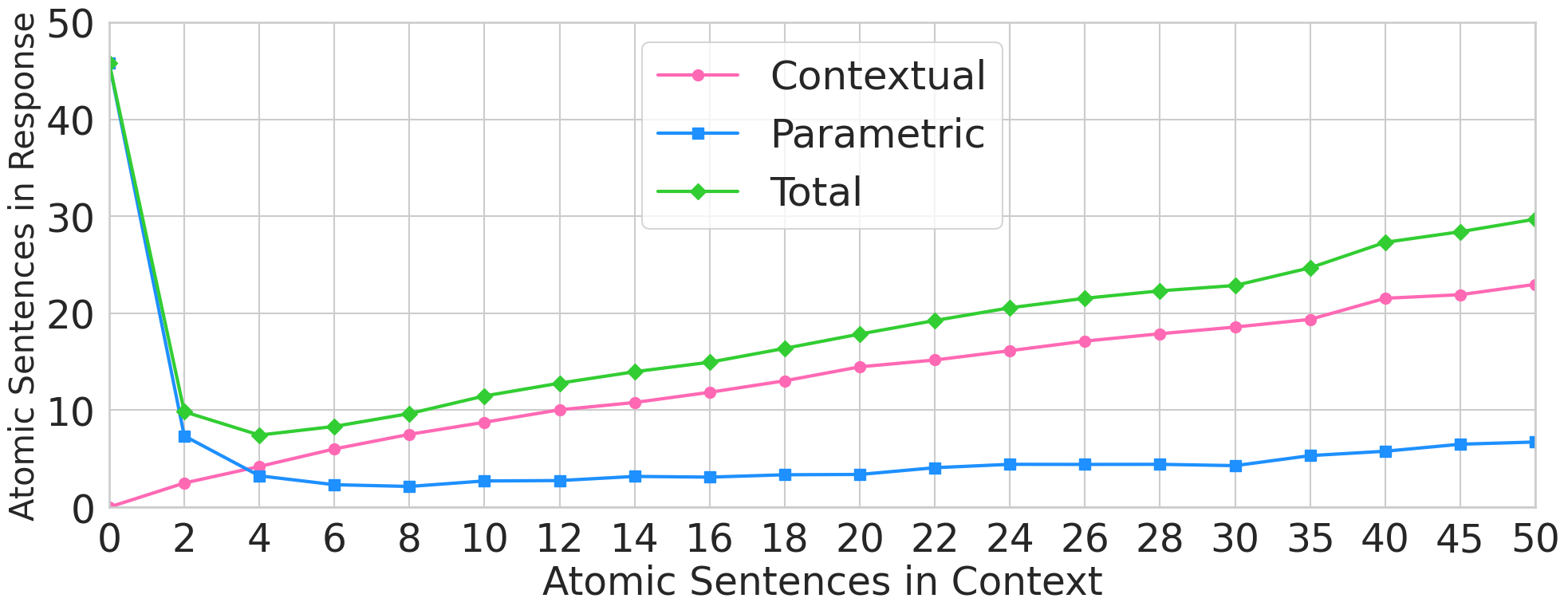}
    \caption{Llama 3 70B}
    \label{fig:llama370b_discard}
    \end{subfigure}
    \begin{subfigure}[t]{0.49\textwidth}
    \includegraphics[width=\textwidth]{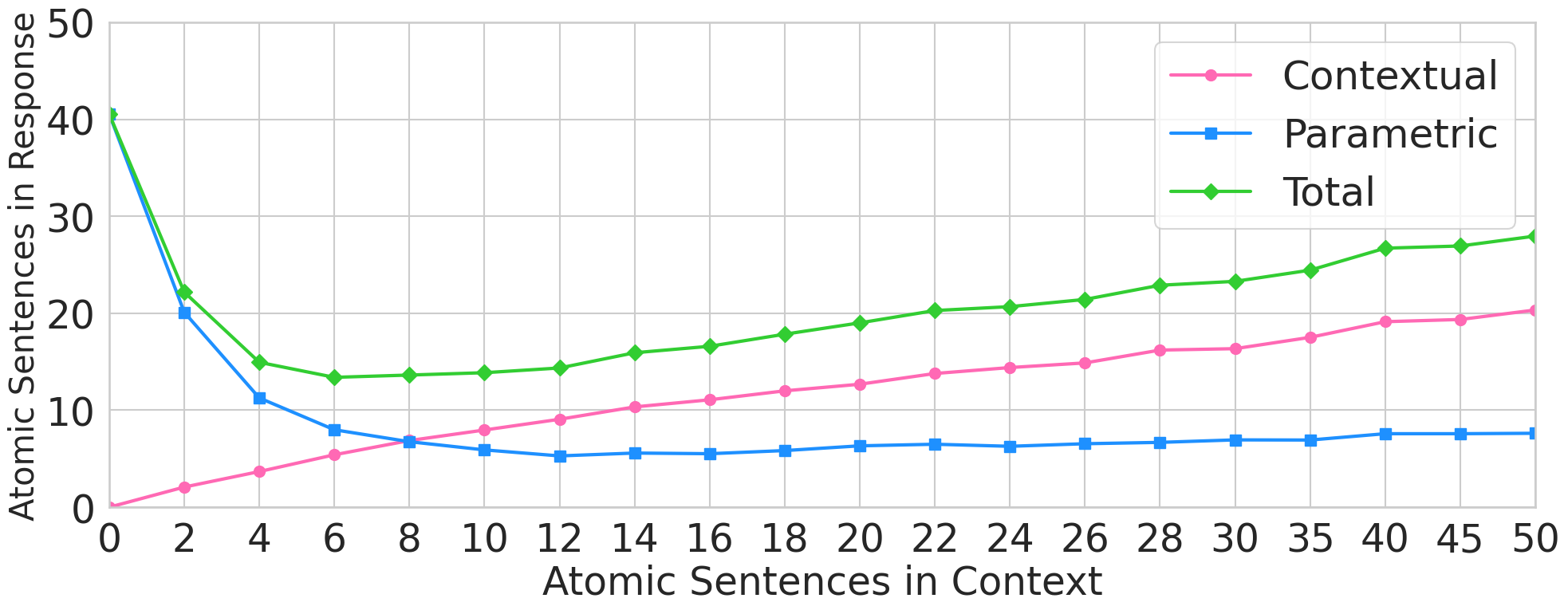}
    \caption{Llama 3 8B}
    \label{fig:llama38b_discard}
    \end{subfigure}

    \hfill
    
    \begin{subfigure}[t]{0.49\textwidth}
    \includegraphics[width=\textwidth]{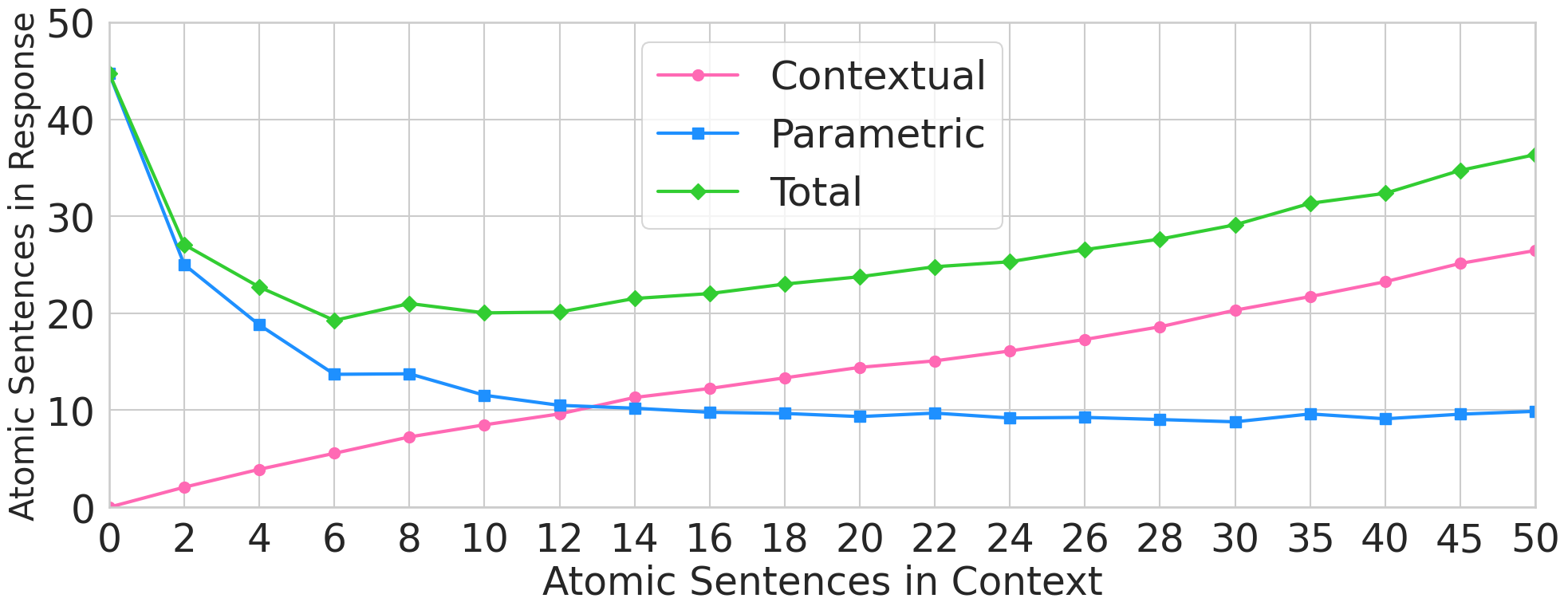}
    \caption{Mistral 7B}
    \label{fig:mistral7b_discard}
    \end{subfigure}
    \begin{subfigure}[t]{0.49\textwidth}
    \includegraphics[width=\textwidth]{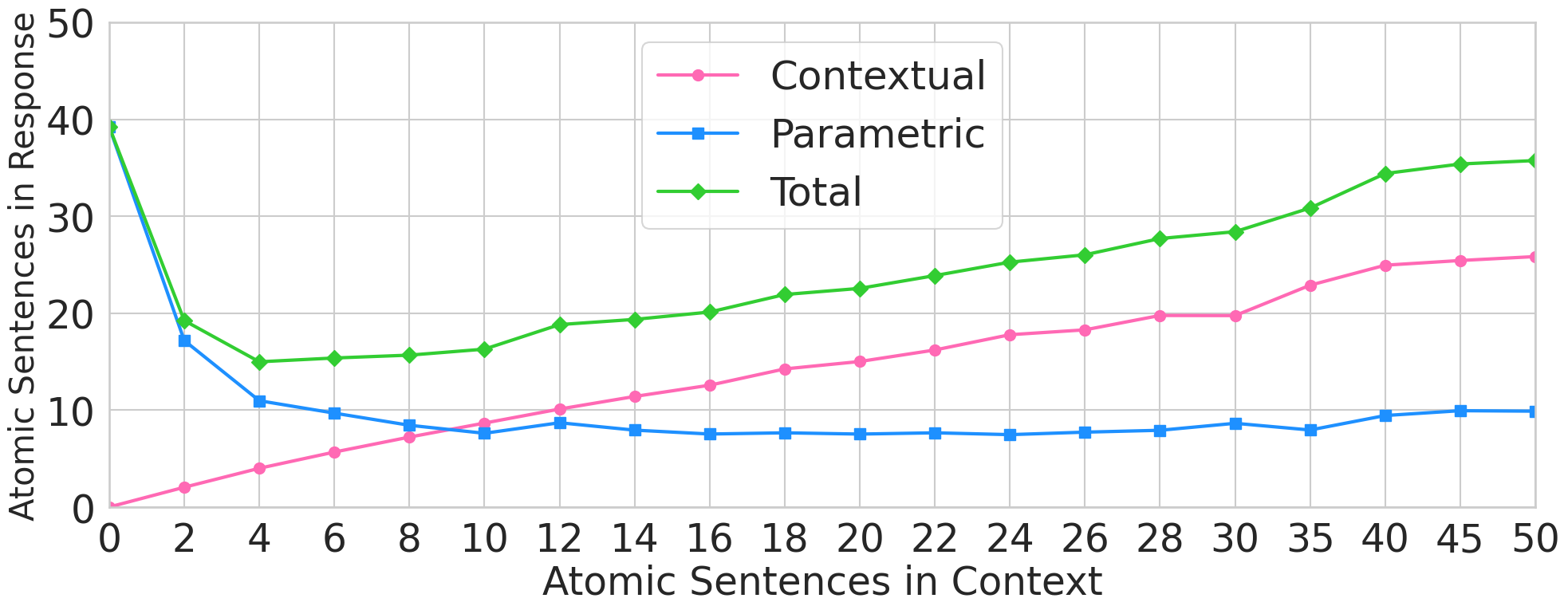}
    \caption{Mixtral 8x22B}
    \label{fig:mixtral_discard}
    \end{subfigure}

    \hfill
    
    \begin{subfigure}[t]{0.49\textwidth}
    \includegraphics[width=\textwidth]{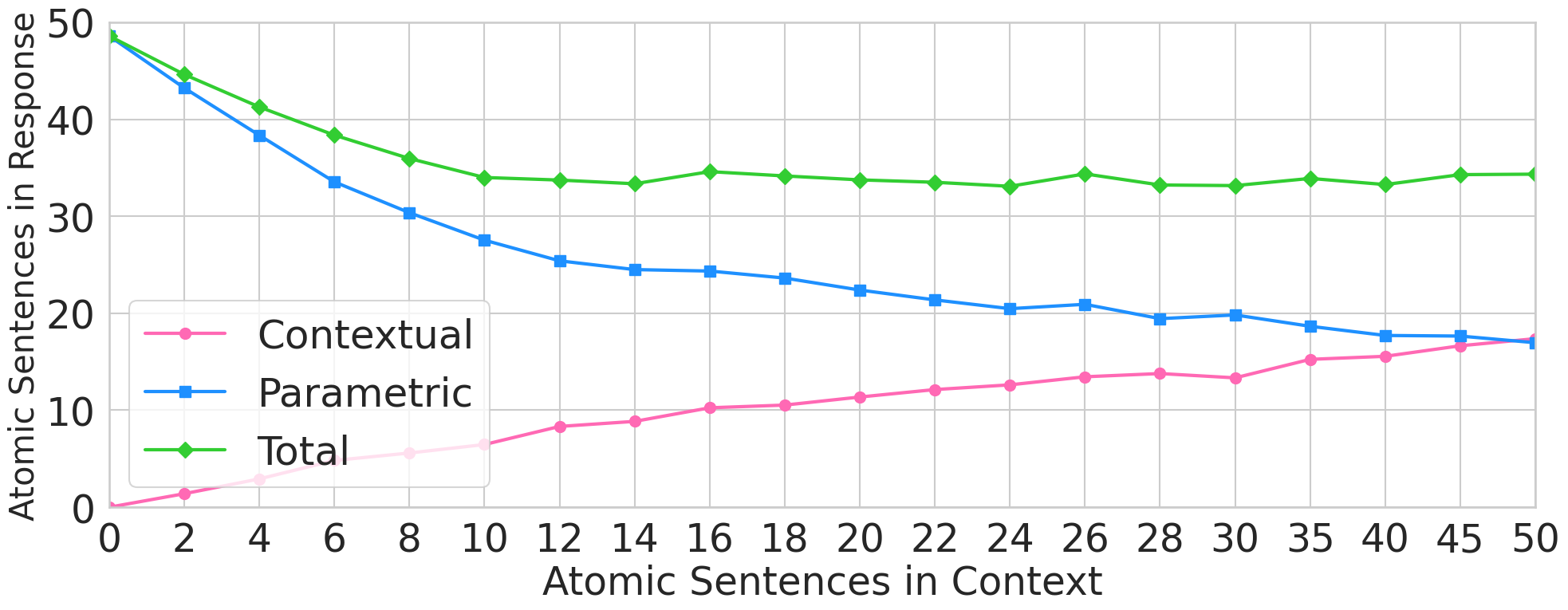}
    \caption{Phi 3}
    \label{fig:phi_discard}
    \end{subfigure}
    
    \caption{Contextual (local), parametric (global), and total sentences in responses for each model while disregarding sentences received INFUSE score between 0.3 and 0.7.}
    \label{fig:ablation_disregarded}
\end{figure*}

\end{document}